\documentclass[conference]{IEEEtran}
\usepackage{times}

\usepackage[numbers]{natbib}
\usepackage{multicol}

\usepackage[bookmarks=true]{hyperref}
\hypersetup{
    colorlinks=true,
    linkcolor=blue,
    filecolor=magenta,      
    urlcolor=blue,
    pdfpagemode=FullScreen,
    }

\usepackage{amsmath}
\usepackage{amssymb}
\usepackage[dvipsnames]{xcolor}
\usepackage{amsthm}
\usepackage{algorithm}
\usepackage[noend]{algpseudocode}
\usepackage{graphics} 
\usepackage{epsfig} 
\usepackage{bm}
\usepackage{stmaryrd}

\usepackage{caption}
\usepackage{subcaption}

\usepackage{tikz}

\usepackage{calrsfs}
\DeclareMathAlphabet{\pazocal}{OMS}{zplm}{m}{n}

\newcommand{\calA}{\pazocal{A}}

\newcommand{\calbC}{\mathcal{C}}
\newcommand{\calC}{\pazocal{C}}

\newcommand{\calE}{\pazocal{E}}

\newcommand{\calF}{\pazocal{F}}

\newcommand{\calI}{\pazocal{I}}

\newcommand{\calL}{\pazocal{L}}

\newcommand{\calN}{\pazocal{N}}

\newcommand{\calO}{\pazocal{O}}

\newcommand{\calP}{\pazocal{P}}

\newcommand{\calQ}{\pazocal{Q}}

\newcommand{\calR}{\pazocal{R}}

\newcommand{\calf}{\mathfrak{f}}

\newcommand{\calm}{\mathfrak{m}}
\newcommand{\caln}{\mathfrak{n}}
\newcommand{\calp}{\mathfrak{p}}
\newcommand{\calq}{\mathfrak{q}}

\newcommand{\cals}{\mathfrak{s}}

\newcommand{\Cb}{\mathbb{C}}
\newcommand{\Db}{\mathbb{D}}
\newcommand{\Eb}{\mathbb{E}}
\newcommand{\Ob}{\mathbb{O}}
\newcommand{\Pb}{\mathbb{P}}

\newcommand{\Rb}{\mathbb{R}}
\newcommand{\Sb}{\mathbb{S}}
\newcommand{\Vb}{\mathbb{V}}
\newcommand{\Zb}{\mathbb{Z}}

\newcommand{\KL}{\Db_{\mathrm{KL}}}

\newcommand{\argmin}{\operatornamewithlimits{argmin}}

\newcommand{\T}{\mathrm{T}}

\newcommand{\Cov}{\Cb \Ob \Vb}

\newcommand{\tr}{\mathrm{tr}}

\newtheorem{proposition}{Proposition}
\newtheorem{corollary}{Corollary}

\newtheorem{remark}{Remark}
\newtheorem{assumption}{Assumption}
\newtheorem{problem}{Problem}

\begin{document}

\title{Distributed Hierarchical Distribution Control for Very-Large-Scale Clustered Multi-Agent Systems}

\author{\authorblockN{Augustinos D. Saravanos,
Yihui Li
and
Evangelos A. Theodorou}
\authorblockA{Georgia Institute of Technology, GA, USA}
\authorblockA{Email: asaravanos@gatech.edu}}



%

\maketitle

\thispagestyle{plain}
\pagestyle{plain}

\begin{abstract}
As the scale and complexity of multi-agent robotic systems are subject to a continuous increase, this paper considers a class of systems labeled as Very-Large-Scale Multi-Agent Systems (VLMAS) with dimensionality that can scale up to the order of millions of agents. In particular, we consider the problem of steering the state distributions of all agents of a VLMAS to prescribed target distributions while satisfying probabilistic safety guarantees. Based on the key assumption that such systems often admit a multi-level hierarchical clustered structure - where the agents are organized into cliques of different levels -  we associate the control of such cliques with the control of distributions, and introduce the Distributed Hierarchical Distribution Control (DHDC) framework. The proposed approach consists of two sub-frameworks. The first one, Distributed Hierarchical Distribution Estimation (DHDE), is a bottom-up hierarchical decentralized algorithm which links the initial and target configurations of the cliques of all levels with suitable Gaussian distributions. The second part, Distributed Hierarchical Distribution Steering (DHDS), is a top-down hierarchical distributed method that steers the distributions of all cliques and agents from the initial to the targets ones assigned by DHDE. Simulation results that scale up to two million agents demonstrate the effectiveness and scalability of the proposed framework. The increased computational efficiency and safety performance of DHDC against related methods is also illustrated. The results of this work indicate the importance of hierarchical distribution control approaches towards achieving safe and scalable solutions for the control of VLMAS. A video with all results is \href{https://youtu.be/0QPyR4bD2q0}{available here}.
\end{abstract}

\IEEEpeerreviewmaketitle

\section{Introduction}

Multi-agent systems in robotics are experiencing an increasing popularity with several significant applications such as multi-robot coordination \cite{cortes2017coordinated}, navigating fleets of vehicles \cite{ren2008distributed}, guiding teams of UAVs \cite{zhang2020multi} and swarm robotics \cite{bayindir2016review}, to name only a few. As the scale and complexity of such systems are continuously growing, a great requirement has emerged for developing algorithmic frameworks that benefit from a distributed structure, high computational efficiency, low communication requirements, and therefore, scalability. In addition, as uncertainty is an integral component of multi-agent systems,  associating such methods with safety guarantees remains of paramount importance.

Most of the existing literature in multi-robot control, has considered systems that range from a handful of units to hundreds or thousands of agents. Some notable approaches can be found in the fields of optimal control \cite{parker2009path, pereira2022decentralized, saravanos2022distributed_ddp, zhu2021adaptive}, path planning \cite{bennewitz2002finding, ferrari1998multirobot, raju2021online}, swarm robotics \cite{bayindir2016review, mohan2009extensive, nedjah2019review, rubenstein2014programmable} and multi-agent reinforcement learning \cite{dasari2020robonet, hernandez2019survey, huttenrauch2019deep, zhang2021multi}. Nevertheless, empirical demonstrations show that the scalability of most methods from the previous classes is practically limited in the order of a few thousands of robots.

\begin{figure}[t]
\centering
\begin{tikzpicture}
    \node[anchor=south west,inner sep=0] at (0,0){\includegraphics[width=0.49\textwidth, trim={4cm 3cm 13.7cm 1cm},clip]{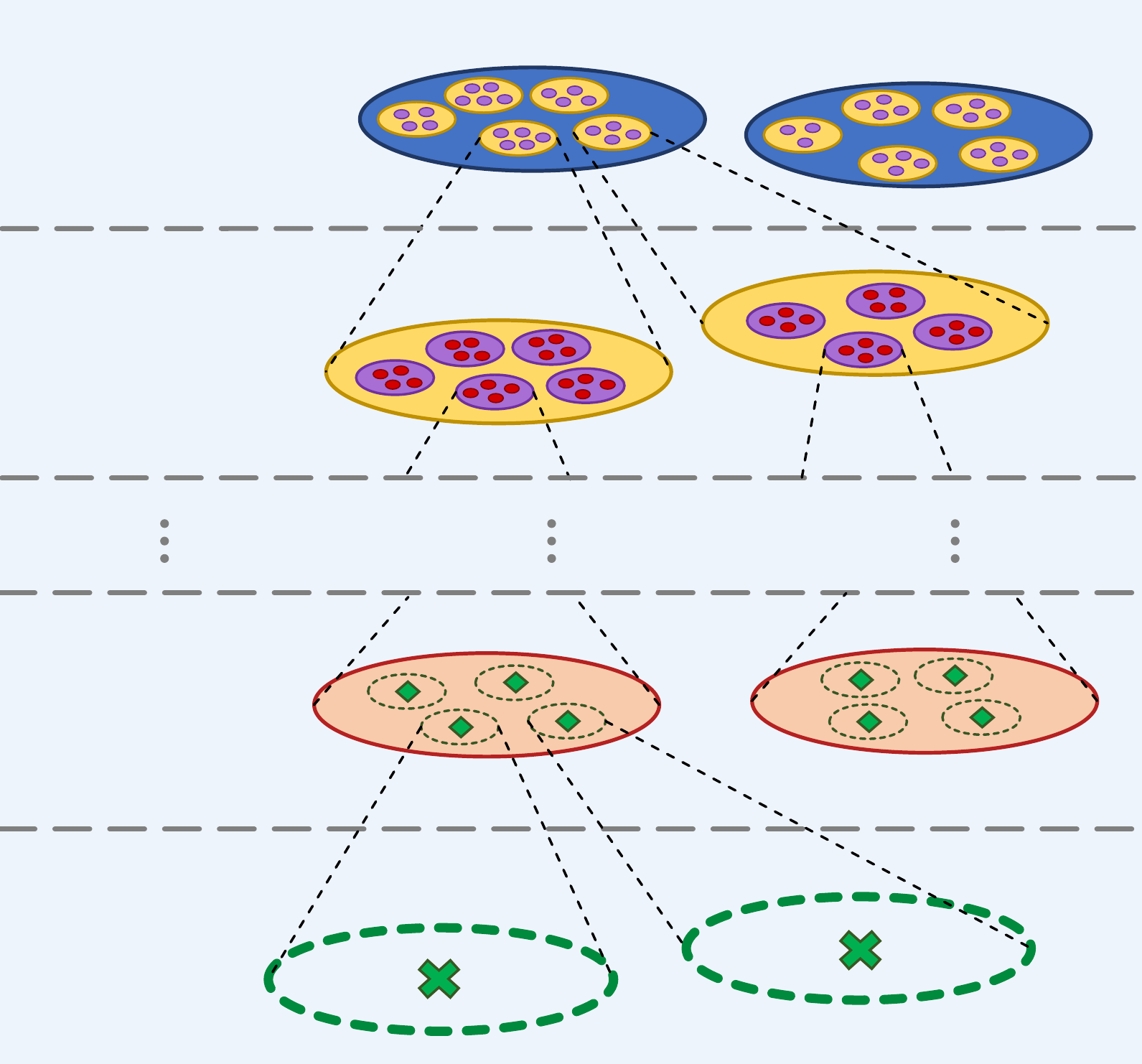}
    };
    \node[align=center] (c) at (1.0, 7.45) {\textbf{Level} $1$ };
    \node[align=center] (c) at (1.0, 5.55) {\textbf{Level} $2$ };
    \node[align=center] (c) at (1.30, 2.75) {\textbf{Level} $L-1$ };
    \node[align=center] (c) at (1.0, 0.93) {\textbf{Level} $L$ };
\end{tikzpicture}
\caption{A VLMAS with a hierarchical clustered structure of $L$ levels. Level $L$ corresponds to individual robots and their distributions (green). All robots are organized into level-($L-1$) cliques (orange) which are then also organized into level-($L-2$) cliques and so on, up to level-$1$ cliques (blue).}
\label{hier_intro_figure}
\end{figure}
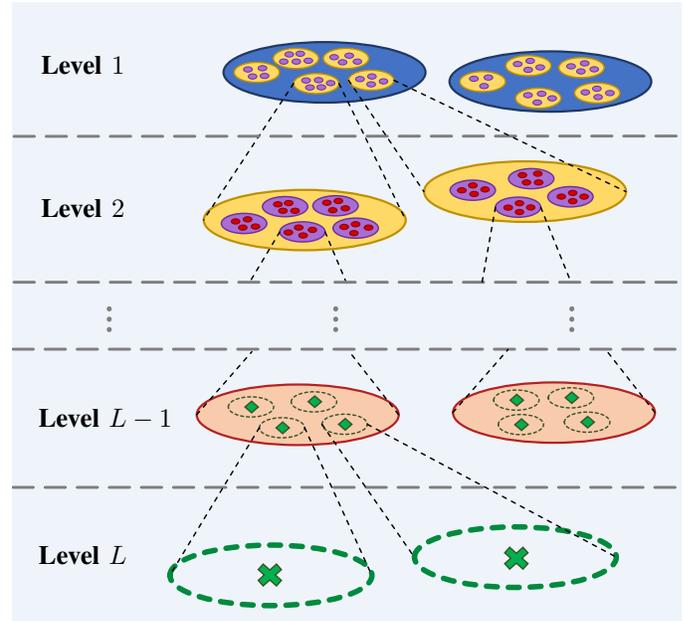

This paper considers a class of systems, labeled as Very-Large-Scale Multi-Agent Systems (VLMAS), that can scale up to the order of millions of robots. To also account for potential uncertainties in VLMAS, all agents are modeled with stochastic dynamics. To our best knowledge, the literature for addressing the control of such systems in a safe, distributed and scalable manner while operating under uncertainty is very scarce. The key insight of this work is that VLMAS often admit a hierarchical clustered structure (Fig. \ref{hier_intro_figure}) where robots are arranged into cliques, which are then organized into greater cliques etc. Such a structure is highly suitable for the proposed direction of hierarchical distribution control. 

Multi-robot control approaches that exploit such hierarchies appear to be quite few in the literature. A multi-robot navigation method that used hierarchical clustering to identify the formation of groups of robots and guide them to their targets was proposed in \cite{arslan2016coordinated}. Furthermore, a distributed algorithm for the coordination of clusters of robots was recently presented in \cite{hu2021decentralized}. While these works have hinted towards the potential of distributed hierarchical control in robotics, their approaches were still only applicable to small-scale multi-robot teams and unrelated to the control of distributions.

As safety requirements in robotics and control are of great significance, covariance steering (CS) theory has recently emerged as a promising approach for guiding the state distribution of a system to prescribed targets while providing probabilistic safety guarantees \cite{bakolas2018finite, balci2021covariance, p:chen2015covariance1, 
liu2022optimal}. Successful robotics applications can be found in trajectory optimization \cite{balci2022constrained, yin2022trajectory}, path planning \cite{p:okamoto2019pathplanning}, flight control \cite{benedikter2022convex, kotsalis2021convex, rapakoulias2023discrete}, multi-robot systems \cite{saravanos2021distributed, saravanos2022dmpcs} and robotic manipulation \cite{lee2022hierarchical}, to name a few. While the main barrier for applying CS methods for multi-agent stochastic control was due to their significant computational requirements, recent distributed optimization based approaches \cite{saravanos2021distributed, saravanos2022dmpcs} have shown that CS is a viable option for multi-agent systems. Nevertheless, the scalability of the aforementioned methods appears to be practically limited to systems with tens or hundreds of agents. In addition, to our best knowledge, combining CS with the control of the distributions of clusters of agents has not been considered yet in the literature.

In this paper, we aspire to surpass the appearing limitations of current CS approaches, by exploiting the fact that VLMAS systems can be subject to a hierarchical clustered structure. In such a multi-level hierarchical setup, the agents are organized into cliques, which are then organized into greater cliques, and so on. Our key insight lies in the fact that CS theory can also be utilized for the control of such cliques in addition to the control of individual agents. Based on this fact, we introduce a novel distributed method for the control of VLMAS, named Distributed Hierarchical Distribution Control (DHDC). The DHDC framework consists of two separate sub-frameworks. The first one, Distributed Hierarchical Distribution Estimation (DHDE), associates all cliques with suitable Gaussian distributions that satisfy the hierarchical structure, while the second one, Distributed Hierarchical Distribution Steering (DHDS), utilizes CS to drive the distributions of all cliques and agents towards their assigned targets. The specific contributions of this work can be listed as follows:
\begin{enumerate}
\item We illustrate how covariance steering theory can be fused with the control of clusters of agents that are linked through a hierarchical structure.
\item We propose DHDE, a bottom-up hierarchical distributed approach for estimating the optimal random distributions to be associated with the cliques of each level of the hierarchy.
\item We propose DHDS, a top-down hierarchical distributed approach for steering the distributions of all cliques and agents to their targets, by exploiting the information acquired with DHDE.
\item We demonstrate the effectiveness and scalability of the proposed approaches through simulation experiments on systems with up to two million agents. 
\end{enumerate}

To our best knowledge, the proposed approach is one of the few existing methods for the distributed and safe control of stochastic systems in the VLMAS scale. In addition, DHDC greatly outperforms the scalability of all current CS approaches for multi-agent control, and therefore, paves the way for the application of CS theory to robotic systems of a much larger scale. Furthermore, this work highlights the potential of hierarchical distributed optimization methods that utilize distribution characteristics for the control of large-scale multi-agent systems.

\section{Problem Statement}

\subsection{Notation}

The set of $n \times n$ symmetric positive semi-definite (definite) matrices is denoted with $\Sb_n^+$ ($\Sb_n^{++}$). Given two matrices $A,B \in \Rb^{n \times n}$, the matrix inequality $A \succeq B$ refers to $A-B \in \Sb_n^+$. Given a random variable (r.v.) $x \in \Rb^n$, its expectation and covariance are given by $\Eb[x] \in \Rb^n$ and $\Cov[x] \in \Sb_n^+$, respectively. If a r.v. $x \in \Rb^n$ is such that $x \sim \calN (\mu, \Sigma)$, then $x$ is subject to a multivariate Gaussian distribution with mean $\mu = \Eb[x]$ and covariance $\Sigma = \Cov[x]$. 
In addition, $x \in \calE_{\theta} [\mu,\Sigma]$ implies that $x \in \Rb^n$ lies within the $\theta$-probability confidence ellipsoid of $\calN (\mu, \Sigma)$, i.e., $\calE_{\theta} [\mu,\Sigma]: (x-\mu)^\T \Sigma^{-1} (x-\mu) \leq \alpha$, where $\alpha = f_{\chi^2, \nu}^{-1} (\theta)$ and $f_{\chi^2, \nu}^{-1} (\cdot)$ is the inverse cumulative distribution function of the chi-square distribution with $\nu$ degrees of freedom. 
The Kullback–Leibler (KL) divergence of a probability distribution $\calP$ from another distribution $\calQ$ is denoted with $\KL(\calP \| \calQ)$.
Furthermore, the cardinality of a set $\calA$ is given by $|\calA|$. Finally, $\llbracket a, b \rrbracket$ denotes the integer set $[a,b] \cap \Zb$ for any $a,b \in \Rb$ with $a \leq b$.


\subsection{Problem Formulation}

Let us consider a VLMAS given by the set $\calR = 
\{1,\dots,M\}$, where $M$ is the total number of agents. Each agent $i \in \calR$ is subject to the following homogeneous discrete-time, stochastic, linear dynamics 
\begin{equation}
\label{linear dynamics}
x_{i,k+1} = A x_{i,k} + B u_{i,k} + w_{i,k},
\end{equation}
where $x_{i,k} \in \Rb^{n_x}$ and $u_{i,k} \in \Rb^{n_u}$ are the state and control of the $i$-th agent at time $k$, $w_{i,k} \in \Rb^{n_x} $ is process noise such that $w_{i,k} \sim \calN(0,W)$ with $W \in \Sb_{n_x}^+$, and $A \in \Rb^{n_x \times n_x}$, $B \in \Rb^{n_x \times n_u}$. If we denote the time horizon with $N$, i.e., $k \in \llbracket 0,N \rrbracket$, then the full state, control and noise sequences of agent $i$ are given by $x_i = [x_{i,0}; \dots; x_{i,N}]$, $u_i = [u_{i,0}; \dots; u_{i,N-1}]$ and $w_i = [w_{i,0}; \dots; w_{i,N-1}]$, respectively. All initial states $x_{i,0}$ are subject to $x_{i,0} \sim \calN(\mu_{i,0}, \Sigma_{i,0})$, where $\mu_{i,0} \in \Rb^{n_x}$ and $\Sigma_{i,0} \in \Sb_{n_x}^{++}$ are known. 

Let us now introduce the VLMAS \textit{distribution control problem}. The main objective is to steer the distributions of the terminal states $x_{i,N}$ of all agents to prescribed target distributions $\calN(\mu_{i,\mathrm{f}}, \Sigma_{i,\mathrm{f}})$ with $\mu_{i,\mathrm{f}} \in \Rb^{n_x}$ and $\Sigma_{i,\mathrm{f}} \in \Sb_{n_x}^{++}$. This can be achieved by enforcing the following constraints
\begin{align}
\label{terminal mean constraint}
\Eb[x_{i,N}] & = \mu_{i,\mathrm{f}},
\\
\label{terminal cov constraint}
\Cov [x_{i,N}] & \preceq \Sigma_{i,\mathrm{f}},
\end{align}
for every agent $i \in \calR$. Furthermore, the position of each agent in space is given by $p_{i,k} = H x_{i,k} \in \Rb^{n_p}$, with $H \in \Rb^{n_p \times n_x}$ defined accordingly. To simplify exposition, in this work, we consider agents that operate on a 2D plane, i.e., $n_p = 2$. Nevertheless, all proposed ideas are readily extendable for 3D multi-robot systems. 
The following probabilistic inter-agent collision avoidance constraints must also be satisfied between all agents,
\begin{equation}
\label{collision avoidance constraints}
\Pb \left( 
\| p_{i,k} - p_{j,k} \|_2 \geq d_{\mathrm{inter}}
\right) \geq \theta, 
\quad \forall i,j \in \calR, \ i \neq j, 
\end{equation}
for all $k \in \llbracket 0, N \rrbracket$, where $d_{\mathrm{inter}}$ is the minimum allowed distance between two agents and $\theta \in (0.5,1)$. In addition, we assume the existence of a set $\calO = \{ 1, \dots, O \}$ of circle obstacles and consider the following probabilistic safety constraints
\begin{equation}
\label{obs avoidance constraints}
\Pb \left( 
\| p_{i,k} - p_o \|_2 \geq d_{\mathrm{obs}} + r_o
\right) \geq \theta, 
\quad \forall i \in \calR, \ \forall o \in \calO,
\end{equation}
for all $k \in \llbracket 0, N \rrbracket$, where $r_o$ is the radius of obstacle $o \in \calO$ and $d_{\mathrm{obs}}$ is the minimum allowed distance between an agent and an obstacle. Finally, all agents also aim to minimize their control effort through the following individual costs
\begin{equation}
J_i (u_i) = \sum_{k = 0}^{N-1} \Eb[u_{i,k}^\T R u_{i,k}], \quad i \in \calR. 
\end{equation}
where $R \in \Sb_n^{++}$.
Consequently, the VLMAS distribution control problem can be formulated as follows.
\begin{problem}[VLMAS Distribution Control Problem]
\label{VLMAS original problem}
Find the optimal control input sequences $u_i^*, \ \forall i \in \calR$, such that 
\begin{align*}
& \{ u_i^* \}_{i \in \calR} = \argmin \sum_{i \in \calR} J_i (u_i)
\\[0.1cm] 
& ~~~ \mathrm{s.t.} \quad \eqref{linear dynamics}, \eqref{terminal mean constraint}, \eqref{terminal cov constraint}, \eqref{collision avoidance constraints}, \eqref{obs avoidance constraints}.
\end{align*}
\end{problem}

\begin{figure*}[!t]
\centering
\includegraphics[width=1\textwidth, trim={0.8cm 1cm 0.5cm 2cm},clip]{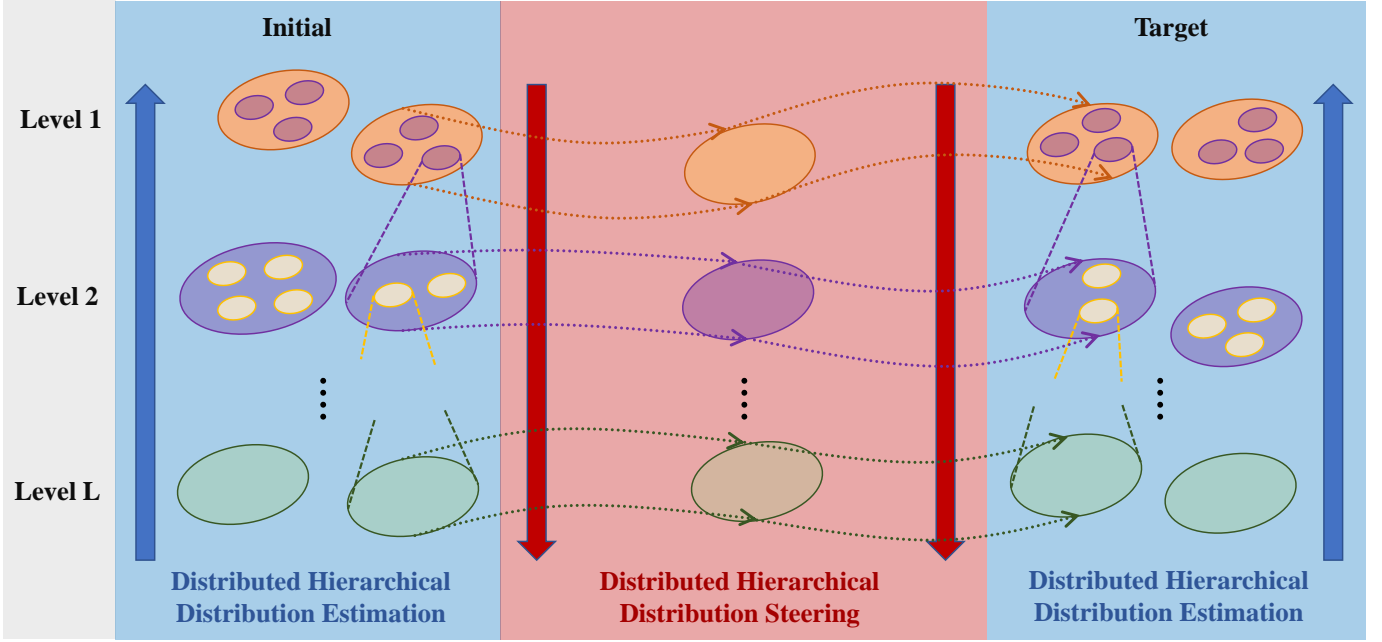}
\caption{The DHDC framework consists of two parts. The first part (DHDE) is a bottom-up approach that estimates the initial and target random distributions that correspond to all cliques of levels $\ell \in \llbracket 1, L-1 \rrbracket$, such that the assumed hierarchical clustered structure is satisfied. The second part (DHDS) is a top-down approach that steers the distributions of the cliques/agents of all levels based on the initial and target distributions provided by DHDE.}
\label{fig: hierarchical}
\end{figure*}

\subsection{Hierarchical Clustered Structure}

This work focuses on VLMAS that are subject to a known \textit{hierarchical clustered structure} (Fig. \ref{hier_intro_figure}). This hierarchy consists of $L$ levels with each level denoted with $\ell \in \{1, \dots, L \}$. The bottom level $\ell = L$ corresponds to individual agents. All agents are organized into level-$(L-1)$ cliques $\calC_i^{L-1} \in \calR^{L-1}$, which are then organized into level-$(L-2)$ cliques $\calC_i^{L-2} \in \calR^{L-2}$, and so on, where $\calR^{\ell}$ denotes the set of all cliques in each level $\ell \in \llbracket 1,L \rrbracket$. If by convention, we let the level-$L$ cliques $\calC_i^L$ correspond to individual agents $i \in \calR$, i.e., $\calR^L \equiv \calR$, then the aforementioned structure can be formally stated as follows. 

\begin{assumption}[Hierarchical Clustered Structure]
\label{assumption hierarchical}
For all levels $\ell \in \llbracket 2, L \rrbracket$, it holds that for every clique $\calC_i^{\ell} \in \calR^{\ell}$, there exists a clique $\calC_a^{\ell-1} \in \calR^{\ell-1}$ such that
\begin{equation}
\calC_i^{\ell} \subseteq \calC_a^{\ell-1}. 
\end{equation}
In other words, every clique in level $\ell$ belongs to a ``parent'' clique of level $\ell-1$. Furthermore, for all levels $\ell \in \llbracket 1, L \rrbracket$, it holds that 
\begin{equation}
\calC_i^{\ell} \cap \calC_j^{\ell} = \emptyset, \quad \forall \calC_i^{\ell},\calC_j^{\ell} \in \calR^{\ell}, \ i \neq j,    
\end{equation}
i.e., the intersection between cliques of the same level is always the empty set.
\end{assumption}
Furthermore, to lighten the notation, given a clique $\calC_i^{\ell}$, we consider the statements $\calC_i^{\ell} \in \calA$ and $i \in \calA$ to be equivalent for any arbitrary set $\calA$. A direct consequence of Assumption \ref{assumption hierarchical} is the following.

\begin{corollary}
For all levels $\ell \in \llbracket 2, L \rrbracket$, every clique $\calC_i^{\ell}$ only has one ``parent'' clique $\calC_a^{\ell-1}$.
\end{corollary}
Subsequently, we introduce the notion of \textit{neighbor cliques}. Given a clique $\calC_i^{\ell}$, the set of its neighbor cliques is defined as $\caln[\calC_i^{\ell}]$. Of course, if $\ell \in \llbracket 2, L \rrbracket$, then all cliques $\calC_j^{\ell}$ with $j \in \caln[\calC_i^{\ell}]$ have the same ``parent'' clique as $\calC_i^{\ell}$, i.e., if $\calC_i^{\ell} \subseteq \calC_a^{\ell-1}$, then $\calC_j^{\ell} \subseteq \calC_a^{\ell-1}$, $\forall j \in \caln[\calC_i^{\ell}]$. The set of cliques $\calC_j^{\ell}$ that include $\calC_i^{\ell}$ as a neighbor clique is defined with $\calm[\calC_i^{\ell}]$, i.e., if $\calC_j^{\ell}$ is such that $i \in \caln[\calC_j^{\ell}]$, then $j \in \calm[\calC_i^{\ell}]$. Note that it is not required that $\caln[\calC_i^{\ell}] \equiv \calm[\calC_i^{\ell}]$.

Next, all necessary communication assumptions are stated. First, we formulate the \textit{basic} communication capabilities of every agent which are only limited in exchanging information with their neighboring agents within the same level-$(L-1)$ clique.

\begin{assumption}[Basic Communication Capabilities]
Every agent $i \in \calR$ (or just $\calC_i^L \in \calR$) is able to exchange information with all agents $j \in \caln[\calC_i^{L}] \cup \calm[\calC_i^{L}]$.    
\end{assumption}

Note that compared to the potential scale of a VLMAS, and assuming that neighbor sets would be relatively small, this is considered to be a minor communication requirement. In the following assumption, we establish how communication between different cliques materializes, by assigning \textit{increased} communication capabilities to specific agents.

\begin{assumption}[Increased Communication Capabilities]
For all levels $\ell \in \llbracket 1, L-1 \rrbracket$, in each clique $\calC_i^\ell$, there exists one agent with ``level-$\ell$ communication capabilities'', or more briefly a ``level-$\ell$ agent''. Each level-$\ell$ agent is able to exchange information with: 
\begin{itemize}
\item all level-$(\ell+1)$ agents of the cliques $\calC_n^{\ell+1}$ such that $\calC_n^{\ell+1} \subseteq \calC_i^\ell$,
\item all level-$\ell$ agents that belong in $\caln[\calC_i^{\ell}] \cup \calm[\calC_i^{\ell}]$.
\end{itemize}
\end{assumption}


\subsection{Virtual States for Clique Dynamics}

Finally, towards associating the control of the cliques of levels $1, \dots, L-1$, with covariance steering, we define their correspoding \textit{virtual} states $x_i^{\ell} \in \Rb^{n_x}$ and controls $u_i^{\ell} \in \Rb^{n_u}$, for all $i \in \calR^{\ell}$. Since the dynamics of all agents are homogeneous, the virtual states are modeled to follow the same dynamics as the agents, 
\begin{equation}
\label{clique linear dynamics}
x_{i,k+1}^{\ell} = A x_{i,k}^{\ell} + B u_{i,k}^{\ell} + w_{i,k}^{\ell},
\end{equation}
where $w_{i,k}^{\ell} \sim \calN(0,W)$. Note that this is still far from associating the control of cliques with covariance steering, since the initial and target random distributions of the cliques of all levels $\ell \in \llbracket 1, L-1 \rrbracket$ are not available and require rigorous selection such that the hierarchical structure is satisfied. In the next section, we propose an approach for estimating these initial and target random distributions as Gaussian distributions $\calN ( \mu_{i,0}^{\ell},\Sigma_{i,0}^{\ell} )$ and $\calN ( \mu_{i,\mathrm{f}}^{\ell},\Sigma_{i,\mathrm{f}}^{\ell} )$, $\forall i \in \calR^{\ell}, \ \ell \in \llbracket 1, L-1 \rrbracket$.

\section{Distributed Hierarchical Distribution Estimation}
\label{sec: estimation}

This section focuses on the problem of finding the optimal Gaussian distributions for representing the initial and target state configurations of the cliques $\calC_i^\ell$, $\forall \calR^\ell$, $\ell \in \llbracket 1,L-1 \rrbracket$, based on the known level-$L$ distributions. We label this problem as the \textit{inter-level distribution estimation} problem. Towards addressing it, we propose Distributed Hierarchical Distribution Estimation (DHDE), a hierarchical distributed approach that operates in a \textit{bottom-up} fashion (Fig. \ref{fig: hierarchical}) for estimating the desired random distributions. 



\subsection{Single-Clique Distribution Estimation}

To facilitate the exposition of our ideas, we first consider the simplified subproblem of \textit{single-clique distribution estimation}, where the objective is to estimate the Gaussian distribution that best describes a clique $\calC_i^{\ell}$, i.e., best captures the random state distributions of all cliques $\calC_n^{\ell+1} \subseteq \calC_i^{\ell}$ with a single Gaussian distribution, without considering potential overlaps in the 2D space between neighboring cliques. Furthermore, note that the distribution estimation problem has the same form, either we refer to the initial or target configurations, thus we make no distinction between the two and drop the corresponding notation. 

In order to find the optimal distribution $\calN_i^{\ell} = \calN ( \mu_i^{\ell},\Sigma_i^{\ell} )$ for capturing all the distributions $\calN_n^{\ell+1} = \calN ( \mu_n^{\ell+1},\Sigma_n^{\ell+1} )$ of the cliques $\calC_n^{\ell+1}$ such that $\calC_n^{\ell+1} \subseteq \calC_i^{\ell}$, we select the KL divergence metric $\KL ( \calN_n^{\ell+1} \| \calN_i^{\ell} )$ to measure discrepancies between distributions. In addition, by defining the parts of the means and covariances that correspond to the position coordinates as $\bar{\mu}_i = H \mu_i$ and $\bar{\Sigma}_i = H \Sigma_i H^\T$, we impose the constraints
\begin{equation}
\calE_\theta [\bar{\mu}_n^{\ell+1} , \bar{\Sigma}_n^{\ell+1}] 
\subseteq
\calE_\theta [\bar{\mu}_i^{\ell} ,  \bar{\Sigma}_i^{\ell}], 
\quad 
n \in \calC_i^{\ell},
\end{equation}
so that the assumed hierarchical clustered structure is indeed satisfied. Therefore, the single-clique distribution estimation problem can be stated for a particular clique $\calC_i^\ell \in \calR^\ell$ of a given level $\ell \in \llbracket 1, L-1 \rrbracket$, as follows.  

\begin{problem}[Single-Clique Distribution Estimation Problem]
\label{estimation problem single clique}
Find the optimal Gaussian distribution $\calN (\mu_i^{\ell},\Sigma_i^{\ell})$ such that 
\begin{subequations}
\begin{align}
& \left\{ \mu_i^{\ell},\Sigma_i^{\ell} \right\} = \argmin J_i^{\mathrm{e}}(\mu_i^{\ell},\Sigma_i^{\ell})
\label{single clique problem: cost}
\\[0.2cm]
\mathrm{s.t.} \quad 
& \calE_\theta [\bar{\mu}_n^{\ell+1} , \bar{\Sigma}_n^{\ell+1}] 
\subseteq
\calE_\theta [\bar{\mu}_i^{\ell} ,  \bar{\Sigma}_i^{\ell}], 
\ 
n \in \calC_i^{\ell},
\label{single clique problem: ellipses constraint}
\\
& \Sigma_i^{\ell} \succ 0,
\label{single clique problem: semidef constraint}
\end{align}
\end{subequations}
where 
\begin{equation}
J_i^{\mathrm{e}} = 
\sum_{n \in \calC_i^{\ell}} \KL ( \calN_n^{\ell+1} \| \calN_i^{\ell} ).
\end{equation}
\end{problem}

In the following proposition, we present a tractable optimization problem whose optimal solution provides the optimal solution of Problem \ref{estimation problem single clique}. From now on, the level superscripts will be omitted unless not obvious from the context. 

\begin{proposition}
\label{single clique proposition}
Let us introduce the auxiliary optimization variables $Q_i = \Sigma_i^{-1}$, $q_i = \Sigma_i^{-1} \mu_i$ and $\tau_n \in \Rb$, $\forall n \in \calC_i^{\ell}$. The optimal solution of Problem \ref{estimation problem single clique} is obtained by solving the following convex optimization problem
\begin{subequations}
\begin{align}
& ~~~~~~~ \min \hat{J}_i^{\mathrm{e}}(Q_i, q_i) 
\label{single clique prop: cost}
\\[0.1cm]
\mathrm{s.t} \quad 
& S_n(\bar{Q}_i, \bar{q}_i, \tau_n) \succeq
0, \quad n \in \calC_i^{\ell},
\label{single clique prop: ellipses constraint 1}
\\
& \tau_n \geq 0, \quad n \in \calC_i^{\ell},
\label{single clique prop: ellipses constraint 2}
\\
& Q_i \succ 0,
\label{single clique prop: semidef constraint}
\end{align}
\end{subequations}
w.r.t. $Q_i$, $q_i$ and $\{\tau_n\}_{n \in \calC_i^{\ell}}$, where 
\begin{subequations}
\begin{align}
& \hat{J}_i^{\mathrm{e}}(Q_i, q_i) = \sum_{n \in \calC_i^{\ell}} - \log|Q_i| + \tr(Q_i \Sigma_n )  
+ q_i^\T Q_i^{-1} q_i
\nonumber
\\
& ~~~~~~~~~~~~~~~~~~~~~~  - 2 \mu_n^\T q_i + \mu_n^\T Q_i \mu_n, 
\\[0.1cm]
& S_n = 
\begin{bmatrix}
S_{11} & S_{12} & 0
\\
S_{12}^\T & S_{22} & S_{23}
\\
0^\T & S_{23}^\T & S_{33}
\end{bmatrix},
\\[0.1cm]
& S_{11} = - \bar{Q}_i + \tau_n \bar{\Sigma}_n^{-1}, \  
S_{12} = \bar{q}_i  - \tau_n \bar{\Sigma}_n^{-1} \bar{\mu}_n,
\label{single clique prop: S terms}
\\
& S_{22} = \alpha + \tau_n \bar{\mu}_n^\T \bar{\Sigma}_n^{-1} \bar{\mu}_n - \alpha \tau_n, \ 
S_{23} = \bar{q}_i^\T, \ S_{33} = \bar{Q}_i, 
\nonumber
\\
& \bar{Q}_i = H Q_i H^\T, \ \bar{q}_i = H q_i, \ \alpha = f_{\chi^2,2}^{-1} (\theta),
\end{align}
\end{subequations}
and setting $\mu_i = Q_i^{-1} q_i$ and $\Sigma_i = Q_i^{-1}$.
\end{proposition}
\begin{proof}
The proof is provided in Section \ref{SM sec: proposition 1 proof} of the Supplementary Material (SM).
\end{proof}

\subsection{Multi-Clique Distribution Estimation}
\label{subsec: multi clique estimation}

Let us now consider an extended version of the previous problem, labeled as the \textit{multi-clique distribution estimation} one, where the objective is to simultaneously estimate the Gaussian distributions $\calN_i^{\ell}$ of all cliques $\calC_i^{\ell}$ such that $\calC_i^{\ell} \subseteq \calC_a^{\ell-1}$, i.e., all cliques that belong in the same parent clique $\calC_a^{\ell-1}$ (or just $\calR^1$ if $\ell = 1$). In this case, it is also necessary to ensure that the $\theta$-probability confidence ellipses $\calE_\theta [\bar{\mu}_i^{\ell} , \bar{\Sigma}_i^{\ell}]$ of neighboring cliques will not overlap with each other, i.e., we also impose the following constraints
\begin{equation}
\calE_\theta [\bar{\mu}_i^{\ell} , \bar{\Sigma}_i^{\ell}] \cap \calE_\theta [ \bar{\mu}_j^{\ell} , \bar{\Sigma}_j^{\ell}] = \emptyset, 
\  j \in \caln[\calC_i^\ell],
\end{equation}
between neighbor cliques of the same parent clique.
Hence, the multi-clique distribution estimation problem can be formulated as follows. 

\begin{problem}[Multi-Clique Distribution Estimation Problem]
\label{estimation problem multi clique}
Given a parent clique $\calC_a^{\ell-1}$ (or $\calR^1$ if $\ell = 1$), find the optimal Gaussian distributions $\calN \left( \mu_i^{\ell},\Sigma_i^{\ell} \right)$ for all $i \in \calC_a^{\ell-1}$, such that 
\begin{subequations}
\begin{align}
& \big\{ \mu_i^{\ell}, \Sigma_i^{\ell} \big\}_{i \in \calC_a^{\ell-1}} = \argmin \sum_{i \in \calC_a^{\ell-1}} 
J_i^{\mathrm{e}}(\mu_i^{\ell},\Sigma_i^{\ell})
\label{multi clique est problem: cost}
\\[0.1cm]
\mathrm{s.t.} \quad
& \calE_\theta [\bar{\mu}_n^{\ell+1} ,  \bar{\Sigma}_n^{\ell+1}] 
\subseteq
\calE_\theta [\bar{\mu}_i^{\ell} ,  \bar{\Sigma}_i^{\ell}], \ 
n \in \calC_i^{\ell}, 
\label{multi clique est problem: ellipses constraint}
\\[0.1cm]
& \calE_\theta [\bar{\mu}_i^{\ell} , \bar{\Sigma}_i^{\ell}] \cap \calE_\theta [ \bar{\mu}_j^{\ell} , \bar{\Sigma}_j^{\ell}] = \emptyset, 
\  j \in \caln[\calC_i^\ell],
\label{multi clique est problem: ellipses overlap constraint}
\\[0.1cm] 
& \Sigma_i^{\ell} \succ 0, \ i \in \calC_a^{\ell-1}, 
\label{multi clique est problem: semidef constraint}
\end{align}
\end{subequations}
where $J_i^{\mathrm{e}}(\mu_i^{\ell},\Sigma_i^{\ell})$ is the same as in Problem \ref{estimation problem single clique}.
\end{problem}

In the following proposition, we present an optimization problem whose optimal solution provides a suboptimal solution for Problem \ref{estimation problem multi clique}. 

\begin{proposition}
\label{multi clique proposition}
Let us introduce the auxiliary optimization variables $Q_i = \Sigma_i^{-1}$, $q_i = \Sigma_i^{-1} \mu_i$, $\phi_i \in \Rb$ and $\tau_{i,n} \in \Rb$, $\forall n \in \calC_i^{\ell}, \ \forall i \in \calC_a^{\ell-1}$. A suboptimal solution for Problem 3 is obtained by solving the following optimization problem
\begin{subequations}
\label{multi clique problem align}
\begin{align}
& ~~~~~~
\min \sum_{i \in \calC_a^{\ell-1}} \hat{J}_i^{\mathrm{e}}(Q_i, q_i)
\label{multi clique prop: cost}
\\[0.2cm]
\mathrm{s.t} \quad 
& S_{i,n}(\bar{Q}_i, \bar{q}_i, \tau_{i,n}) \succeq
0, \ n \in \calC_i^{\ell},
\label{multi clique prop: ellipses constraint 1}
\\
& \tau_{i,n} \geq 0, \ n \in \calC_i^{\ell},
\label{multi clique prop: ellipses constraint 2}
\\
& h_{i,j}(\bar{Q}_i, \bar{q}_i, \phi_i, \bar{Q}_j, \bar{q}_j, \phi_j) \leq 0,
\  j \in \caln[\calC_i^\ell],
\label{multi clique prop: ellipses overlap constraint 1}
\\
& T_i(\bar{Q}_i, \phi_i) \succeq 0,
\label{multi clique prop: ellipses overlap constraint 2}
\\ 
& Q_i \succ 0, \ i \in \calC_a^{\ell-1},
\label{multi clique prop: semidef constraint}
\end{align}
\end{subequations}
w.r.t. $Q_i$, $q_i$, $\{\tau_n\}_{n \in \calC_i^{\ell}}$ and $\phi_i$, $\forall i \in \calC_a^{\ell-1}$, where
\begin{subequations}
\begin{align}
h_{i,j} & = 
r_i(\phi_i) + r_j(\phi_j)
- \left\| \bar{Q}_i^{-1} \bar{q}_i - \bar{Q}_j^{-1} \bar{q}_j \right\|_2, 
\\
r_i & = \phi_i^{-1/2}, \ T_i = \bar{Q}_i - \phi_i \alpha I, 
\end{align}
\end{subequations}
$\hat{J}_i^{\mathrm{e}}$, $S_{i,n}$, $\bar{Q}_i$ and $\bar{q}_i$ are as in Proposition \ref{single clique proposition},
and setting $\mu_i = Q_i^{-1} q_i$ and $\Sigma_i = Q_i^{-1}$ for every $i \in \calC_a^{\ell-1}$.
\end{proposition}
\begin{proof}
Provided in Section \ref{SM sec: proposition 2 proof} of the SM.
\end{proof}

\begin{remark}
The optimal solution of the problem in Proposition \ref{multi clique proposition} might be suboptimal - but always feasible - for Problem \ref{estimation problem multi clique}, in the sense that the constraints \eqref{multi clique prop: ellipses overlap constraint 1} and \eqref{multi clique prop: ellipses constraint 2} formulate a ``tighter'' version of \eqref{multi clique est problem: ellipses overlap constraint}, as shown in Section \ref{SM sec: proposition 2 proof}.
\end{remark}

Note that the cost and all constraints in \eqref{multi clique problem align} are convex except for \eqref{multi clique prop: ellipses overlap constraint 1}. As explained in the next section, we address that through iterative linearization of the constraint.

\subsection{Distributed Hierarchical Distribution Estimation}

Let us now formulate the full \textit{inter-level distribution estimation} problem, whose objective is to estimate the optimal distributions $\calN ( \mu_i^{\ell},\Sigma_i^{\ell} )$ of all cliques $\calC_i^\ell \in \calR^\ell$ of all levels $\ell \in \llbracket 1, L-1 \rrbracket$. Of course, this problem will consist of many interconnected instances of Problem \ref{estimation problem multi clique}. In fact, the inter-level distribution estimation problem can be formulated as follows.

\begin{problem}[Inter-Level Distribution Estimation Problem]
\label{problem: estimation full interlevel}
For all $\ell \in \llbracket 1, L-1 \rrbracket$ 
and for all $a \in \calR^{\ell-1}$ (if $\ell \geq 2)$
, find the sets of optimal Gaussian distributions $\{ \calN \left( \mu_i^{\ell},\Sigma_i^{\ell} \right) \}, \ i \in \calC_a^{\ell-1}$ , that solve each corresponding Problem \ref{estimation problem multi clique}.
\end{problem}

The goal of the proposed Distributed Hierarchical Distribution Estimation (DHDE) framework is to solve Problem \ref{problem: estimation full interlevel} using a bottom-up strategy, since the only a priori known distributions are the level-$L$ ones. In this direction, we propose first solving all instances of Problem \ref{estimation problem multi clique} in Level $L-1$, then based on the acquired information, i.e. all $\calN_i^{L-1}$, solve all instances of Problem \ref{estimation problem multi clique} in Level $L-2$, to obtain all $\calN_i^{L-2}$, and so on. Of course, based on the results of Section \ref{subsec: multi clique estimation}, we also replace all these instances of Problem \ref{estimation problem multi clique} with the Proposition \ref{multi clique proposition} problems. Finally, to further distribute computations, we propose solving each such problem in a distributed manner using an approach based on the Alternating Direction Method of Multipliers (ADMM) \cite{boyd2011distributed}.

\begin{algorithm}[t]
\caption{DHDE Algorithm} \label{DHDE Algorithm}
\begin{algorithmic}[1] 
\State \textbf{Inputs:} \textsc{Hier\_Structure}, $\mu_i^L, \Sigma_i^L, \forall i \in \calR$, $\theta$
\For{$ \ell = L-1:-1:1$}
\If{$\ell > 1$}
\State $\calA \leftarrow \calR^{\ell - 1}$
\Else
\State $\calA \leftarrow \{1\}$
\EndIf
\For{$a \in  \calA$} (in parallel)
\For{$i \in \calC_a^{\ell-1} $} (in parallel)
\State \textit{All} $n \in \calC_i^{\ell}$ \textit{send} $\mu_n^{\ell+1}, \Sigma_n^{\ell+1}$ \textit{to} $i$.
\EndFor
\While{\textsc{Term\_Criterion} $==$ False}
\For{$i \in \calC_a^{\ell-1} $} (in parallel)
\State $\{ \tilde{Q}_i, \tilde{q}_i, \tilde{\phi}_i \} \leftarrow$ Solve \eqref{estimation admm local updates}. 
\State \textit{All}  $j \in \calm[\calC_i^\ell]$ \textit{send} $\{ Q_i^j, q_i^j, \phi_i^j \}$ \textit{to} $i$.
\EndFor
\For{$i \in \calC_a^{\ell-1} $} (in parallel)
\State $\{ G_i, g_i, z_i \} \leftarrow$ Update with \eqref{estimation admm global updates}. 
\State \textit{All} $j \in \caln[\calC_i^\ell]$ \textit{send} $\{ G_j, g_j, z_j \}$ \textit{to} $i$. 
\EndFor
\For{$i \in \calC_a^{\ell-1} $} (in parallel)
\State $\{ \Xi_i, \xi_i, y_i \} \leftarrow$ Update with \eqref{estimation admm dual updates}. 
\EndFor
\EndWhile
\For{$i \in \calC_a^{\ell-1} $} (in parallel)
\State $\{ \mu_i^\ell, \Sigma_i^\ell \} \leftarrow$ Compute based on $Q_i, q_i$.
\EndFor
\EndFor
\EndFor
\end{algorithmic}
\end{algorithm}

In order to solve problem \eqref{multi clique problem align} in a distributed fashion, we first need to address the coupling induced by the constraints \eqref{multi clique prop: ellipses overlap constraint 1} between neighboring cliques. Hence, we define the copy variables $Q_j^i$, $q_j^i$ and $\phi_j^i$, for all $j \in \caln[\calC_i^\ell]$, and subsequently the augmented variables 
\begin{subequations}
\begin{align}
\tilde{Q}_i & = \big[ Q_i; \{Q_j^i\}_{j \in \caln[\calC_i^\ell]} \big], \\ 
\tilde{q}_i & = \big[ q_i; \{q_j^i\}_{j \in \caln[\calC_i^\ell]} \big], \\ 
\tilde{\phi}_i & = \big[ \phi_i; \{\phi_j^i\}_{j \in \caln[\calC_i^\ell]} \big].
\end{align}
\end{subequations}
Note that the addition of the copy variables might have created multiple variables for each agent. To accommodate for that, we also define the global variables 
$G = [ \{G_i\}_{i \in \calC_a^{\ell-1}} ]$, 
$g = [ \{g_i\}_{i \in \calC_a^{\ell-1}} ]$, 
$z = [ \{z_i\}_{i \in \calC_a^{\ell-1}} ]$, 
and impose the consensus constraints 
\begin{equation}
\tilde{Q}_i = \tilde{G}_i, \ 
\tilde{q}_i = \tilde{g}_i, \ 
\tilde{\phi}_i = \tilde{z}_i,
\end{equation}
where 
$\tilde{G}_i = [ G_i; \{G_j\}_{j \in \caln[\calC_i^\ell]} ]$, 
$\tilde{g}_i = [ g_i; \{g_j\}_{j \in \caln[\calC_i^\ell]} ]$ and 
$\tilde{z}_i = [ z_i; \{z_j\}_{j \in \caln[\calC_i^\ell]} ]$.

Next, we present the algorithm updates; for a full derivation, the reader is referred to Section \ref{SM sec: DHDE admm} of the SM. First, each level-$\ell$ agent $i$ updates its local variables $\tilde{Q}_i$, $\tilde{q}_i$ and $\tilde{\phi}_i$ by solving the optimization problem 
\begin{subequations}
\label{estimation admm local updates}
\begin{align}
& 
\{\tilde{Q}_i, \tilde{q}_i, \tilde{\phi}_i\} = \argmin \tilde{J}_i^{\mathrm{e}}(\tilde{Q}_i, \tilde{q}_i, \tilde{\phi}_i)
\label{DHDE local problem: cost}
\\[0.1cm]
\mathrm{s.t} \quad 
& S_{i,n}(Q_i, q_i, \tau_{i,n}) \succeq
0, \ n \in \calC_i^{\ell},
\label{DHDE local problem: ellipses constraint 1}
\\
& \tau_{i,n} \geq 0, \ n \in \calC_i^{\ell},
\label{DHDE local problem: ellipses constraint 2}
\\
& \bar{h}_{i,j}(Q_i, q_i, \phi_i, Q_j^i, q_j^i, \phi_j^i) \leq 0,
\  j \in \caln[\calC_i^\ell],
\label{DHDE local problem: ellipses overlap constraint 1}
\\
& T_i(Q_i, \phi_i) \succeq 0,
\label{DHDE local problem: ellipses overlap constraint 2}
\\
& Q_i \succ 0
\end{align}
\end{subequations}
where the augmented costs $\tilde{J}_i^{\mathrm{e}}(\tilde{Q}_i, \tilde{q}_i, \tilde{\phi}_i)$ are given by
\begin{align}
\tilde{J}_i^{\mathrm{e}} & = \hat{J}_i^{\mathrm{e}}(Q_i, q_i) + \tr(\Xi_i^\T(\tilde{Q}_i - \tilde{G}_i)) + \xi_i^\T(\tilde{q}_i - \tilde{g}_i) 
\nonumber
\\[0.1cm]
& ~~ + y_i^\T(\tilde{\phi}_i - \tilde{z}_i)
+ \frac{\rho_Q}{2} \| \tilde{Q}_i - \tilde{G}_i \|_F^2
+ \frac{\rho_q}{2} \| \tilde{q}_i - \tilde{g}_i \|_2^2
\nonumber
\\
& ~~ + \frac{\rho_\phi}{2} \| \tilde{\phi}_i - \tilde{z}_i \|_2^2,
\end{align}
with $\rho_Q, \rho_q, \rho_\phi > 0$ being penalty parameters and $\Xi_i, \xi_i, y_i$ being the dual variables of the corresponding constraints. In addition, we take advantage of the iterative nature of ADMM and replace $h_{i,j}(\cdot)$ in \eqref{DHDE local problem: ellipses overlap constraint 1} with its linear approximation $\bar{h}_{i,j}(\cdot)$ to convexify problem \eqref{estimation admm local updates}. For more details, the reader is referred to Section \ref{SM sec: DHDE details} of the SM.

Subsequently, the components of the global variables are updated locally by each level-$\ell$ agent as follows, 
\begin{equation}
\label{estimation admm global updates}
G_i \leftarrow \frac{1}{|\calm'[\calC_i^\ell]|} \sum_{j \in \calm'[\calC_i^\ell]} Q_i^j
\end{equation}
where $\calm'[\calC_i^\ell] = \calm[\calC_i^\ell] \cup \{ i \}$. The updates for $g_i$ and $z_i$ have the same form as \eqref{estimation admm global updates} if we replace $Q_i^j$ with $q_i^j$ and $\phi_i^j$, respectively. Finally, the dual variables are updated with
\begin{subequations}
\begin{align}
\Xi_i & \leftarrow \Xi_i + \rho_Q (\tilde{Q}_i - \tilde{G}_i)
\\ 
\xi_i & \leftarrow \xi_i + \rho_q (\tilde{q}_i - \tilde{g}_i)
\\
y_i & \leftarrow y_i + \rho_{\phi} (\tilde{\phi}_i - \tilde{z}_i).
\end{align}
\label{estimation admm dual updates}%
\end{subequations}
Note that all updates in \eqref{estimation admm local updates}, \eqref{estimation admm global updates}, \eqref{estimation admm dual updates} can be performed in parallel by each level-$\ell$ agent $i$. Therefore, all computations take place in a decentralized manner. 

The DHDE algorithm with all necessary computation and communication steps is illustrated in Alg. \ref{DHDE Algorithm}. The method operates in a bottom-up fashion for $\ell = \{ L-1,L-2,\dots,1 \}$. For a particular level $\ell$, the first step is that all agents $n$ such that $\calC_n^{\ell+1} \subseteq \calC_i^\ell$ send $\mu_n^{\ell+1}, \Sigma_n^{\ell+1}$ to the level-$\ell$ agent $i$ that corresponds to the clique $\calC_i^\ell$ (Line 9). Then the iterative ADMM procedure starts for every different group of agents that corresponds to a clique $\calC_a^{\ell-1}$ (Lines 10-18). Note that these procedures can of course take place in parallel. Focusing into a particular group of agents that belong in clique $\calC_a^{\ell-1}$, the ADMM updates are performed as follows. First, each agent $i \in \calC_a^{\ell-1}$ solves \eqref{estimation admm local updates} to update its local variables $\{ \tilde{Q}_i$, $\tilde{q}_i, \tilde{\phi}_i \}$ (Line 12). Subsequently, each $i$ receives the copy variables $\{ Q_i^j, q_i^j, \phi_i^j \}$ from all $j \in \calm[\calC_i^\ell]$ (Line 13). As a result, each agent $i$ can now obtain the new iterates for $\{G_i, g_i, z_i\}$ (Line 15) through updates of the form \eqref{estimation admm global updates}. Afterwards, all agents $j \in \caln[\calC_i^\ell]$ send the variables $\{ G_j, g_j, z_j \}$ to each $i$ (Line 16) so that each agent $i$ updates its dual variables $\{\Xi_i, \xi_i, y_i\}$ (Line 18) with \eqref{estimation admm dual updates}. Once a predefined termination criterion is satisfied, each agent $i$ computes the variables $ \mu_i^\ell$ and $\Sigma_i^\ell$ (Line 20). Now that the estimates $ \mu_i^\ell, \Sigma_i^\ell$ have been found for all $\ell \in \calR^\ell$, this procedure repeats for the above level $\ell-1$, and so on, until level $1$ is reached.

\section{Distributed Hierarchical Distribution Steering}
\label{sec: steering}

After associating all cliques of all levels with their initial and target Gaussian distributions, we proceed with addressing the problem of steering all state distributions from the initial to the target ones. We label this problem as the \textit{inter-level distribution steering} one. To solve this problem, we propose a \textit{top-down} hierarchical distributed method (Fig. \ref{fig: hierarchical}) called Distributed Hierarchical Distribution Steering (DHDS).

\subsection{Multi-Clique Distribution Steering}

Before formulating the inter-level problem, let us again first state a subproblem that serves as the basic component of the full problem. In particular, we consider the \textit{multi-clique distribution steering} problem whose objective is to steer the Gaussian distributions of (the virtual states of) all $i$ such that the cliques $\calC_i^{\ell} \subseteq \calC_a^{\ell-1}$, for a specific parent clique $\calC_a^{\ell-1}$ and level $\ell$. Of course, in the case where $\ell = 1$, we consider $\calR^1$ as the parent clique. Thus, the multi-clique distribution steering problem can be formulated as follows.

\begin{problem}[Multi-Clique Distribution Steering Problem]
\label{steering problem multi clique}
Given a parent clique $\calC_a^{\ell-1}$ (or $\calR^1$ if $\ell = 1$), find the optimal control sequences $u_i^{\ell}$ for all $i \in \calC_a^{\ell-1}$, such that 
\begin{subequations}
\begin{align}
& \big\{ u_i^{\ell} \big\}_{i \in \calC_a^{\ell-1}} = \argmin \sum_{i \in \calC_a^{\ell-1}} 
J_i^{\mathrm{s}}(u_i^{\ell})
\label{multi clique steering problem: cost}
\\[0.1cm]
\mathrm{s.t.} \quad
& x_{i,k+1}^{\ell} = A x_{i,k}^{\ell} + B u_{i,k}^{\ell} + w_{i,k}^{\ell},
\label{multi clique steering problem: dynamics}
\\[0.1cm]
& \Eb[x_{i,N}^{\ell}] = \mu_{i,\mathrm{f}}^{\ell}, ~ 
\Cov[x_{i,N}^{\ell}] \preceq \Sigma_{i,\mathrm{f}}^{\ell},
\\[0.1cm]
& q_{i,j,k}^{\ell}(p_{i,k}^{\ell}, p_{j,k}^{\ell}) \geq 0,
~  j \in \caln[\calC_i^\ell], 
\label{multi clique steering problem: inter-agent constraint}
\\[0.1cm]
& s_{i,k}^{\ell}(p_{i,k}^{\ell}) \geq 0,
\label{multi clique steering problem: obs constraint}
\\[0.1cm] 
& \calE_\theta [\bar{\mu}_{i,k}^{\ell} , \bar{\Sigma}_{i,k}^{\ell}] 
\subseteq
\calE_\theta [\bar{\mu}_{a,k}^{\ell-1} ,  \bar{\Sigma}_{a,k}^{\ell-1}], 
\label{multi clique steering problem: in ellipse constraint}
\\
& k \in \llbracket 0, N \rrbracket, ~ 
i \in \calC_a^{\ell-1},
\nonumber
\end{align}
\end{subequations}
where
\begin{subequations}
\begin{align}
& J_i^{\mathrm{s}}(u_i^{\ell}) = \sum_{k = 0}^{N-1} \Eb[u_{i,k}^{\ell \ \mathrm{T}} R u_{i,k}^{\ell}],
\\
& q_{i,j,k}^{\ell}(p_{i,k}^{\ell}, p_{j,k}^{\ell}) = 
\Pb \left( 
\| p_{i,k}^{\ell} - p_{j,k}^{\ell} \|_2 \geq d_{\mathrm{inter}}^{\ell}
\right) - \theta,
\\[0.1cm]
& s_{i,k}^{\ell}(p_{i,k}^{\ell}) = \Pb \left( 
\| p_{i,k}^{\ell} - p_o \|_2 \geq d_{\mathrm{obs}}^{\ell} + r_o
\right) - \theta.
\end{align}
\end{subequations}
and $d_{\mathrm{inter}}^{\ell}$, $d_{\mathrm{obs}}^{\ell}$ are prespecified parameters.
\end{problem}

In the following, we will be omitting level superscripts, unless not clear by the context. 

\subsection{Problem Transformation}

In order to address Problem \ref{steering problem multi clique}, we consider the affine disturbance feedback control policies proposed in \cite{balci2021covariance}, with
\begin{equation}
u_{i,k} = \bar{u}_{i,k} + L_{i,k} (x_{i,0} - \mu_{i,0}) +  \sum^{k-1}_{\kappa=0} K_{i,(k-1,\kappa)} w_{i,\kappa}, 
\end{equation}
where $\bar{u}_{i,k} \in \Rb^{n_u}$ are the feed-forward control terms and $L_{i,k}, \ K_{i,(k-1,\kappa)} \in \Rb^{n_u \times n_x}$ are feedback gain matrices. After concentrating these variables for the entire time horizon, we obtain the decision variables $\bar{u}_i \in \Rb^{N n_u}$, $L_i \in \Rb^{N n_u \times n_x}$ and $K_i \in \Rb^{N n_u \times N n_x}$, with their exact forms provided in Section \ref{SM sec: DHDS detailed expressions} of the SM. Thus, the control sequences $u_i$ can be rewritten as
\begin{equation}
\label{control policy}
u_i = \bar{u}_i + L_i (x_{i,0} - \mu_{i,0}) + K_i w_i.
\end{equation}
If we also write the dynamics \eqref{clique linear dynamics} in their concatenated form, 
\begin{equation}
x_i = \Psi_0 x_{i,0} + \Psi_u u_i + \Psi_w w_i, 
\label{state compact}
\end{equation}
where the matrices $\Psi_0$, $\Psi_u$ and $\Psi_w$ are provided in the SM, then through \eqref{control policy}, we obtain 
\begin{align}
x_i & = \Psi_0 x_{i,0} + \Psi_u \bar{u}_i + \Psi_u L_i (x_{i,0} - \mu_{i,0})
\nonumber
\\
& ~~~~ + (\Psi_{w} + \Psi_{u} K_i) w_i.
\end{align}
Therefore, given that affine transformations preserve Gaussianity, the entire state sequence $x_i$ is a Gaussian vector, 
which implies that
$x_{i,k} \sim \calN(\mu_{i,k}, \Sigma_{i,k})$, with  
%
%
\begin{equation}
\mu_{i,k} = f_{i,k}(\bar{u}_i),
\quad 
\Sigma_{i,k} = F_{i,k}(L_i,K_i),
\label{state mean covs}
\end{equation}
where the exact expressions for $f_{i,k}(\bar{u}_i)$ and $F_{i,k}(L_i,K_i)$ are provided in the SM. Note that the mean states only depend on the feed-forward controls, while the state covariances only depend on the feedback matrices. This is a useful fact that we will later exploit to further increase computational efficiency.

In the following proposition, we present an new optimization problem whose optimal solution provides a suboptimal one for Problem \ref{steering problem multi clique}.

\begin{proposition}
\label{proposition: multi clique steering}
A suboptimal solution for Problem \ref{steering problem multi clique} is obtained by solving the following optimization problem.
In particular, given a parent clique $\calC_a^{\ell-1}$ (or $\calR^1$ if $\ell = 1$), find the optimal decision variables $\bar{u}_i, L_i, K_i$ for all $i \in \calC_a^{\ell-1}$, such that 
\begin{subequations}
\label{proposition: multi clique steering align}
\begin{align}
\big\{ \bar{u}_i, & L_i, K_i \big\}_{i \in \calC_a^{\ell-1}} = \argmin \sum_{i \in \calC_a^{\ell-1}} 
\hat{J}_i^{\mathrm{s}}(\bar{u}_i, L_i, K_i)
\label{prop: multi clique steering problem: cost}
\\[0.1cm]
\mathrm{s.t.} \quad
& \calf_{i,N}(\bar{u}_i) = 0, ~ 
\calF_{i,N}(L_i, K_i) \succeq 0, 
\\[0.1cm] 
& \calQ_{i,k}(L_i, K_i) \succeq 0, ~ 
\cals_{i,k}(\bar{u}_i) \geq 0, 
\label{prop: multi clique steering problem: single obs constraint}
\\[0.1cm] 
& \calq_{i,j,k}(\bar{u}_i, \bar{u}_j) \geq 0,
~  j \in \caln[\calC_i^\ell],
\label{prop: multi clique steering problem: interagent coupling constraint}
\\[0.1cm] 
& \calp_{i,k}(\bar{u}_i) \leq 0, ~ k \in \llbracket 0, N \rrbracket, ~  
i \in \calC_a^{\ell-1},
\end{align}
\end{subequations}
with
\begin{subequations}
\begin{align}
& \hat{J}_i^{\mathrm{s}} = \bar{u}_i^\T \bar{R} \bar{u}_i + \tr(\bar{R} K_i W K_i^\T+ \bar{R} L_i \Sigma_{i,0} L_i^\T),
\\[0.1cm] 
& \calf_{i,N}(\bar{u}_i) = \mu_{i,\mathrm{f}} - f_{i,N}(\bar{u}_i), 
\\[0.1cm] 
& \calF_{i,N}(L_i, K_i) = 
\begin{bmatrix}
\Sigma_{i,\mathrm{f}} & \Phi_{i,N}(L_i, K_i) \\ 
\Phi_{i,N}(L_i, K_i)^\T & I
\end{bmatrix},
\\[0.1cm] 
& \calQ_{i,k}(L_i, K_i) = 
\begin{bmatrix}
\big(\frac{r^\ell}{\sqrt{\alpha}}\big)^2 I & H \Phi_{i,k}(L_i, K_i) \\ 
\Phi_{i,k}(L_i, K_i)^\T H^\T & I
\end{bmatrix}, 
\\[0.1cm] 
& \cals_{i,k} (\bar{u}_i, \bar{u}_j) = 
\| \bar{f}_{i,k}(\bar{u}_i) - p_o \|_2 - r^{\ell} - r_o - d_{\mathrm{obs}}^{\ell}, 
\\[0.1cm] 
& \calp(\bar{u}_i) = \| \bar{f}_{i,k}(\bar{u}_i) - \bar{\mu}_{a,k}^{\ell-1} \|_{\hat{P}} - 1 \leq 0,
\\[0.1cm] 
& \calq_{i,j,k} (\bar{u}_i, \bar{u}_j) = 
\| \bar{f}_{i,k}(\bar{u}_i) - \bar{f}_{i,k}(\bar{u}_j) \|_2 - 2 r^{\ell} - d_{\mathrm{inter}}^{\ell}, 
\end{align}
\end{subequations}
where $\bar{R} = \mathrm{blkdiag(R, \dots, R)} \in \Sb_n^{++}$ $\bar{f}_{i,k} = H f_{i,k}$, $r^{\ell}$ is a prespecified parameter and $\Phi_{i,k}(L_i, K_i)$ is an affine function of $L_i, K_i$, 
provided in Section \ref{SM sec: proposition 3 proof} of the SM.
The matrix $\hat{P} = \frac{1}{\alpha} U \hat{\Lambda}^{-1} U^\T$, where
\begin{equation}
\hat{\Lambda} 
= \left( \Lambda^{1/2} - \frac{r}{\sqrt{\alpha}} I  \right)^2,
\end{equation}
with $[\Lambda,U]$ being the eigendecomposition of the matrix $\bar{\Sigma}_{a,k}^{\ell-1}$.
\end{proposition}
\begin{proof}
Provided in Section \ref{SM sec: proposition 3 proof} of the SM.
\end{proof}

\begin{remark}
The optimal solution of the problem in Proposition \ref{proposition: multi clique steering} might be suboptimal - but always feasible - for Problem \ref{steering problem multi clique}, since the constraints \eqref{prop: multi clique steering problem: single obs constraint}-\eqref{prop: multi clique steering problem: interagent coupling constraint} provide a ``tighter'' version of \eqref{multi clique steering problem: inter-agent constraint}-\eqref{multi clique steering problem: obs constraint}, as shown in Section \ref{SM sec: proposition 3 proof}.
\end{remark}

\begin{remark}
\label{remark: mean and cov separable}
A major advantage of the problem presented in Proposition \ref{proposition: multi clique steering}, is that it is separable w.r.t. the feed-forward control variables $\bar{u}_i$ and the feedback matrices $L_i, K_i$. In addition, an inter-agent coupling only appears through the variables $\bar{u}_i$ - because of \eqref{prop: multi clique steering problem: interagent coupling constraint} - and not through $L_i, K_i$. As shown in the next section, these two facts can be exploited to significantly increase computational efficiency.
\end{remark}

\subsection{Distributed Hierarchical Distribution Steering}

Let us now introduce the full \textit{inter-level distribution steering} problem, where the objective lies in finding the optimal control sequences that will steer the state distributions of all cliques $\calC_i^\ell \in \calR^\ell$ of all levels $\ell \in \llbracket 1, L \rrbracket$ from the initial distributions $\calN ( \mu_{i,0}^{\ell},\Sigma_{i,0}^{\ell} )$ to the target ones $\calN ( \mu_{i,\mathrm{f}}^{\ell},\Sigma_{i,\mathrm{f}}^{\ell})$. Being in symmetry with its estimation counterpart, this problem will consist of many interconnected instances of Problem \ref{steering problem multi clique}. Indeed, the inter-level distribution steering problem is stated as follows.

\begin{problem}[Inter-Level Distribution Steering Problem]
\label{problem: steering full interlevel}
For all $\ell \in \llbracket 1, L \rrbracket$ and for all $a \in \calR^{\ell-1}$ (where by convention $\calR^0 = \{1\}$), find the optimal control policies $\{ u_i \}, i \in \calC_a^{\ell-1}$, that solve each corresponding Problem \ref{steering problem multi clique}.
\end{problem}
\begin{remark}
The combination of Problems \ref{problem: estimation full interlevel} and \ref{problem: steering full interlevel} yields a relaxed form of Problem \ref{VLMAS original problem}, that is substantially more computationally tractable as the VLMAS dimensionality increases.
\end{remark}

\begin{algorithm}[t]
\caption{DHDS Algorithm}\label{DHDS Algorithm}
\begin{algorithmic}[1] 
\State \textbf{Inputs:} \textsc{Hier\_Structure}, $A, B, W$, $\mu_{i,0}^\ell, \Sigma_{i,0}^\ell, \mu_{i,\mathrm{f}}^\ell, \Sigma_{i,\mathrm{f}}^\ell$ $\forall i \in \calR^\ell$, $d_{\mathrm{obs}}^{\ell}$, $d_{\mathrm{inter}}^{\ell}$, $r^\ell$, $ \forall \ell \in \llbracket 1, L\rrbracket$, $\theta$ 
\For{$ \ell = 1:L$}
\If{$\ell > 1$}
\State $\calA \leftarrow \calR^{\ell - 1}$
\Else
\State $\calA \leftarrow \{1\}$
\EndIf
\For{$a \in  \calA$} (in parallel)
\If{$\ell > 1$}
\State \textit{Each} $a \in \calR^{\ell-1}$ \textit{sends} $\mu_{a,k}^{\ell-1}, \Sigma_{a,k}^{\ell-1}, \ k \in \llbracket 0,N \rrbracket$ $~~~~~~~~~~~~{\color{white}.}$ \textit{to all} $i \in \calC_a^{\ell-1}$.
\EndIf
\While{\textsc{Term\_Criterion} $==$ False}
\For{$i \in \calC_a^{\ell-1} $} (in parallel)
\State $\tilde{u}_i \leftarrow$ Solve \eqref{DHDS ubar: local update}. 
\State \textit{All}  $j \in \calm[\calC_i^\ell]$ \textit{send} $\bar{u}_j^i$ \textit{to each} $i$.
\EndFor
\For{$i \in \calC_a^{\ell-1} $} (in parallel)
\State $b_i \leftarrow$ Update with \eqref{DHDS ubar: global update}. 
\State \textit{All} $j \in \caln[\calC_i^\ell]$ \textit{send} $b_j$ \textit{to each} $i$. 
\EndFor
\For{$i \in \calC_a^{\ell-1} $} (in parallel)
\State $v_i \leftarrow$ Update with \eqref{DHDS ubar: dual update}. 
\EndFor
\EndWhile
\For{$i \in \calC_a^{\ell-1} $} (in parallel)
\State $\{ L_i, K_i \} \leftarrow$ Solve \eqref{DHDS: feedback problem 2}.
\State $\mu_{i,k}^{\ell}, \Sigma_{i,k}^{\ell} \leftarrow$ Get with
\eqref{state mean covs} for all $k \in \llbracket 0,N \rrbracket$.
\EndFor
\EndFor
\EndFor
\end{algorithmic}
\end{algorithm}

In contrast with DHDE, the aim of DHDS is to solve Problem \ref{problem: steering full interlevel} with a top-down procedure, since every level-$\ell$ subproblem depends on a level-$(\ell-1)$ subproblem. Therefore, we intend to first solve all instances of Problem \ref{steering problem multi clique} in Level $1$, then based on the acquired distributions $\calN(\mu_{i,k}^1, \Sigma_{i,k}^1)$, solve all instances of Problem \ref{steering problem multi clique} in Level $2$, and so on, until Level $L$ is reached. After replacing all instances of Problem \ref{steering problem multi clique} with the problem proposed in Proposition \ref{proposition: multi clique steering}, we propose again an ADMM-based approach to solve each such problem in a distributed manner.

\begin{figure*}
     \centering
          \begin{subfigure}[b]{0.24\textwidth}
         \centering
         \includegraphics[width=\textwidth, trim={0cm 0cm 0cm 0cm
         },clip]{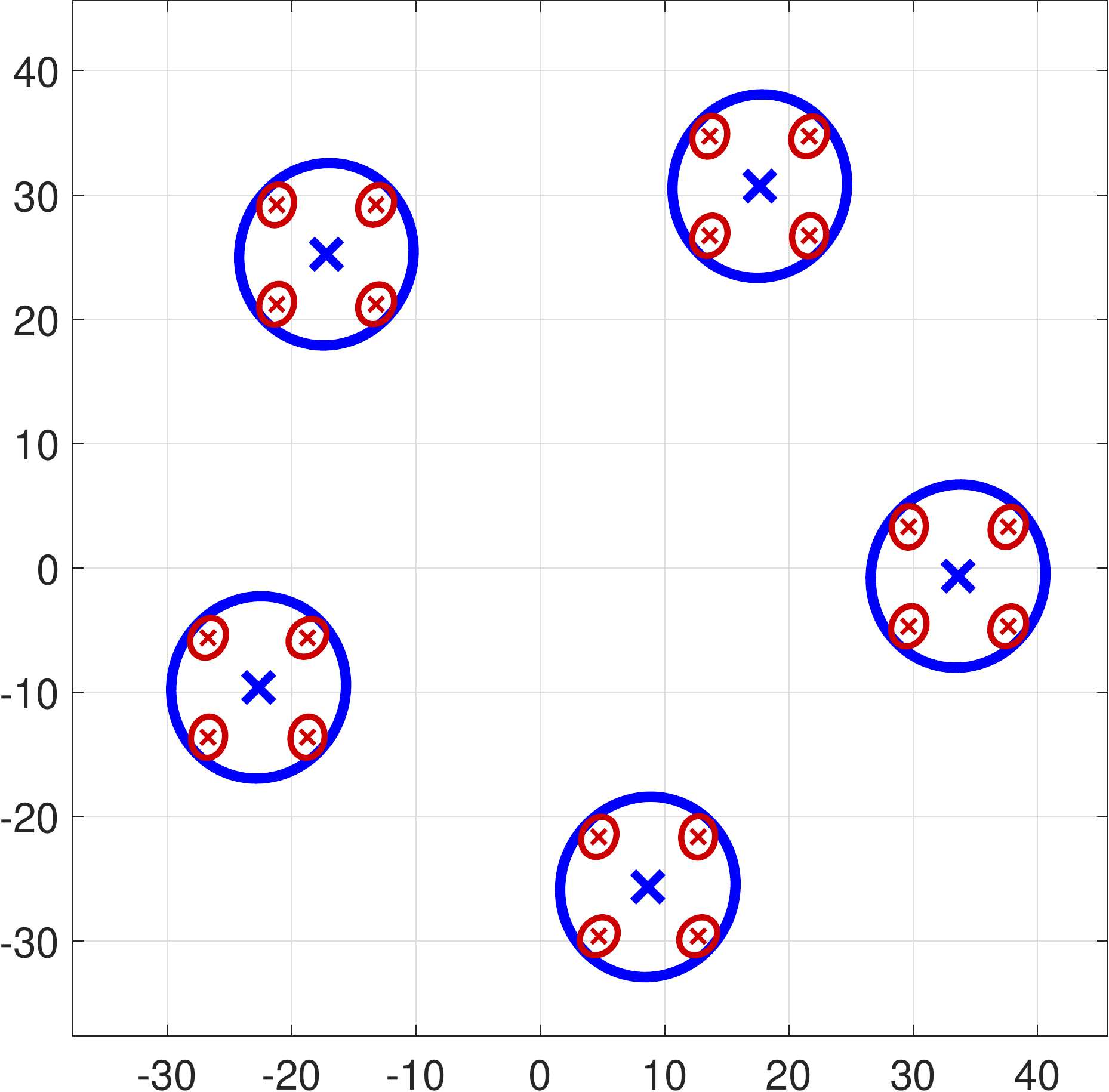}
         \caption{}
         \label{fig: small scale a}
     \end{subfigure}
          \centering
          \hfill
          \begin{subfigure}[b]{0.24\textwidth}
         \centering
         \includegraphics[width=\textwidth, trim={0cm 0cm 0cm 0cm
         },clip]{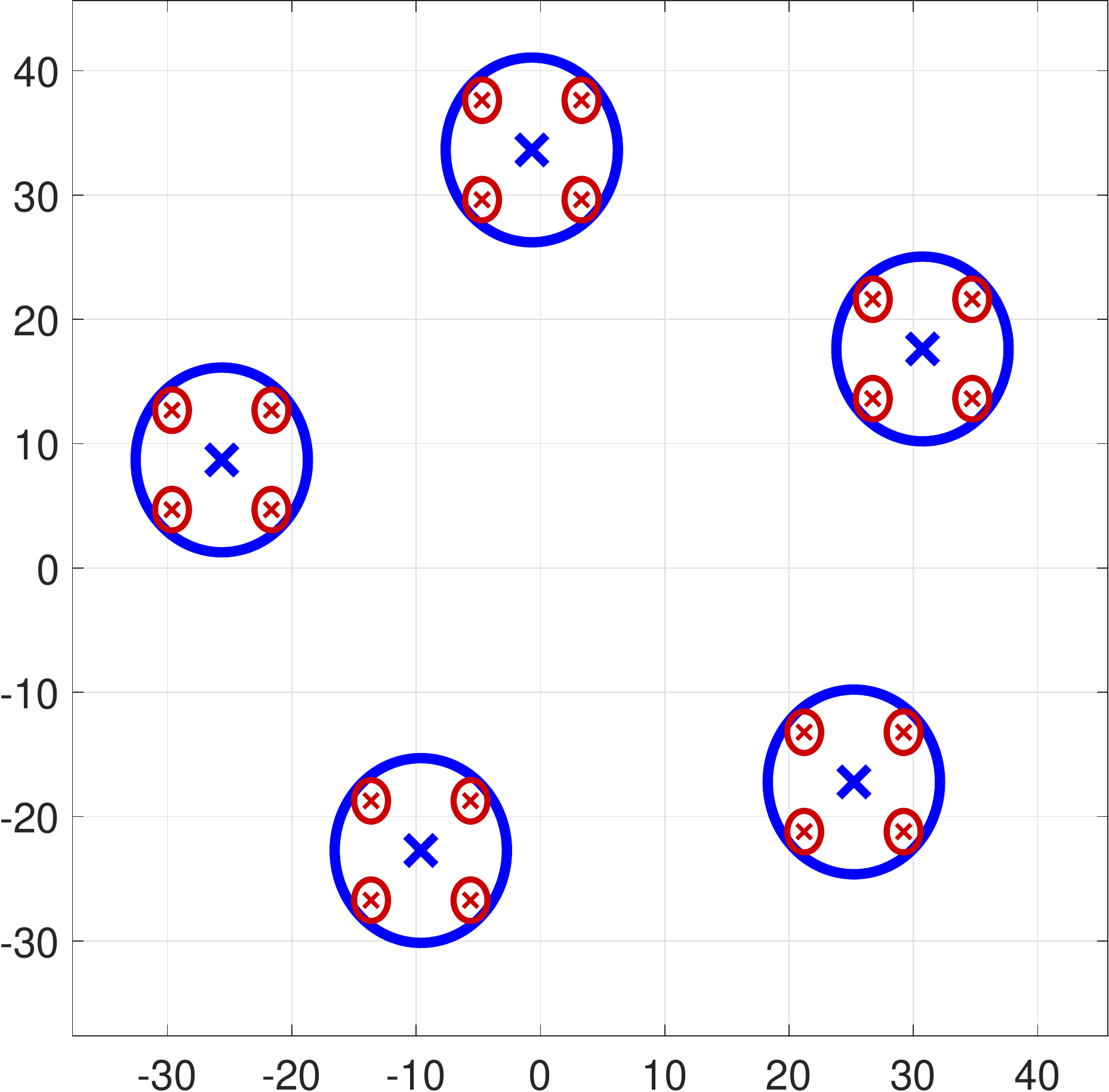}
         \caption{}
         \label{fig: small scale b}
     \end{subfigure}
          \centering
          \hfill
          \begin{subfigure}[b]{0.24\textwidth}
         \centering
         \includegraphics[width=\textwidth, trim={0cm 0cm 0cm 0cm
         },clip]{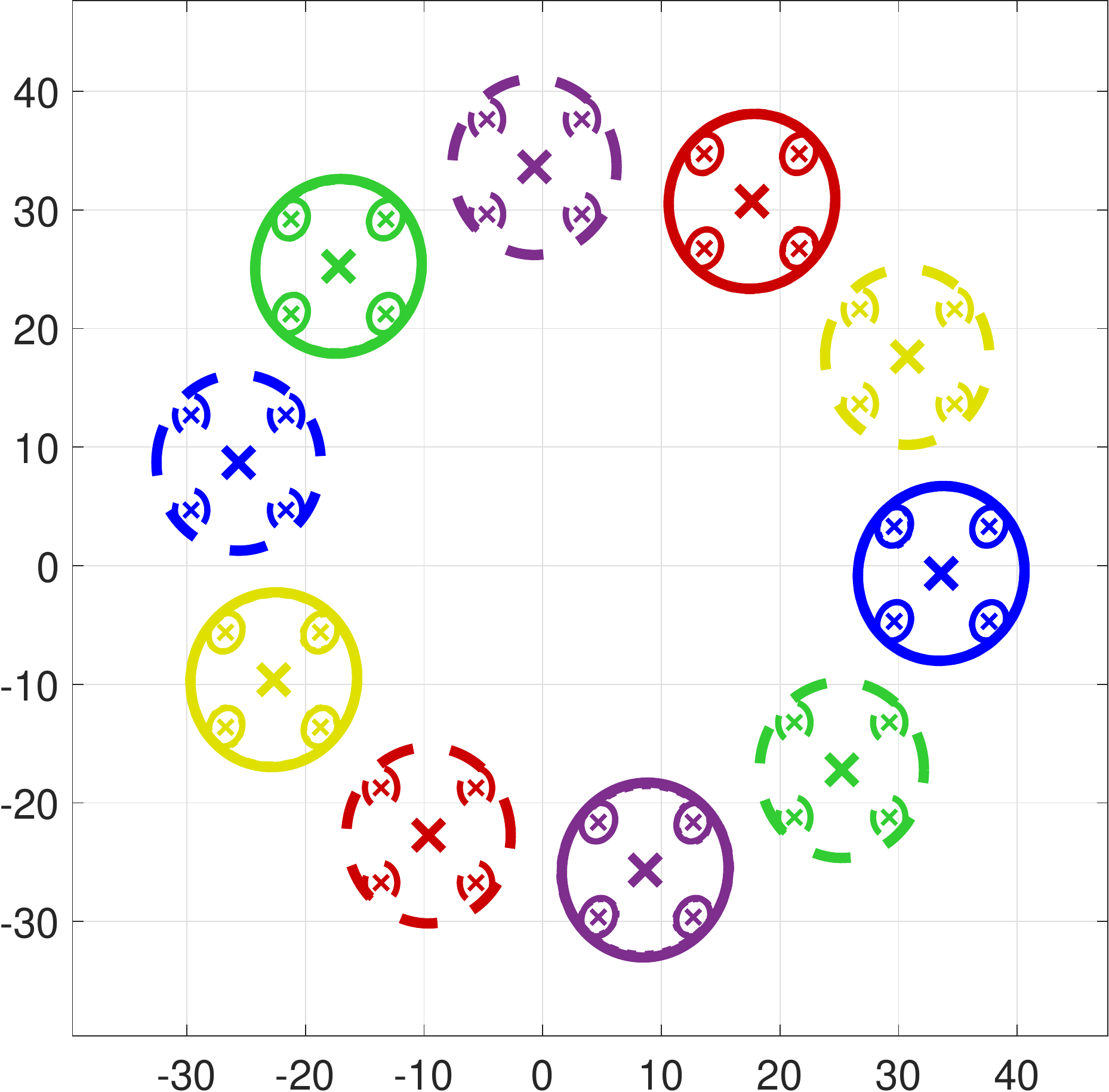}
         \caption{}
         \label{fig: small scale c}
     \end{subfigure}
          \centering
          \hfill
          \begin{subfigure}[b]{0.24\textwidth}
         \centering
         \includegraphics[width=\textwidth, trim={0cm 0cm 0cm 0cm
         },clip]{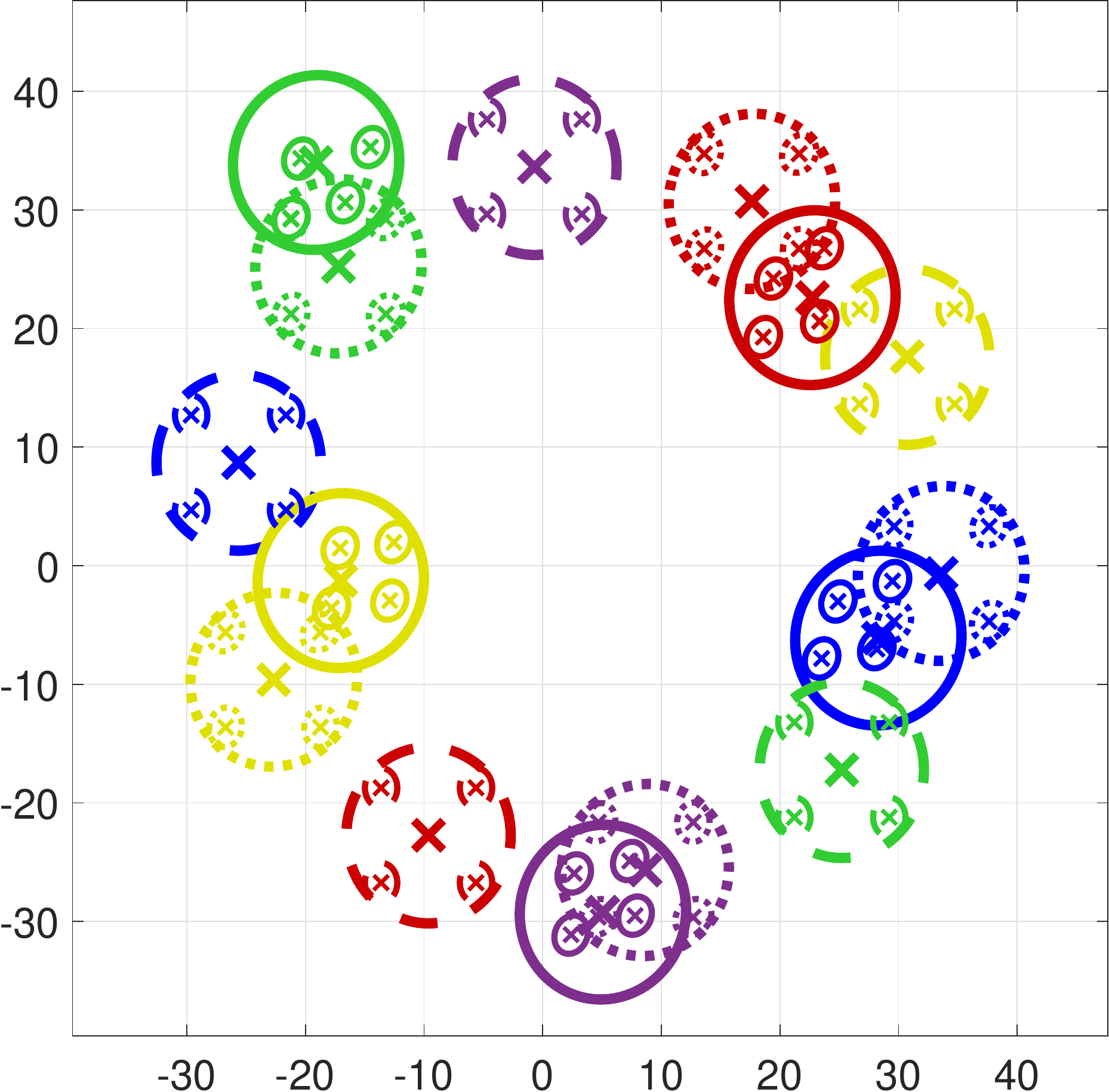}
         \caption{}
         \label{fig: small scale d}
     \end{subfigure}
     \\[0.2cm]
          \centering
          \begin{subfigure}[b]{0.24\textwidth}
         \centering
         \includegraphics[width=\textwidth, trim={0cm 0cm 0cm 0cm
         },clip]{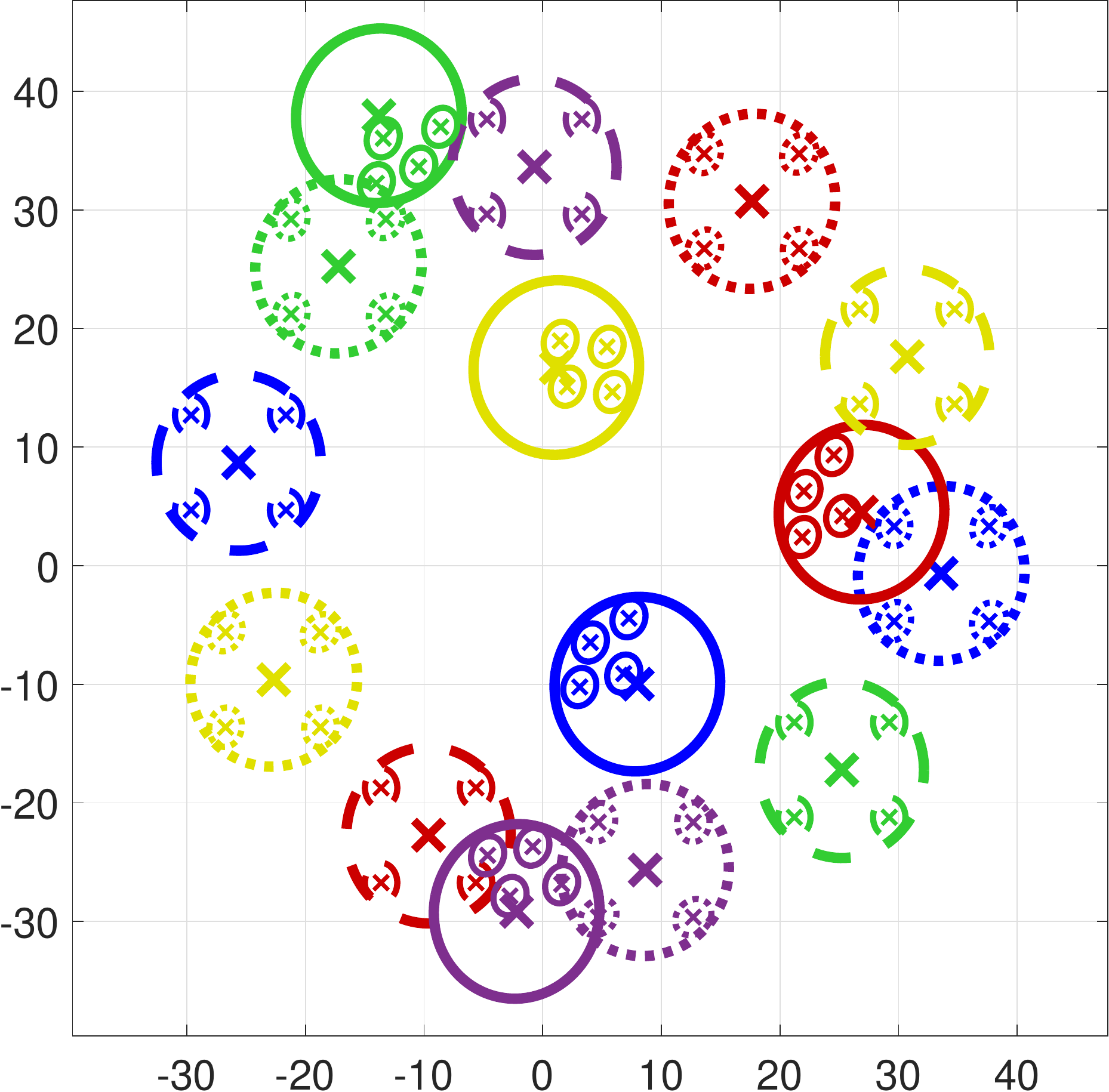}
         \caption{}
         \label{fig: small scale e}
     \end{subfigure}
          \centering
          \hfill
          \begin{subfigure}[b]{0.24\textwidth}
         \centering
         \includegraphics[width=\textwidth, trim={0cm 0cm 0cm 0cm
         },clip]{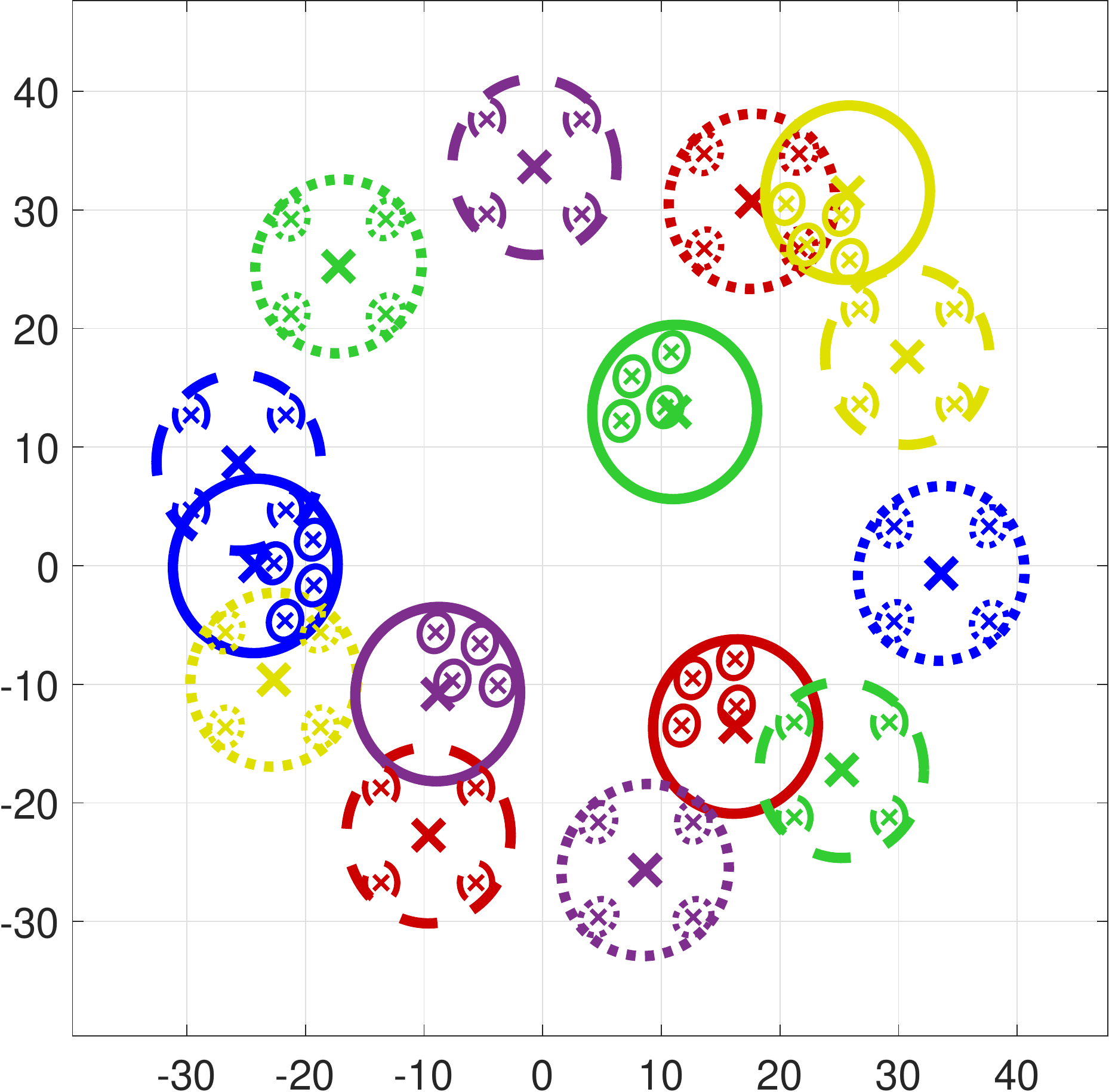}
         \caption{}
         \label{fig: small scale f}
     \end{subfigure}
          \centering
          \hfill
          \begin{subfigure}[b]{0.24\textwidth}
         \centering
         \includegraphics[width=\textwidth, trim={0cm 0cm 0cm 0cm
         },clip]{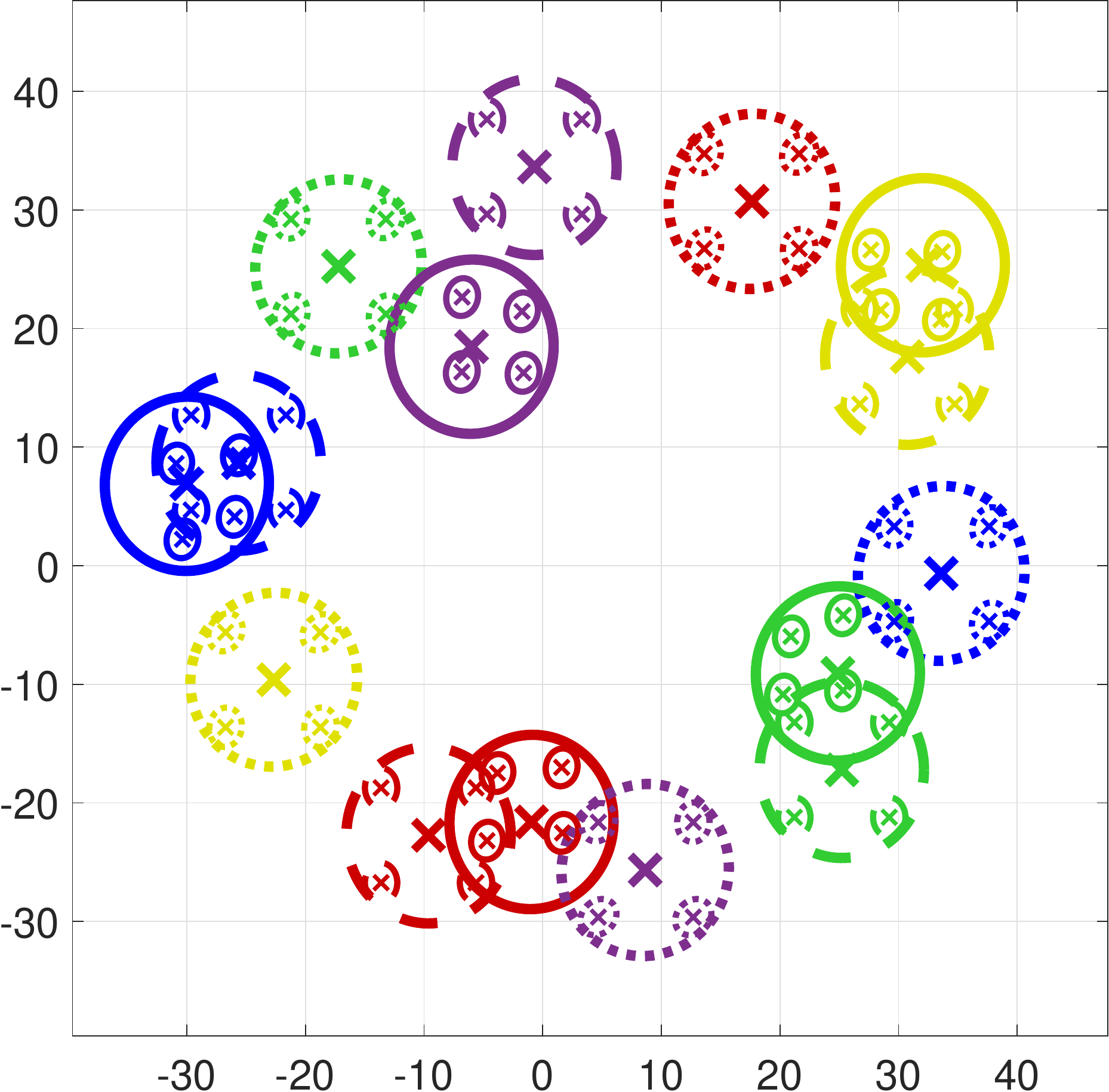}
         \caption{}
         \label{fig: small scale g}
     \end{subfigure}
          \centering
          \hfill
          \begin{subfigure}[b]{0.24\textwidth}
         \centering
         \includegraphics[width=\textwidth, trim={0cm 0cm 0cm 0cm
         },clip]{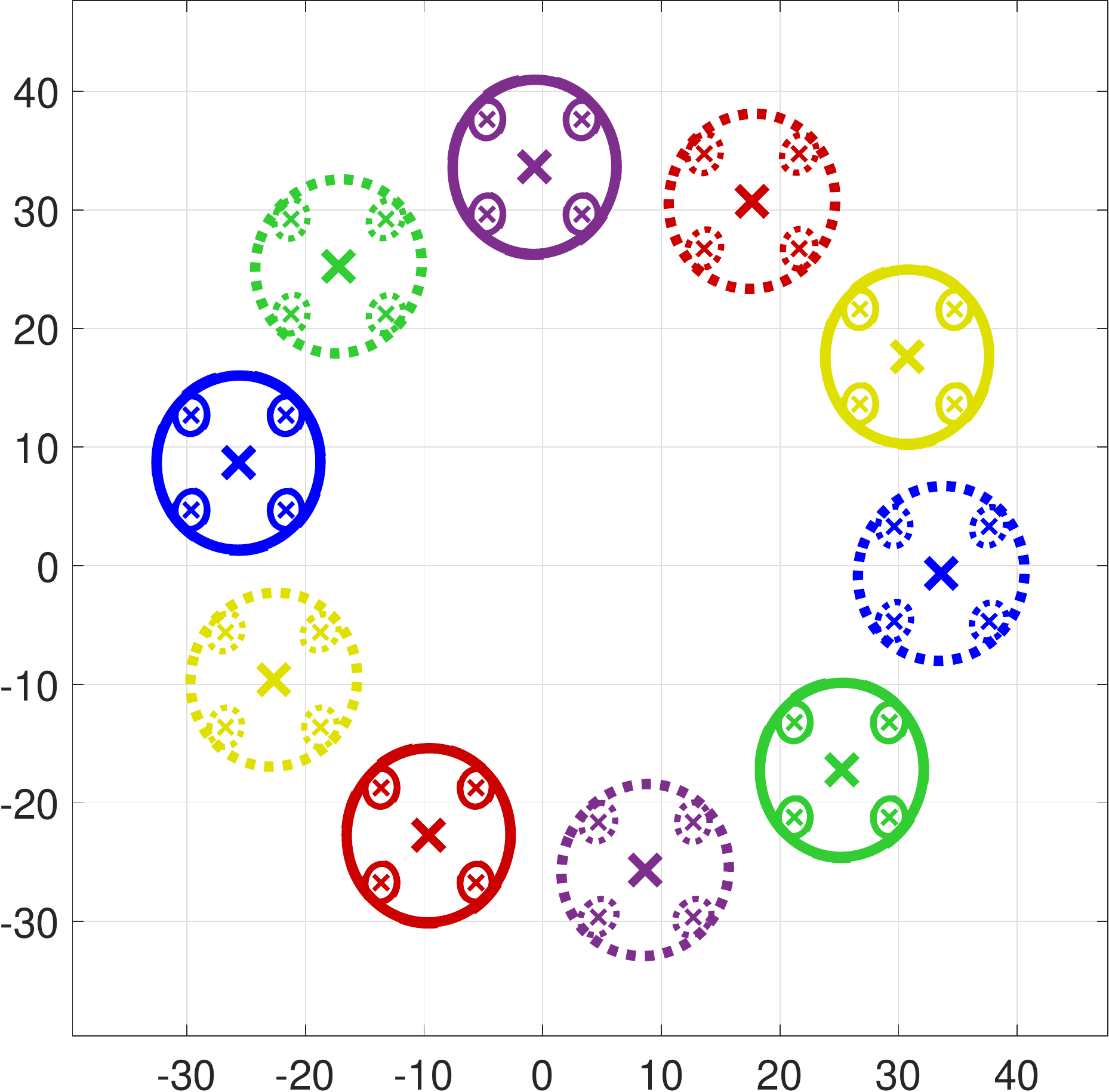}
         \caption{}
         \label{fig: small scale h}
     \end{subfigure}
        \caption{ \textbf{A two-level small-scale scenario.}  In (a), the $99.7\%$-confidence ellipses of the initial distributions are shown. Each level corresponds to a specific color (1: blue, 2: red). Similarly, (b) shows the target distributions for the two levels. In (c)-(h), the steering of all distributions is demonstrated for time instants $k=0,20,40,80,80,100$, respectively. At each snapshot, the current, initial and target distributions are shown with solid, dotted and dashed ellipses, respectively. Here, each color corresponds to a specific level-$1$ clique.}
        \label{fig: small scale}
\end{figure*}

Note that as emphasized in Remark \ref{remark: mean and cov separable}, we can separate solving \eqref{proposition: multi clique steering align} into a part that would only involve the variables $\bar{u}_i$, $i \in \calC_a^{\ell-1}$ and a second part including only the variables $L_i, K_i$, $i \in \calC_a^{\ell-1}$. Let us start with addressing the former part, which can be formulated as 
\begin{subequations}
\label{DHDS ubar: problem}
\begin{align}
& \{ \bar{u}_i \}_{i \in \calC_a^{\ell-1}} = \argmin \sum_{i \in \calC_a^{\ell-1}} 
\hat{J}_{i,1}^{\mathrm{s}}(\bar{u}_i)
\\[0.1cm]
\mathrm{s.t.} \quad
& \calf_{i,N}(\bar{u}_i) = 0, ~
\cals_{i,k}(\bar{u}_i) \geq 0, ~ 
\calp_{i,k}(\bar{u}_i) \leq 0, 
\\[0.1cm]
& \calq_{i,j,k}(\bar{u}_i, \bar{u}_j) \geq 0,
~  j \in \caln[\calC_i^\ell],
\label{local problem: interagent coupling constraint}
\\[0.1cm]
& k \in \llbracket 0, N \rrbracket, ~  
i \in \calC_a^{\ell-1}.
\nonumber
\end{align}
\end{subequations}
where $\hat{J}_{i,1}^{\mathrm{s}}(\bar{u}_i) = \bar{u}_i^\T \bar{R} \bar{u}_i$. To solve this problem in a distributed manner, we first need to introduce the augmented local variables
$
\tilde{u}_i = [ \bar{u}_i; \{ \bar{u}_j^i\}_{j \in \caln[\calC_i^\ell]} \big] 
$,
where $\bar{u}_j^i$ are copy variables for all $j \in \caln[\calC_i^\ell]$.
As in DHDE, we also need to introduce a global variable $b = [ \{b_i\}_{i \in \calC_a^{\ell-1}} ]$ and the constraints $\tilde{u}_i = \tilde{b}_i$, where $\tilde{b}_i = [ b_i; \{b_j\}_{j \in \caln[\calC_i^\ell]} ]$. Subsequently, we can arrive to a distributed ADMM algorithm for solving \eqref{DHDS ubar: problem}. For the full derivation and additional implementation details, the reader is referred to Sections \ref{SM sec: DHDS admm} and \ref{SM sec: DHDS details} of the SM.

The updates of the resulting algorithm are presented below. First, each level-$\ell$ agent updates its local variable $\tilde{u}_i$ by solving the following local optimization problem
\begin{subequations}
\label{DHDS ubar: local update}
\begin{align}
& ~~~~~~~~~ \tilde{u}_i = \argmin 
\tilde{J}_{i,1}^{\mathrm{s}}(\bar{u}_i)
\\[0.1cm]
\mathrm{s.t.} \quad
& \calf_{i,N}(\bar{u}_i) = 0, ~
\cals_{i,k}(\bar{u}_i) \geq 0, ~ 
\calp_{i,k}(\bar{u}_i) \leq 0, 
\\[0.1cm]
& \calq_{i,j,k}(\bar{u}_i, \bar{u}_j^i) \geq 0,
~  j \in \caln[\calC_i^\ell], ~ k \in \llbracket 0, N \rrbracket.
\end{align}
\end{subequations}
with each cost $\tilde{J}_{i,1}^{\mathrm{s}}(\tilde{u}_i)$ given by
\begin{align}
\tilde{J}_{i,1}^{\mathrm{s}}(\tilde{u}_i) & = \hat{J}_{i,1}^{\mathrm{s}}(\bar{u}_i) 
+ v_i^\T(\tilde{u}_i - \tilde{b}_i)
+ \frac{\rho_u}{2} \| \tilde{u}_i - \tilde{b}_i \|_2^2.
\end{align}
where $\rho_u > 0$ and $v_i$ are the dual variables corresponding to the constraints $\tilde{u}_i = \tilde{b}_i$. Subsequently, all global variables components are updated with 
\begin{equation}
\label{DHDS ubar: global update}
b_i \leftarrow \frac{1}{|\calm'[\calC_i^\ell]|} \sum_{j \in \calm'[\calC_i^\ell]} \bar{u}_i^j
\end{equation}
and, finally, the dual updates are given by 
\begin{equation}
\label{DHDS ubar: dual update}
v_i \leftarrow v_i + \rho_{u} (\tilde{u}_i - \tilde{b}_i).
\end{equation}
Again, all updates in \eqref{DHDS ubar: local update}, \eqref{DHDS ubar: global update} and \eqref{DHDS ubar: dual update} can be performed in parallel by each level-$\ell$ agent $i$. Therefore, all computations are operated in a decentralized fashion. This iterative algorithm repeats until a prespecified termination criterion is met.

Let us now focus on the part of problem \eqref{proposition: multi clique steering align} that only involves the variables $L_i, K_i$. This subproblem can be formulated as 
\begin{subequations}
\label{DHDS: feedback problem}
\begin{align}
& \big\{ L_i, K_i \big\}_{i \in \calC_a^{\ell-1}} = \argmin \sum_{i \in \calC_a^{\ell-1}} 
\hat{J}_{i,2}^{\mathrm{s}}(L_i, K_i)
\\[0.1cm]
\mathrm{s.t.} \quad
& \calF_{i,N}(L_i, K_i) \succeq 0, 
~ \calQ_{i,k}(L_i, K_i) \succeq 0, 
\\[0.1cm] 
& k \in \llbracket 0, N \rrbracket, ~  
i \in \calC_a^{\ell-1},
\nonumber
\end{align}
\end{subequations}
where $\hat{J}_{i,2}^{\mathrm{s}}(\bar{u}_i) = \tr(\bar{R} K_i W K_i^\T+ \bar{R} L_i \Sigma_{i,0} L_i^\T)$. A key observation here is that there exists no inter-agent coupling between the optimization variables of different agents, so we can further split the problem and independently solve each agent's problem as follows
\begin{align}
& ~~~~~~~~ \{L_i, K_i\} = \argmin 
\hat{J}_{i,2}^{\mathrm{s}}(L_i, K_i)
\label{DHDS: feedback problem 2}
\\[0.1cm]
\mathrm{s.t.} \quad
& \calF_{i,N}(L_i, K_i) \succeq 0, 
~ \calQ_{i,k}(L_i, K_i) \succeq 0,
~ k \in \llbracket 0, N \rrbracket.
\nonumber
\end{align}
for each $i \in \calC_a^{\ell-1}$.

The DHDS algorithm with all required computation and communication steps is demonstrated in Alg. \ref{DHDS Algorithm}. The algorithm is executed in a top-down manner for $\ell = \{1, \dots, L\}$. For a specific level $\ell$, each agent $a$ that correspond to a parent clique $\calC_a^{\ell -1} \in \calR^{\ell-1}$ (if $\ell > 1$) must send the variables $\mu_{a,k}^{\ell-1}, \Sigma_{a,k}^{\ell-1}, \ k \in \llbracket 0,N \rrbracket$, to all agents $i \in \calC_a^{\ell -1}$ (Line 9). Afterwards, the iterative ADMM algorithm starts for every separate group of agents that corresponds to a particular clique $\calC_a^{\ell -1}$ (Lines 10-18). First, each agent $i \in \calC_a^{\ell -1}$ solves its local subproblem \eqref{DHDS ubar: local update} to update its local variable $\tilde{u}_i$ (Line 12). Then, each agent $i$ receives the copy variables $\bar{u}_i^j$ from all $j \in \calm[\calC_i^\ell]$ (Line 13), so that the variables $b_i$ are updated (Line 15) through \eqref{DHDS ubar: global update}. Subsequently, all agents $j \in \caln[\calC_i^\ell]$ send $b_j$ to each $i$ (Line 16) and each agent $i$ updates its dual variable $v_i$ (Line 18) with \eqref{DHDS ubar: dual update}. This iterative procedure terminates once a predefined termination criterion is met. Finally, each agent $i \in \calC_a^{\ell -1}$ also computes its optimal feedback gains $\{L_i, K_i\}$ (Line 20) through solving \eqref{DHDS: feedback problem 2}. After the optimal variables $\{\bar{u}_i, L_i, K_i\}$ have been found for all $i \in \calC_a^{\ell -1}$, then the state means and covariances $\mu_{i,k}^\ell, \Sigma_{i,k}^\ell$ can be obtained (Line 21) through \eqref{state mean covs}. Subsequently, the same procedure repeats for the next level $\ell+1$, and so on, until level $L$ is reached.

\begin{figure*}
    \hfill
     \centering
          \begin{subfigure}[b]{0.32\textwidth}
         \centering
         \includegraphics[width=\textwidth, trim={0cm 0cm 0cm 0cm
         },clip]{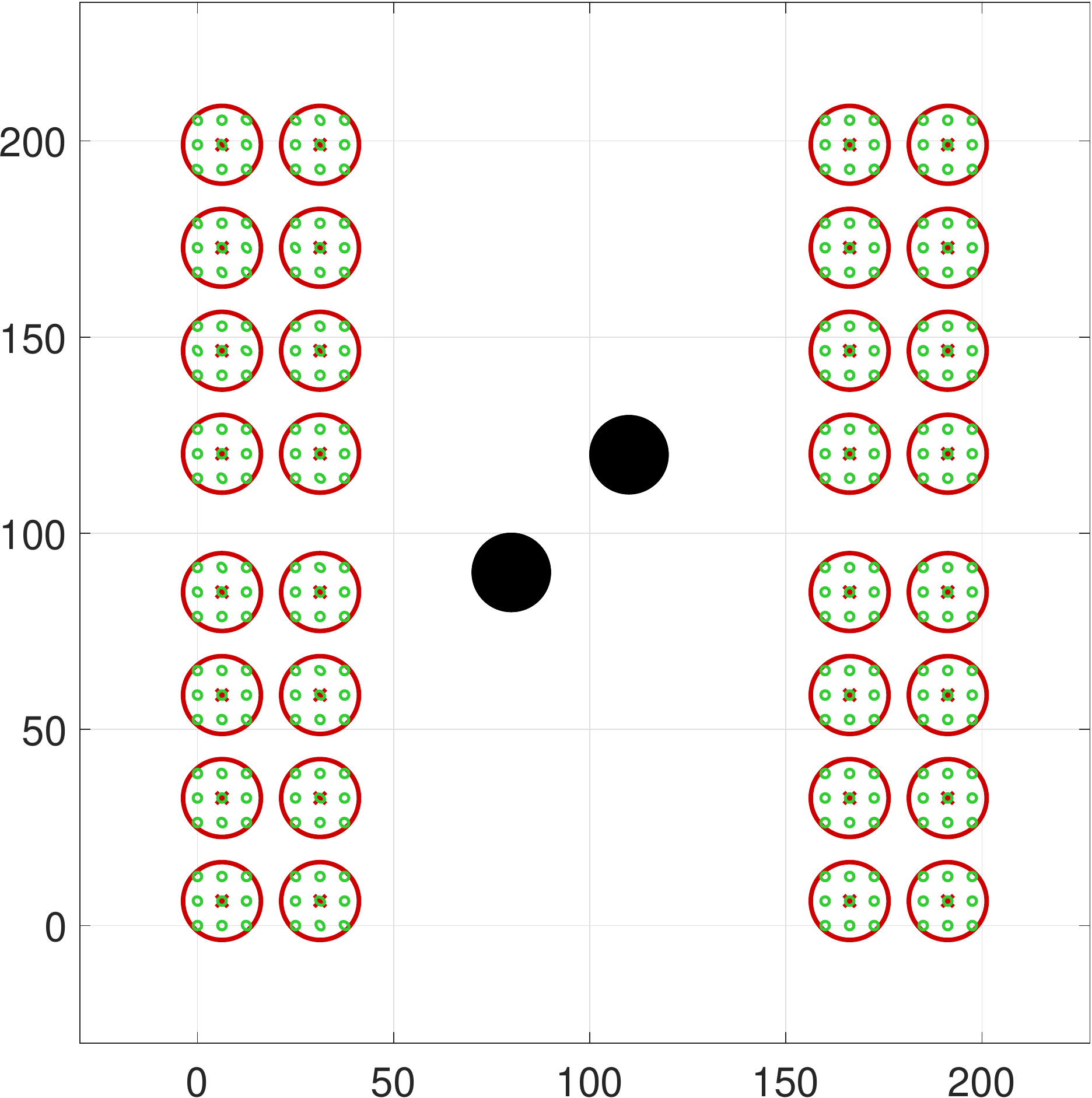}
         \caption{}
         \label{fig: mid scale a}
     \end{subfigure}
    \hfill
     \centering
          \begin{subfigure}[b]{0.32\textwidth}
         \centering
         \includegraphics[width=\textwidth, trim={0cm 0cm 0cm 0cm
         },clip]{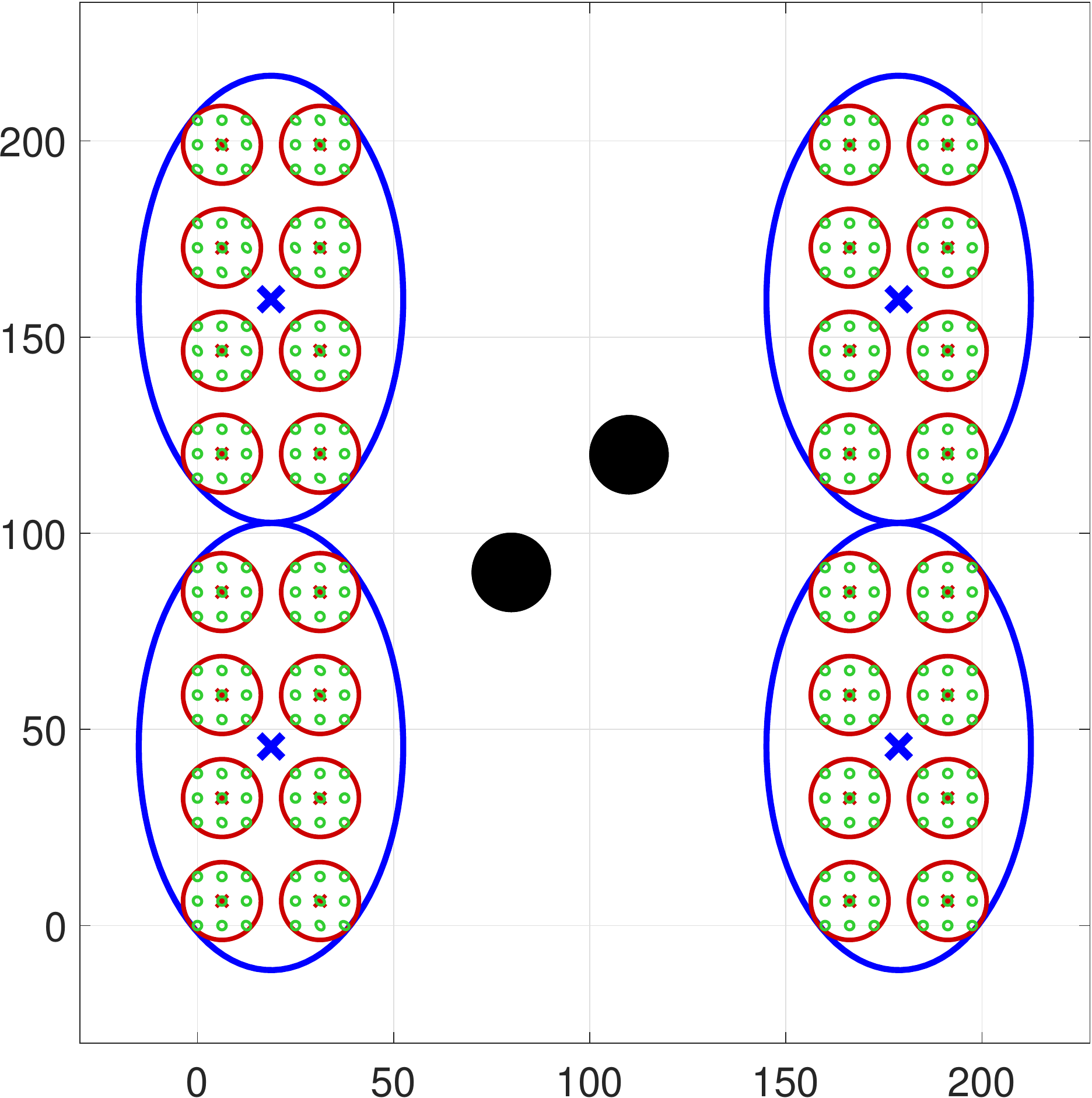}
         \caption{}
         \label{fig: mid scale b}
     \end{subfigure}
     \hfill
          \centering
          \begin{subfigure}[b]{0.32\textwidth}
         \centering
         \includegraphics[width=\textwidth, trim={0cm 0cm 0cm 0cm
         },clip]{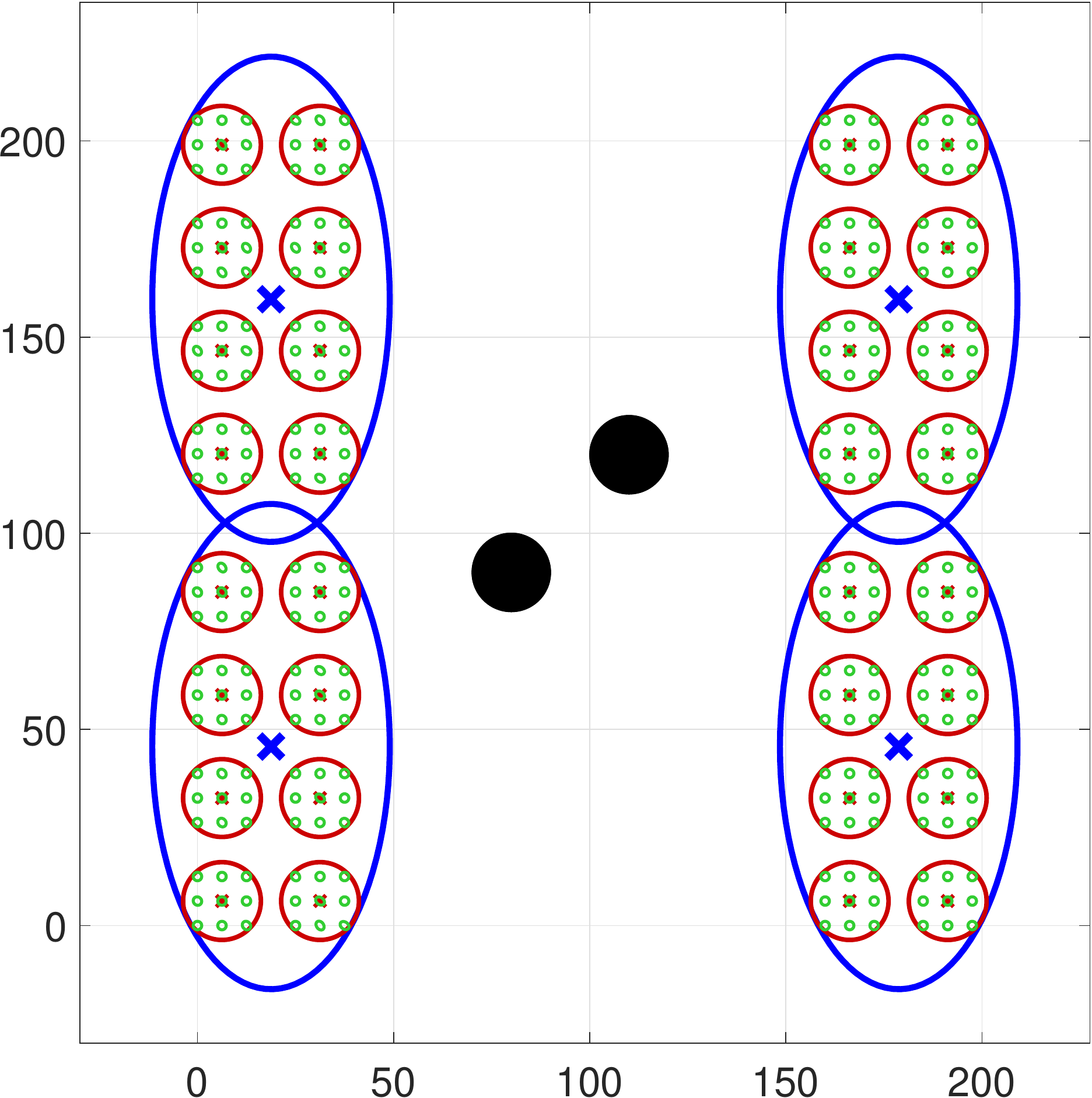}
         \caption{}
         \label{fig: mid scale c}
     \end{subfigure}
     \hfill
     \\[0.2cm]
    \hfill
     \centering
          \begin{subfigure}[b]{0.32\textwidth}
         \centering
         \includegraphics[width=\textwidth, trim={0cm 0cm 0cm 0cm
         },clip]{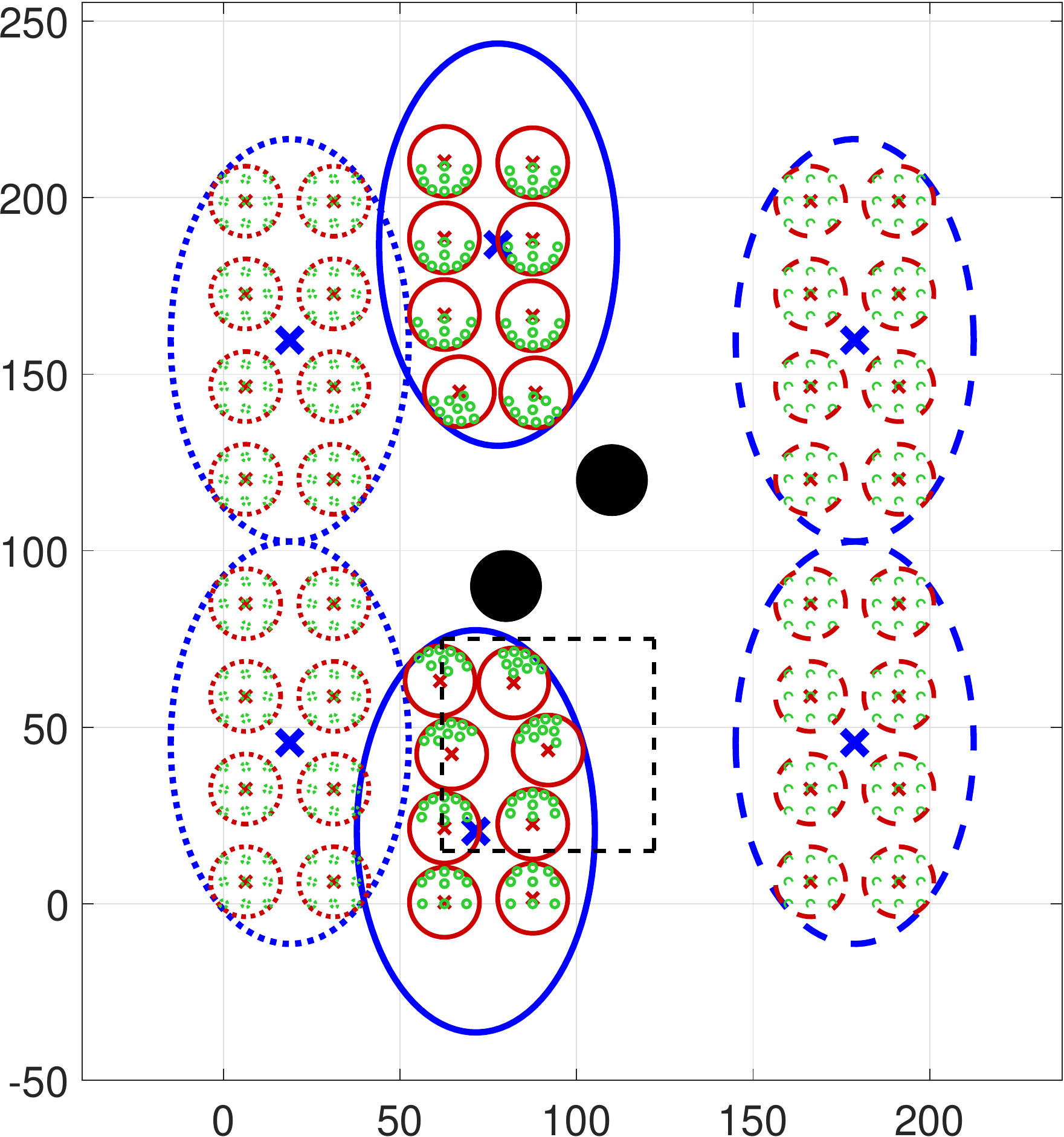}
         \caption{}
         \label{fig: mid scale d}
     \end{subfigure}
               \hfill
          \centering
          \begin{subfigure}[b]{0.32\textwidth}
         \centering
         \includegraphics[width=\textwidth, trim={0cm 0cm 0cm 0cm
         },clip]{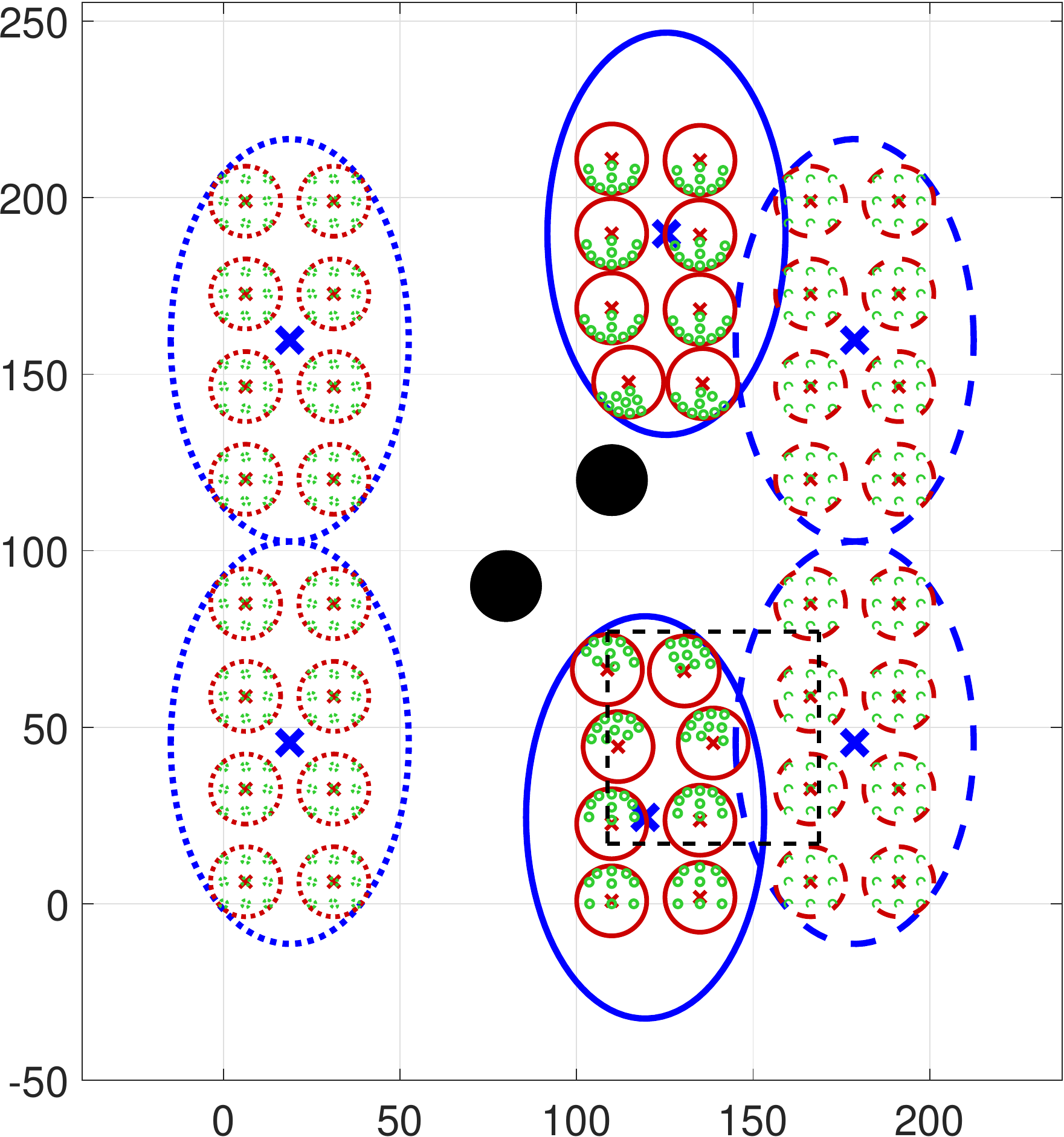}
         \caption{}
         \label{fig: mid scale e}
     \end{subfigure}
               \hfill
          \centering
          \begin{subfigure}[b]{0.32\textwidth}
         \centering
         \includegraphics[width=\textwidth, trim={0cm 0cm 0cm 0cm
         },clip]{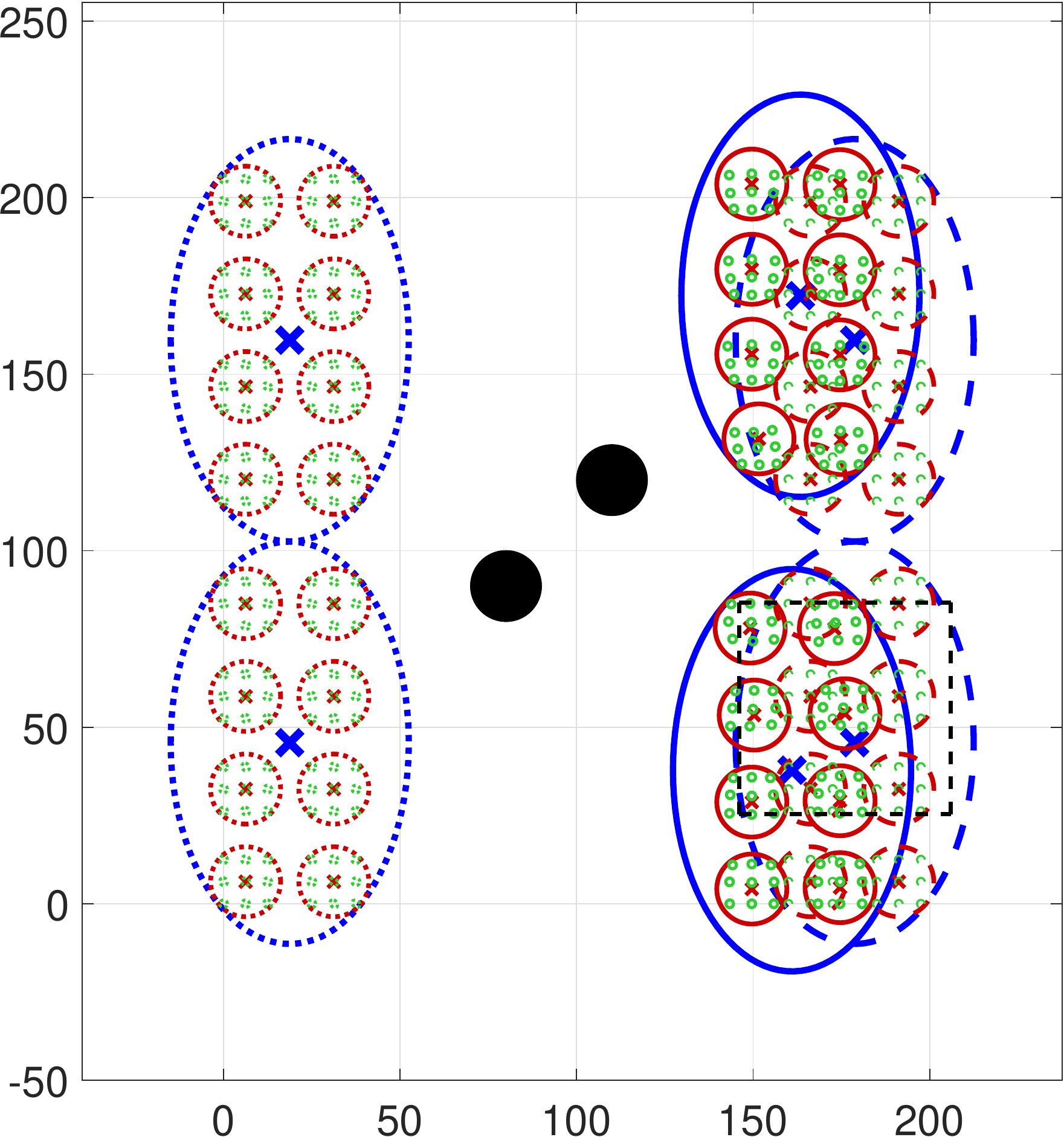}
         \caption{}
         \label{fig: mid scale f}
     \end{subfigure}
               \hfill
     \\[0.2cm]
               \hfill
          \centering
          \begin{subfigure}[b]{0.32\textwidth}
         \centering
         \includegraphics[width=\textwidth, trim={0cm 0cm 0cm 0cm
         },clip]{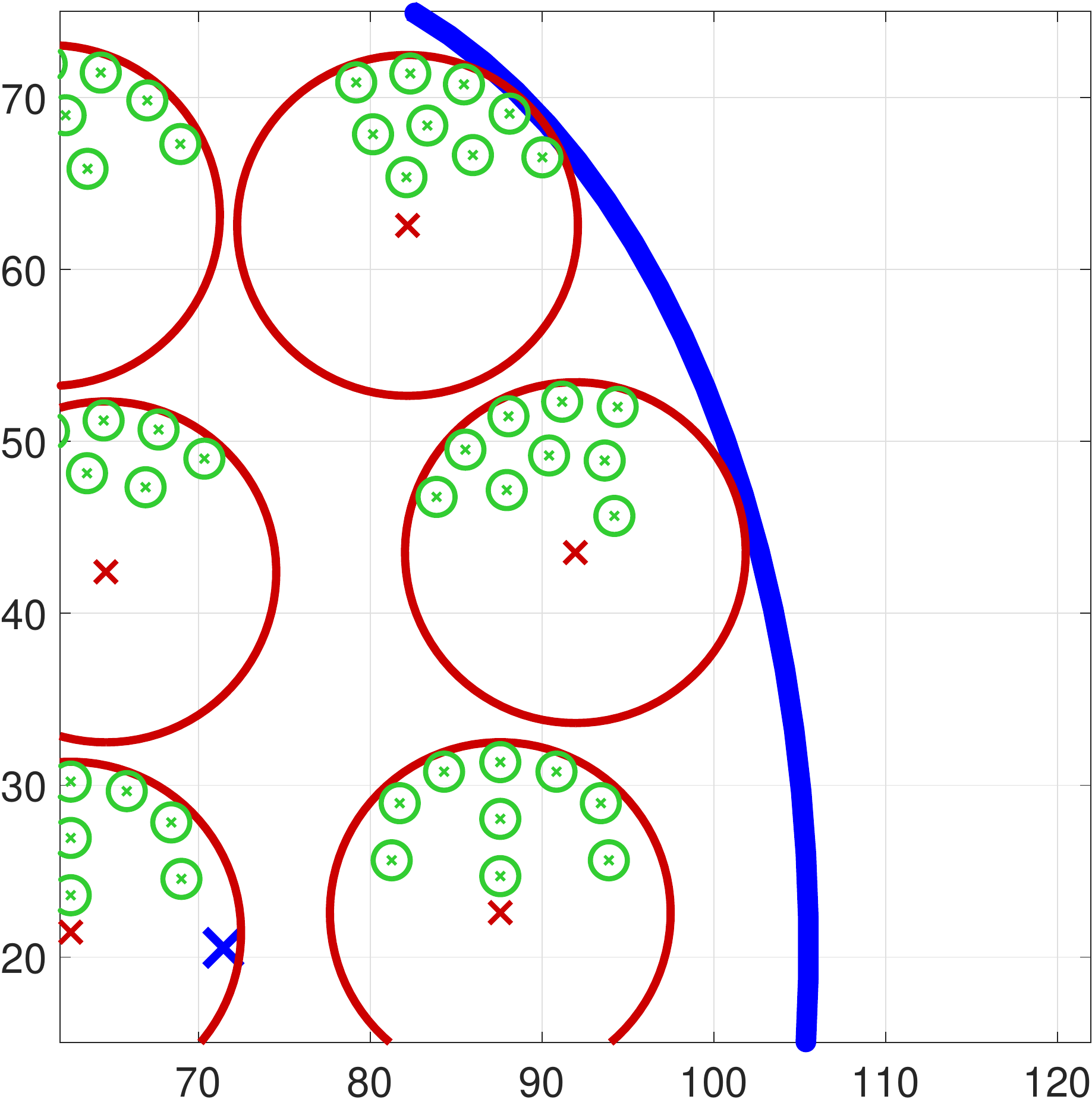}
         \caption{}
         \label{fig: mid scale g}
     \end{subfigure}
     \hfill
     \centering
          \begin{subfigure}[b]{0.32\textwidth}
         \centering
         \includegraphics[width=\textwidth, trim={0cm 0cm 0cm 0cm
         },clip]{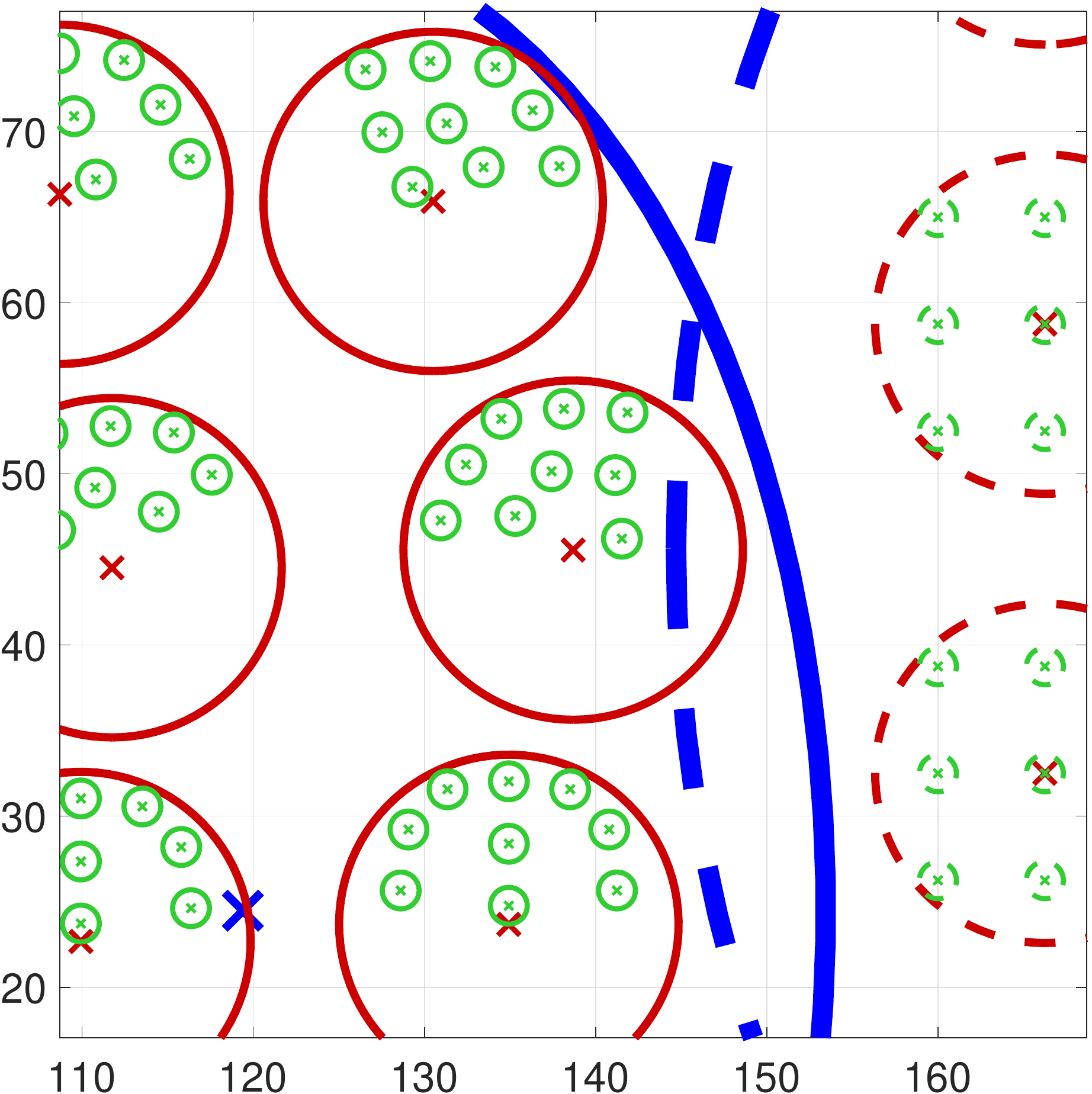}
         \caption{}
         \label{fig: mid scale h}
     \end{subfigure}
          \hfill
          \centering
          \begin{subfigure}[b]{0.32\textwidth}
         \centering
         \includegraphics[width=\textwidth, trim={0cm 0cm 0cm 0cm
         },clip]{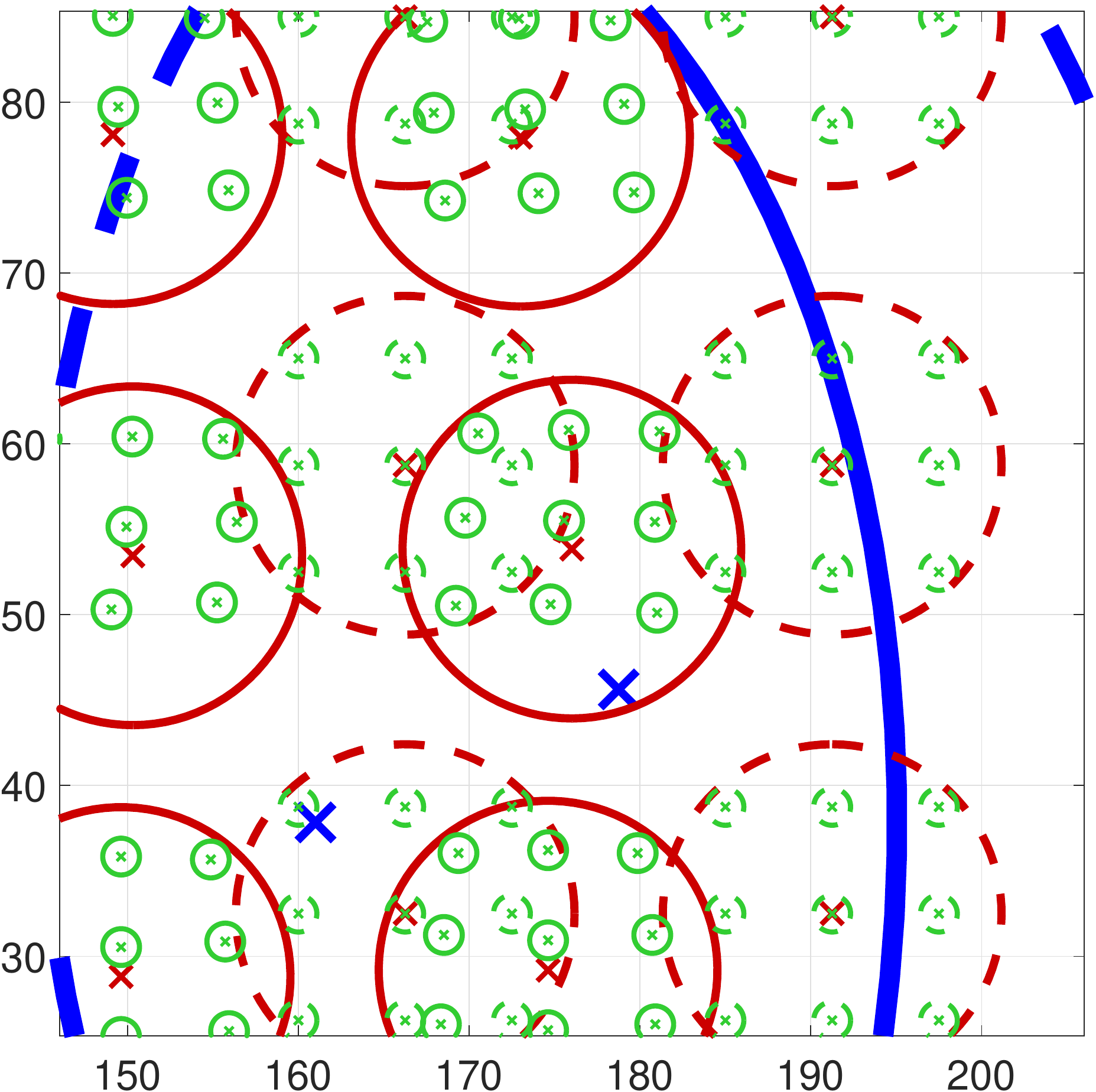}
         \caption{}
         \label{fig: mid scale i}
     \end{subfigure}
     
        \caption{ \textbf{A three-level large-scale scenario.} In (a), the initial (LHS) and target (RHS) distributions of the robots (level $3$, green) are shown along with the level-$2$ ones (red) that are computed by DHDE. In (b), the level-$1$ distributions (blue) that are obtained by DHDE are illustrated. In (c), we show the resulting (overlapping) level-$1$ distributions if the inter-clique non-overlapping constraints are omitted. In (d)-(f), the distributions of all levels are steered to their targets (time instants $k=40,60,80$, respectively), while avoiding the obstacles in the middle (black circles). In (g)-(i), we focus into the black dotted boxes of the above figures and show the motion of the level-$3$ distributions in more detail.}
        \label{fig: mid scale}
\end{figure*}

\begin{figure*}
     \centering
          \begin{subfigure}[b]{0.495\textwidth}
         \centering
         \includegraphics[width=\textwidth, trim={0cm 0cm 0cm 0cm
         },clip]{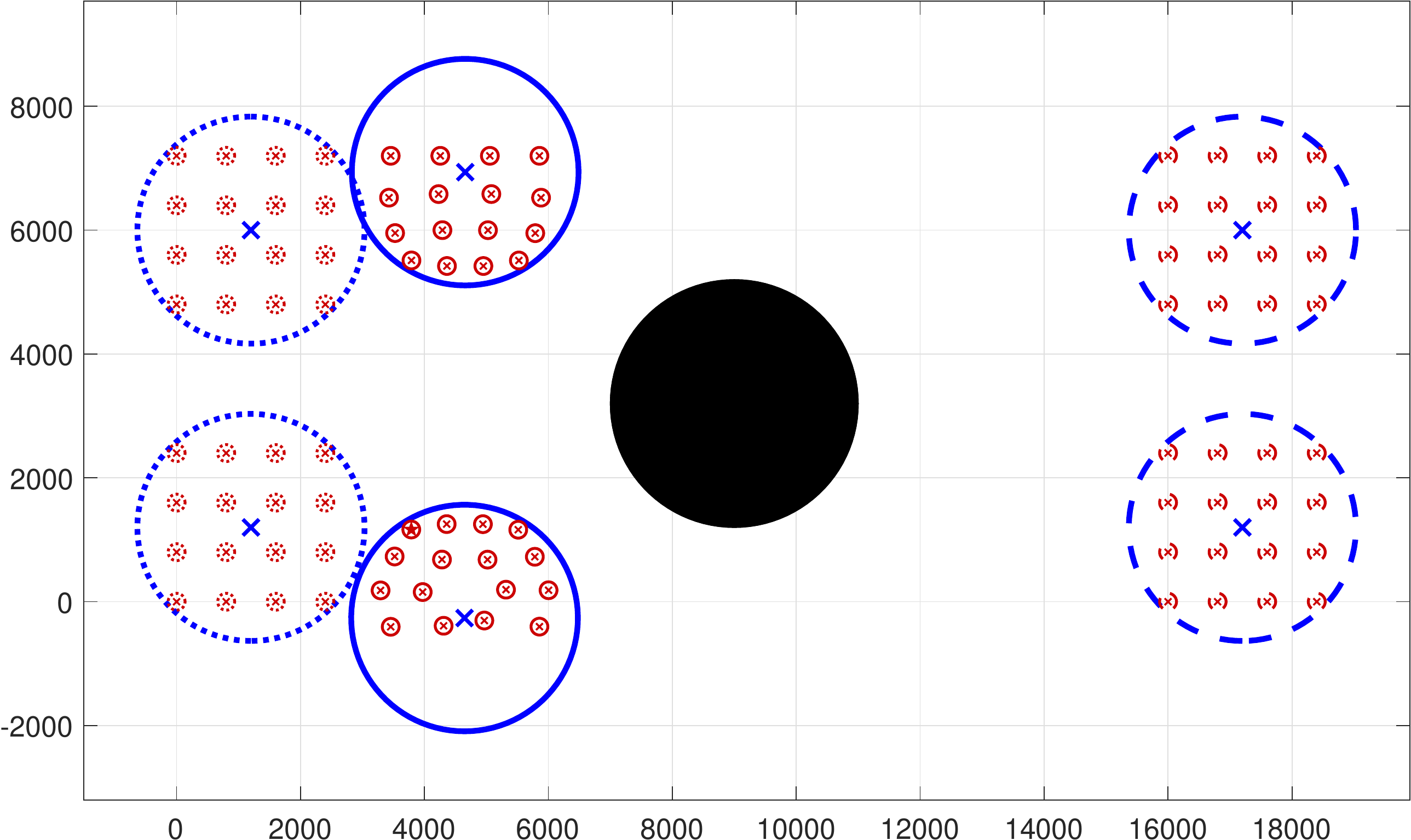}
         \caption{}
         \label{fig: VLMAS scale a}
     \end{subfigure}
          \centering
          \hfill
          \begin{subfigure}[b]{0.495\textwidth}
         \centering
         \includegraphics[width=\textwidth, trim={0cm 0cm 0cm 0cm
         },clip]{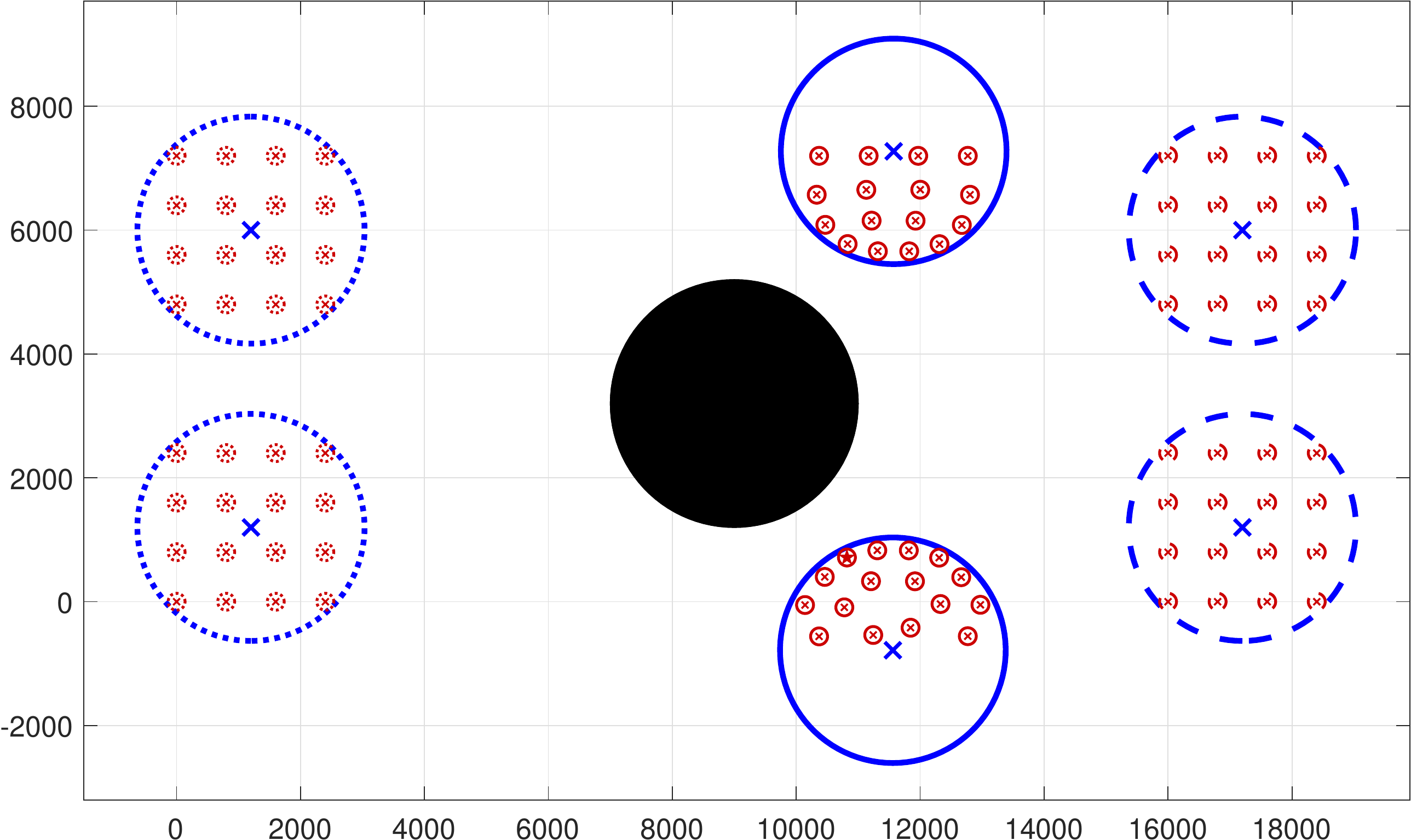}
         \caption{}
         \label{fig: VLMAS scale b}
     \end{subfigure}
     \\[0.15cm]
          \centering
          \begin{subfigure}[b]{0.495\textwidth}
         \centering
         \includegraphics[width=\textwidth, trim={0cm 0cm 0cm 0cm
         },clip]{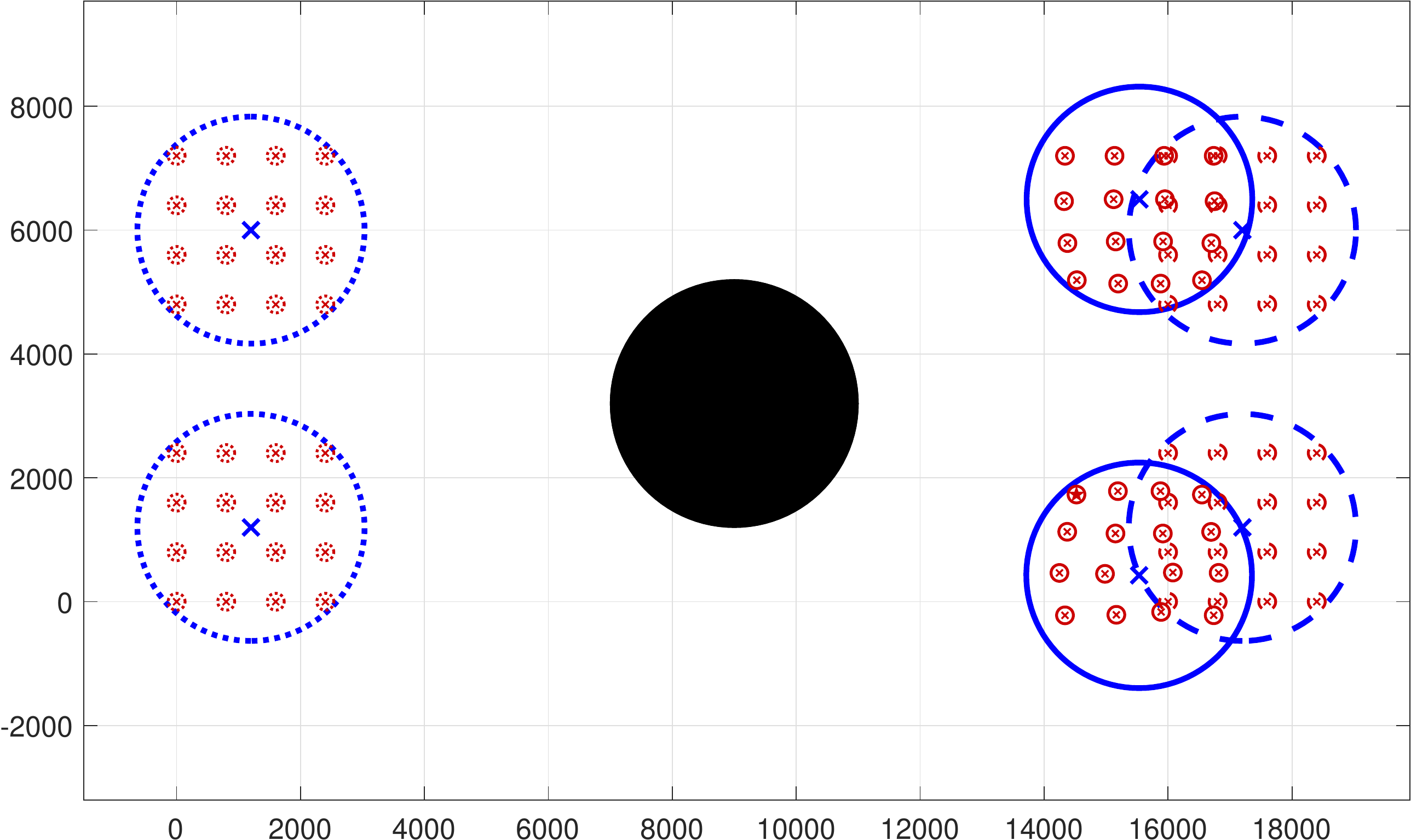}
         \caption{}
         \label{fig: VLMAS scale c}
     \end{subfigure}
       \centering
          \hfill
          \begin{subfigure}[b]{0.495\textwidth}
         \centering
         \includegraphics[width=\textwidth, trim={0cm 0cm 0cm 0cm
         },clip]{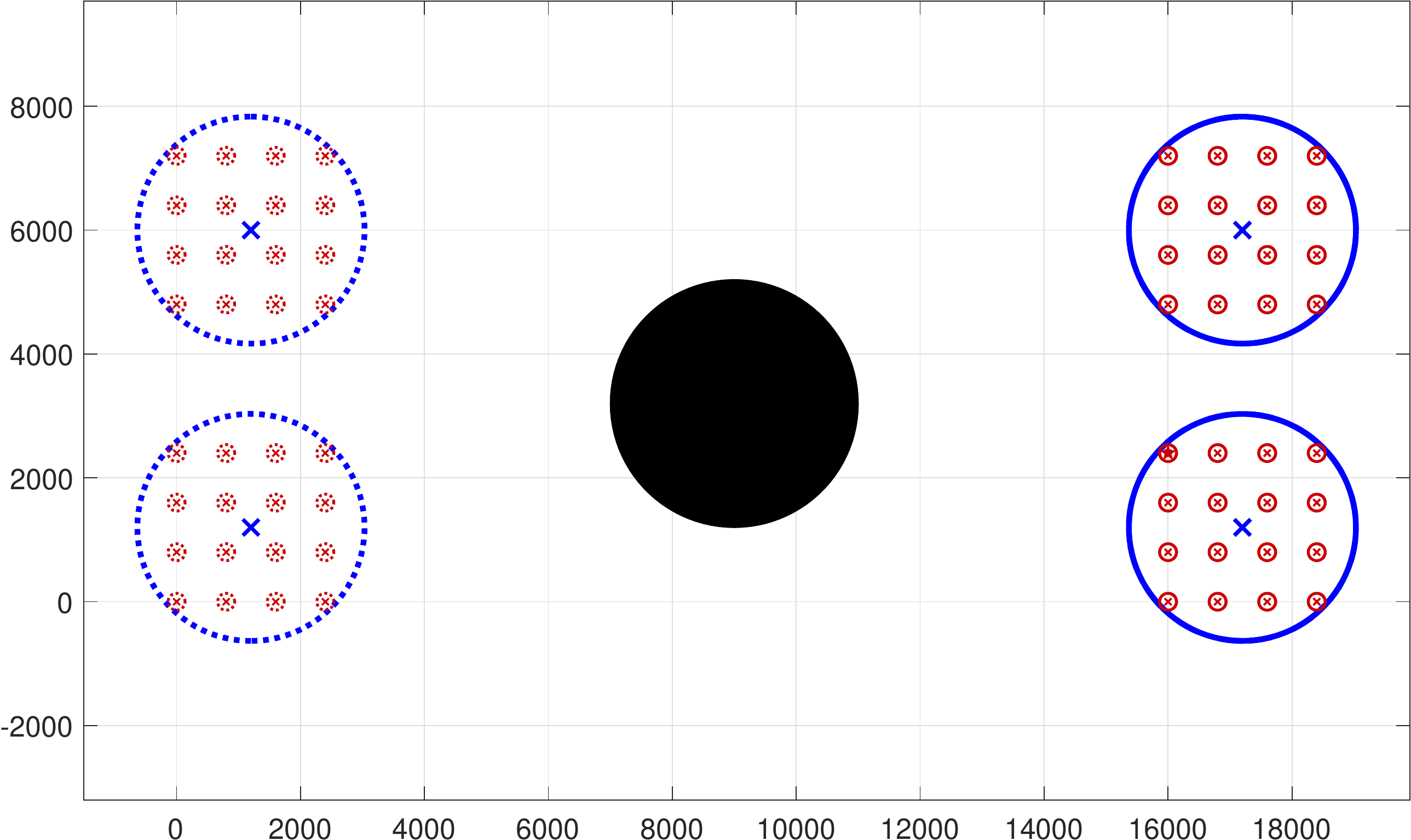}
         \caption{}
         \label{fig: VLMAS scale d}
     \end{subfigure}
     \\[0.2cm]
    \centering
          \begin{subfigure}[b]{0.16\textwidth}
         \centering
         \includegraphics[width=\textwidth, trim={0cm 0cm 0cm 0cm
         },clip]{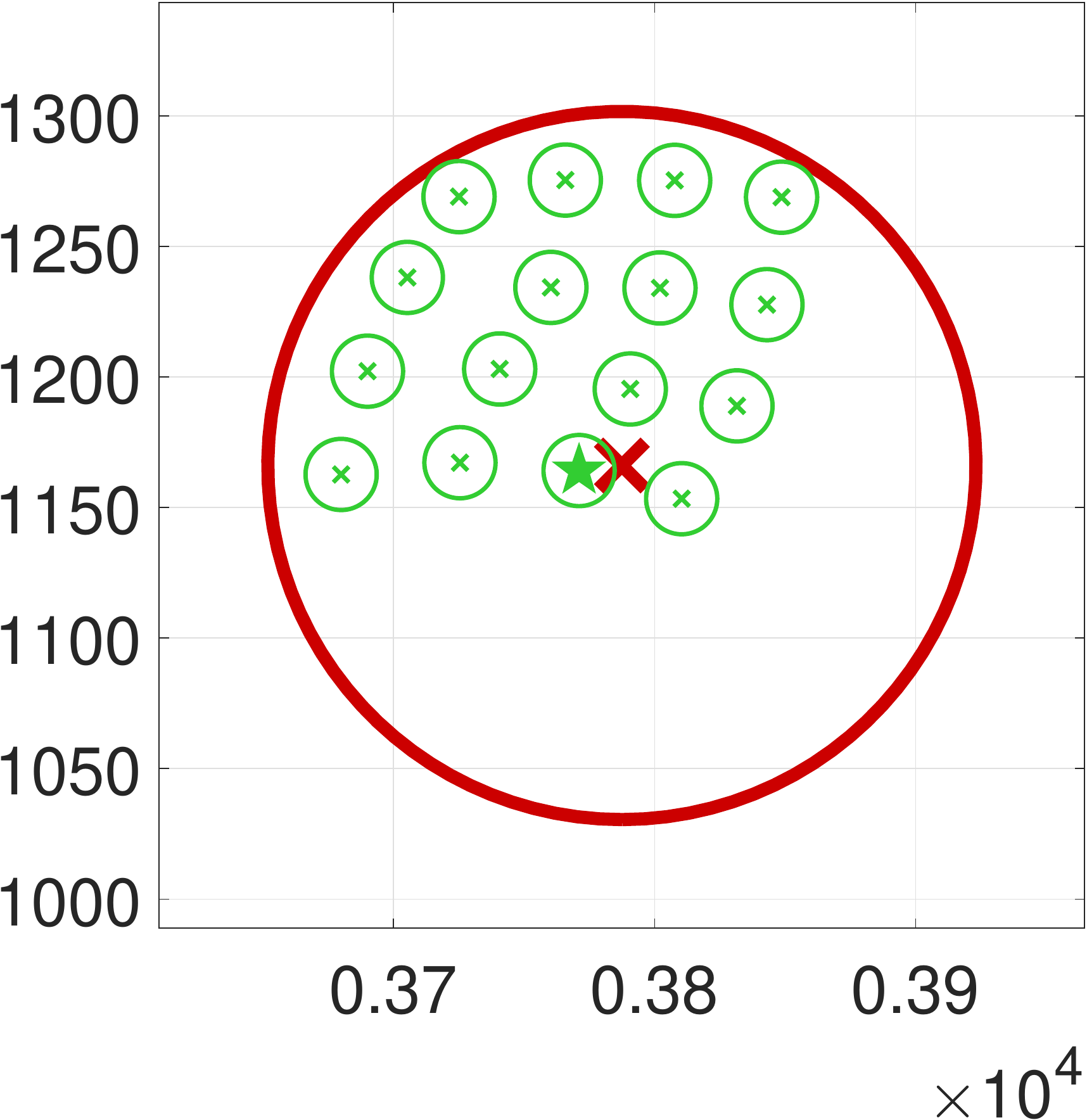}
         \caption{}
         \label{fig: VLMAS scale e}
     \end{subfigure}
          \centering
          \hfill
          \begin{subfigure}[b]{0.155\textwidth}
         \centering
         \includegraphics[width=\textwidth, trim={0cm 0cm 0cm 0cm
         },clip]{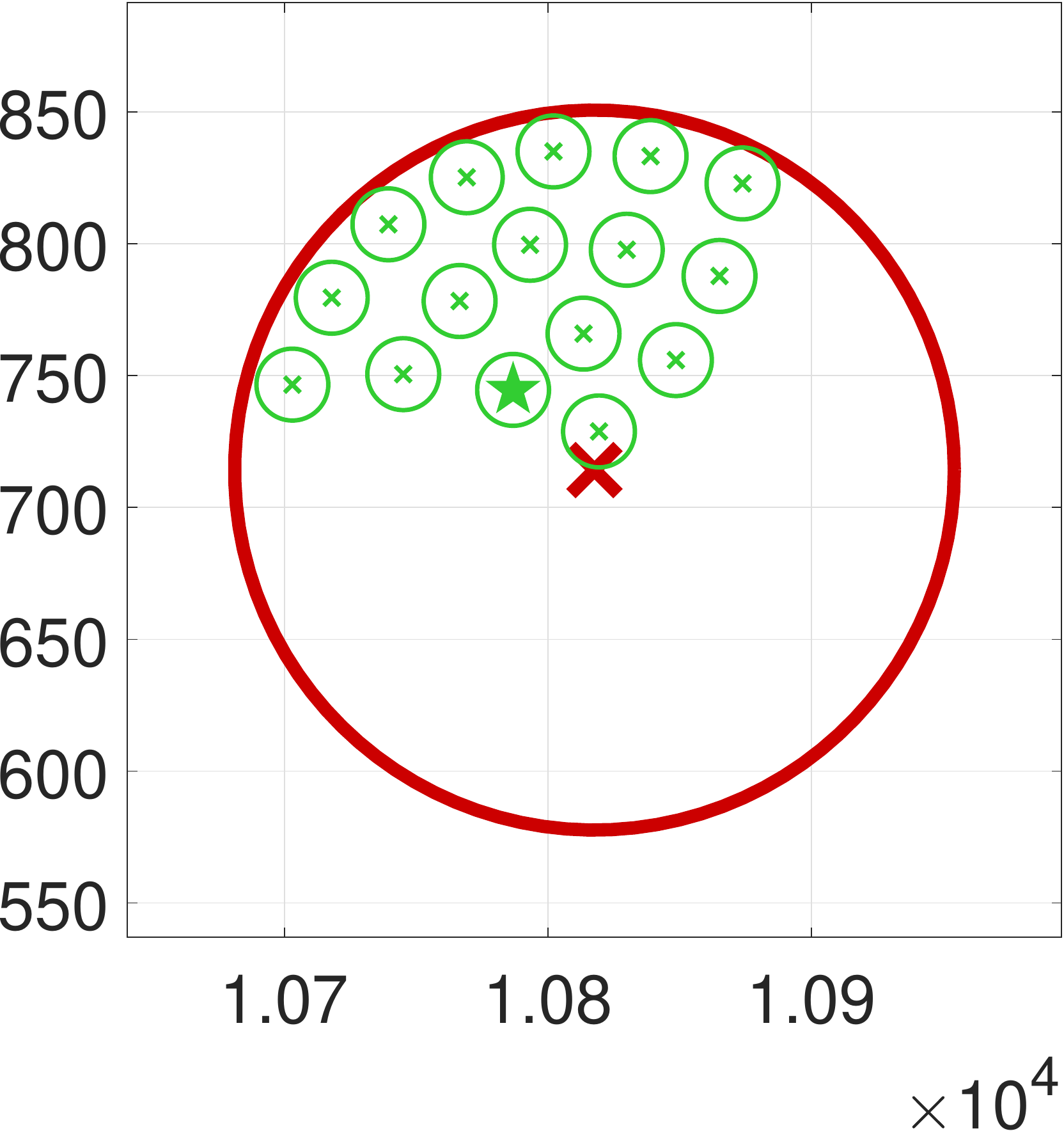}
         \caption{}
         \label{fig: VLMAS scale f}
     \end{subfigure}
               \hfill
          \begin{subfigure}[b]{0.16\textwidth}
         \centering
         \includegraphics[width=\textwidth, trim={0cm 0cm 0cm 0cm
         },clip]{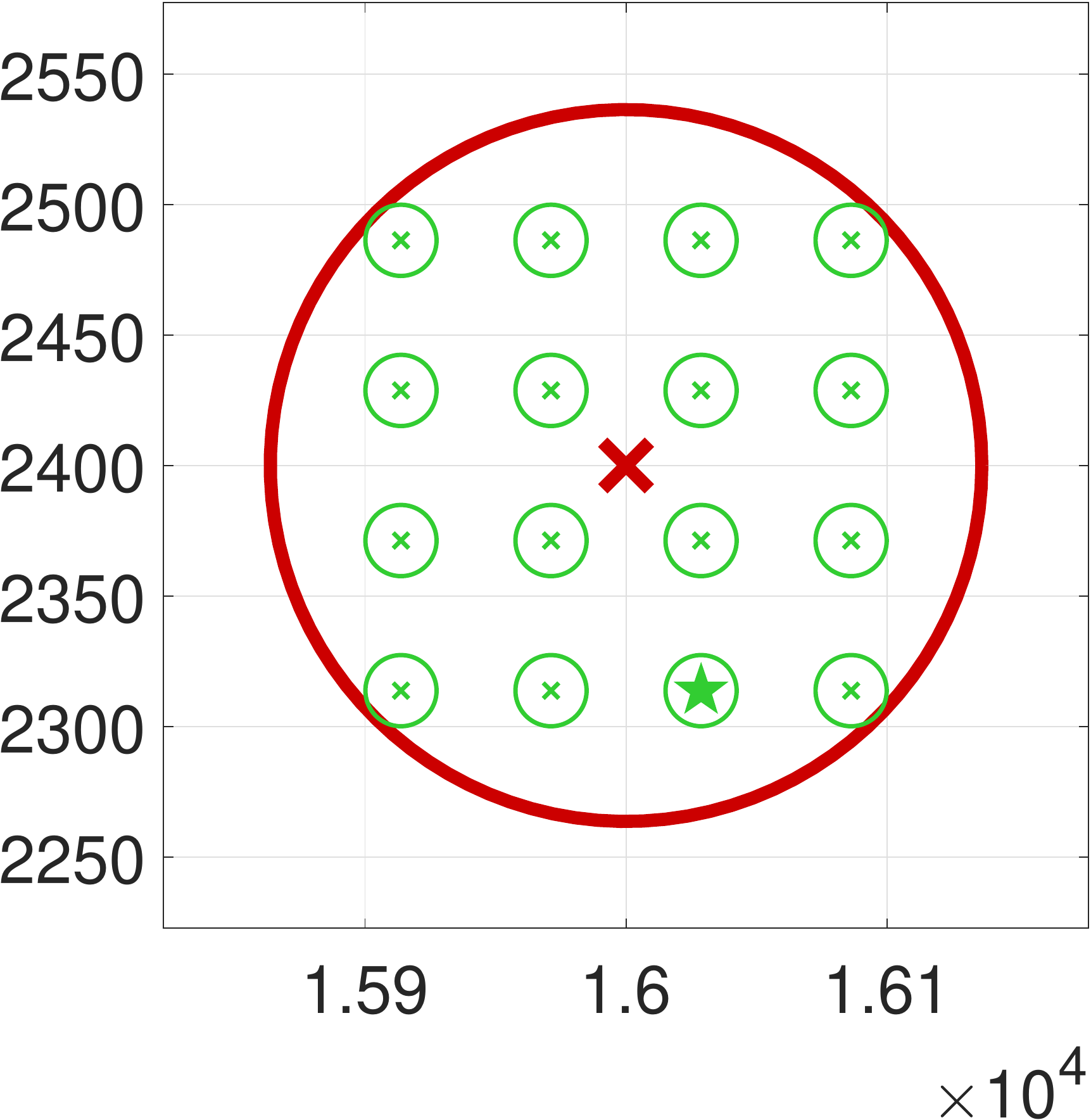}
         \caption{}
         \label{fig: VLMAS scale g}
     \end{subfigure}
       \centering
          \hfill
          \begin{subfigure}[b]{0.16\textwidth}
         \centering
         \includegraphics[width=\textwidth, trim={0cm 0cm 0cm 0cm
         },clip]{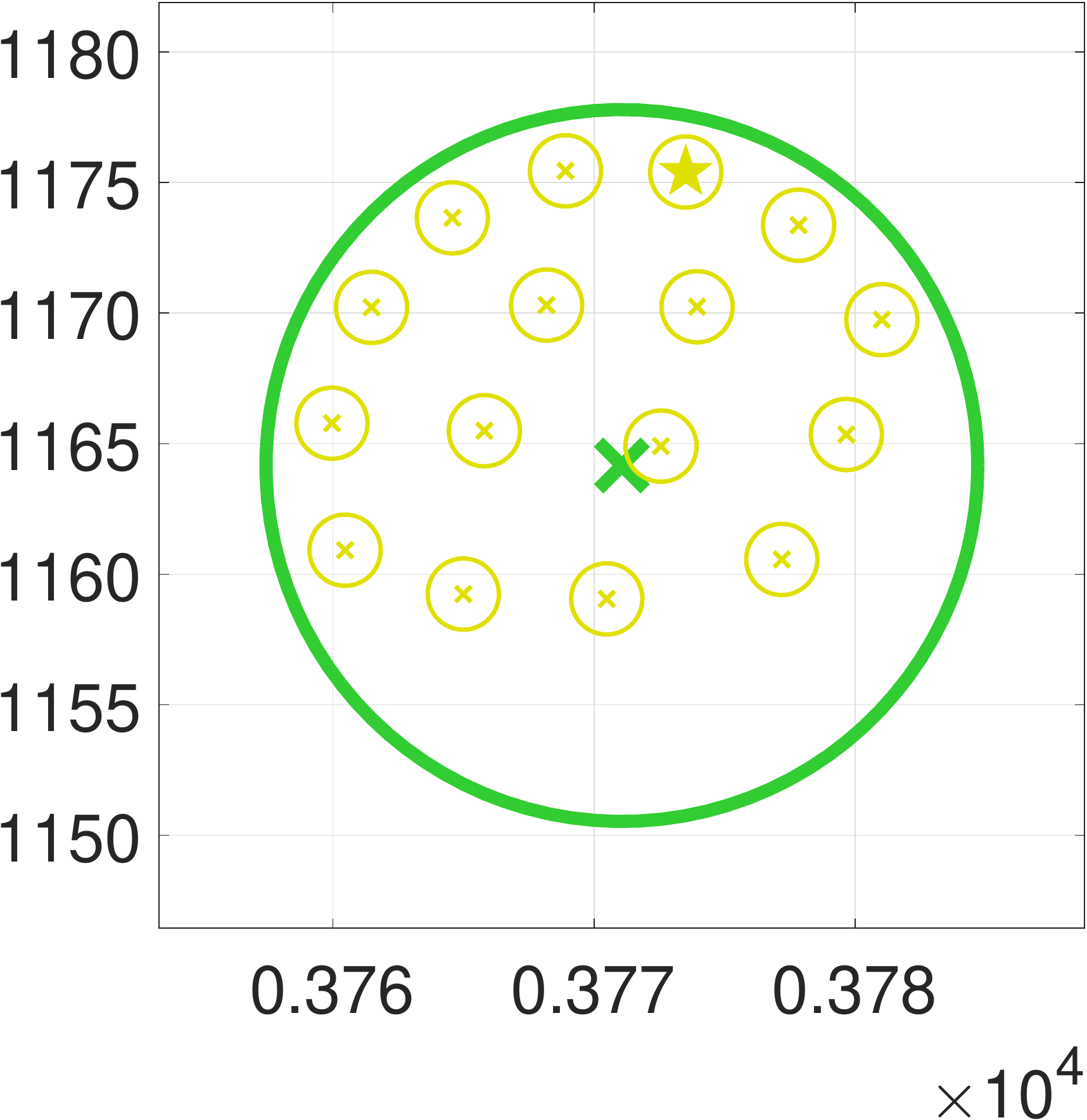}
         \caption{}
         \label{fig: VLMAS scale h}
     \end{subfigure}
            \centering
          \hfill
          \begin{subfigure}[b]{0.155\textwidth}
         \centering
         \includegraphics[width=\textwidth, trim={0cm 0cm 0cm 0cm
         },clip]{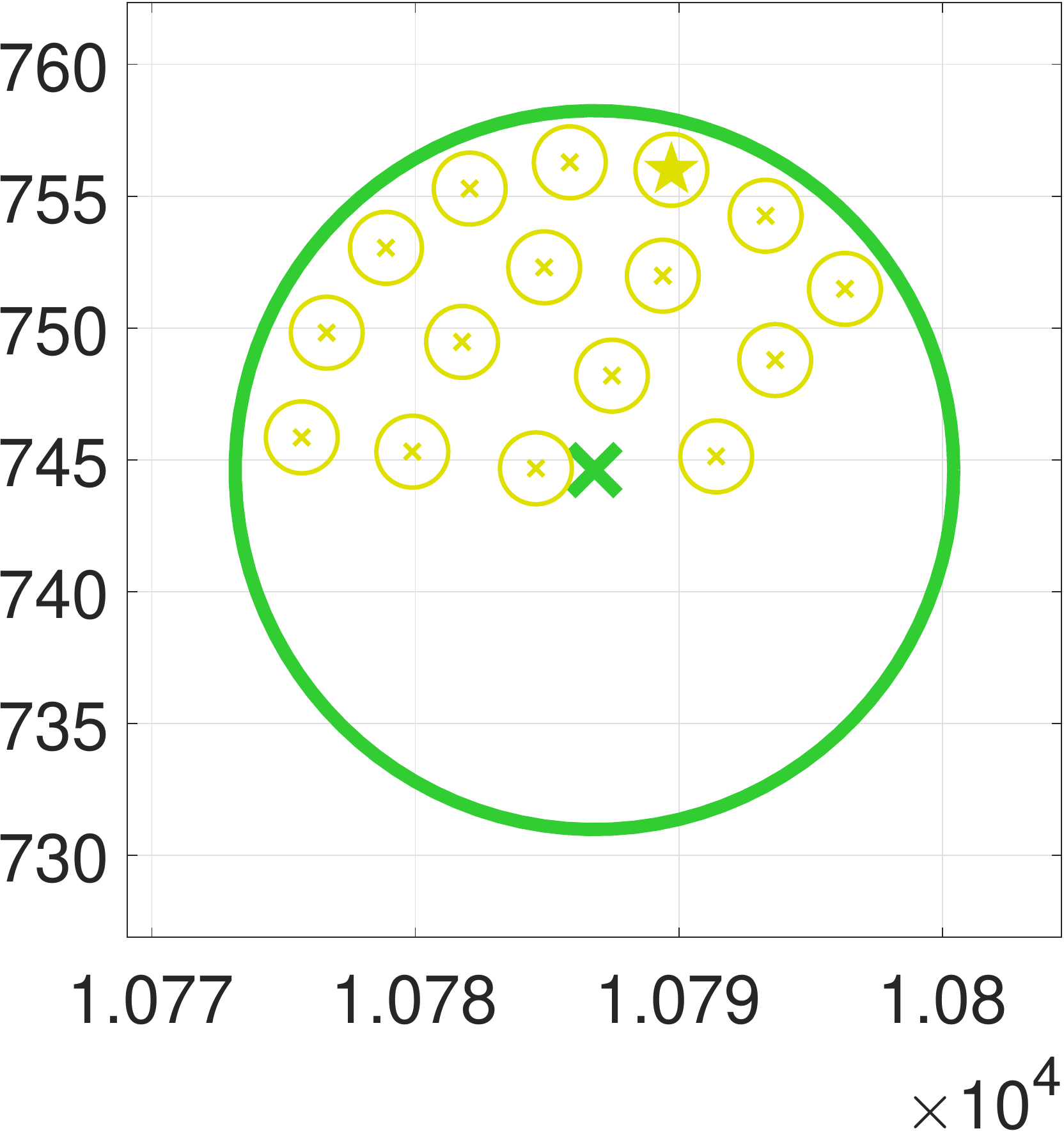}
         \caption{}
         \label{fig: VLMAS scale i}
     \end{subfigure}
            \centering
          \hfill
          \begin{subfigure}[b]{0.16\textwidth}
         \centering
         \includegraphics[width=\textwidth, trim={0cm 0cm 0cm 0cm
         },clip]{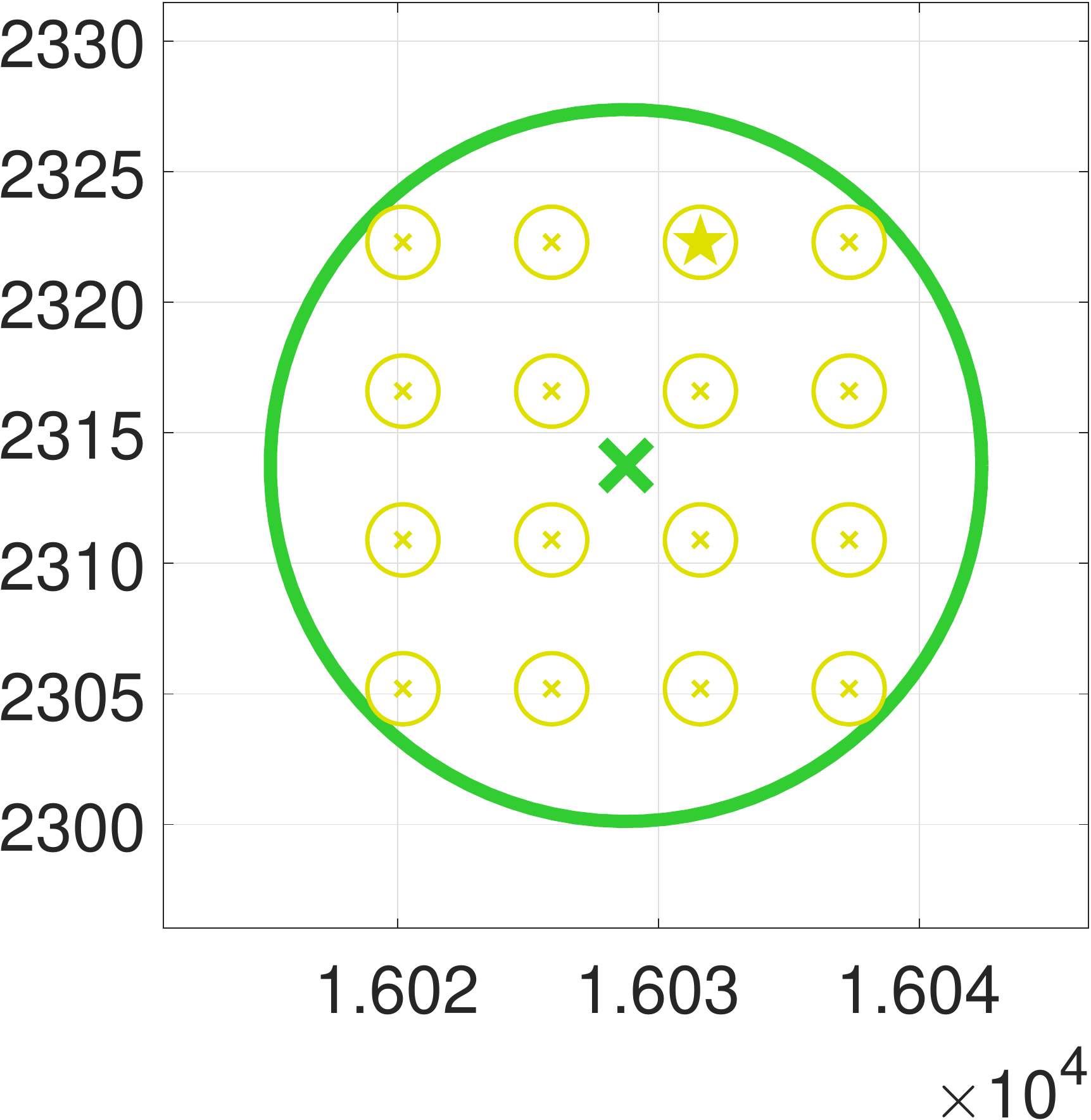}
         \caption{}
         \label{fig: VLMAS scale j}
     \end{subfigure}
     \\[0.1cm]
\centering
          \begin{subfigure}[b]{0.16\textwidth}
         \centering
         \includegraphics[width=\textwidth, trim={0cm 0cm 0cm 0cm
         },clip]{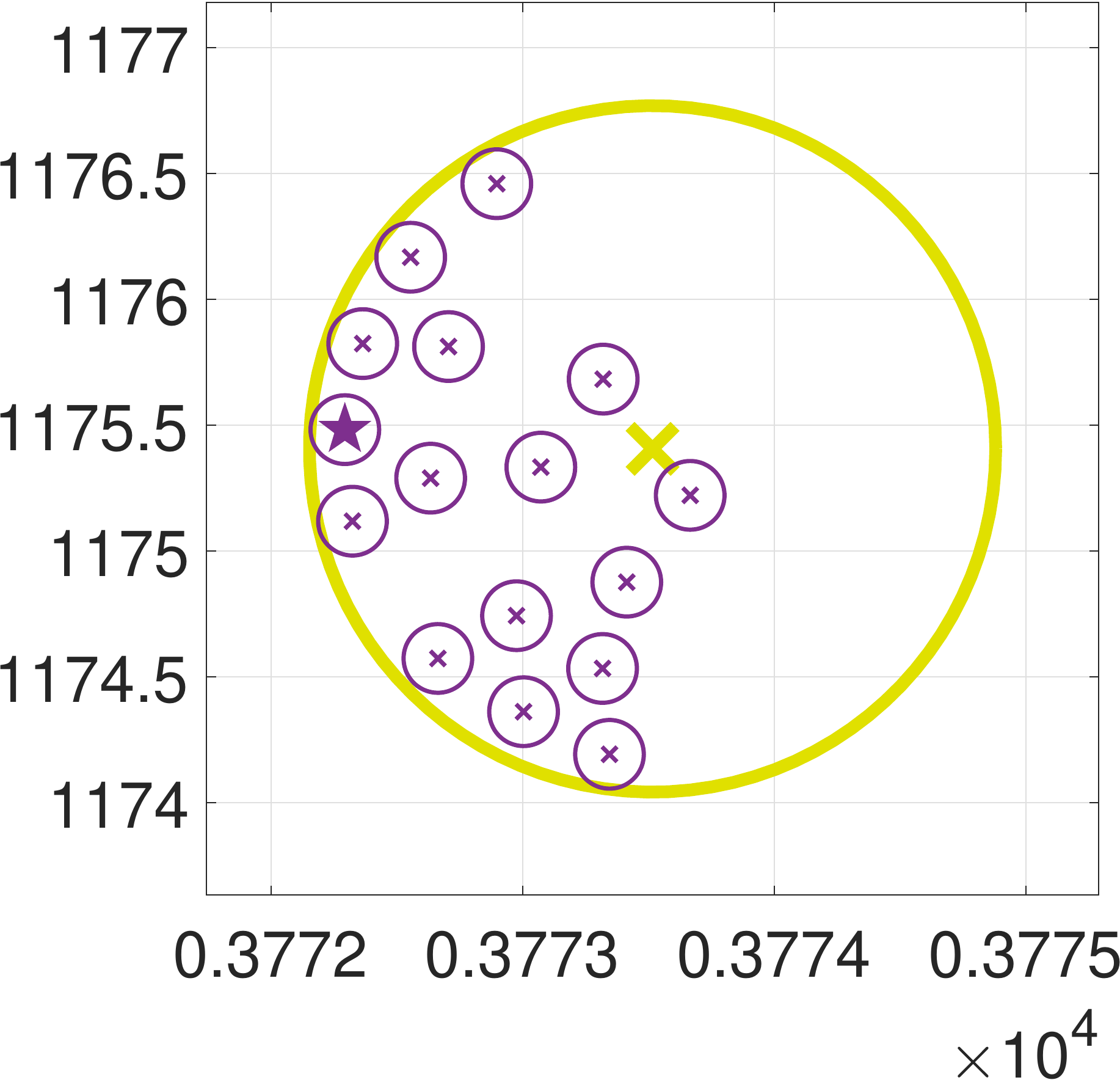}
         \caption{}
         \label{fig: VLMAS scale k}
     \end{subfigure}
          \centering
          \hfill
          \begin{subfigure}[b]{0.155\textwidth}
         \centering
         \includegraphics[width=\textwidth, trim={0cm 0cm 0cm 0cm
         },clip]{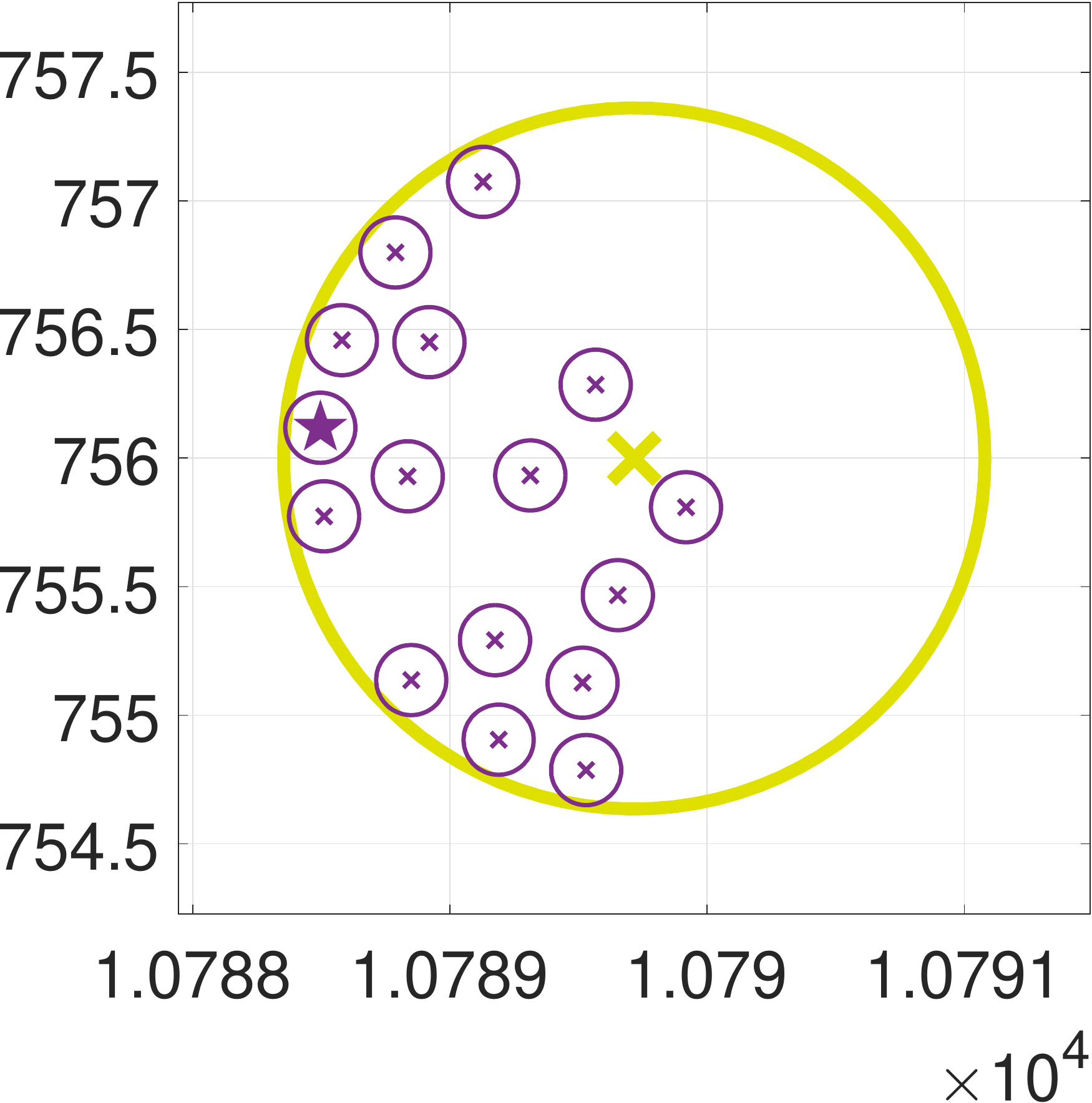}
         \caption{}
         \label{fig: VLMAS scale l}
     \end{subfigure}
               \hfill
          \begin{subfigure}[b]{0.16\textwidth}
         \centering
         \includegraphics[width=\textwidth, trim={0cm 0cm 0cm 0cm
         },clip]{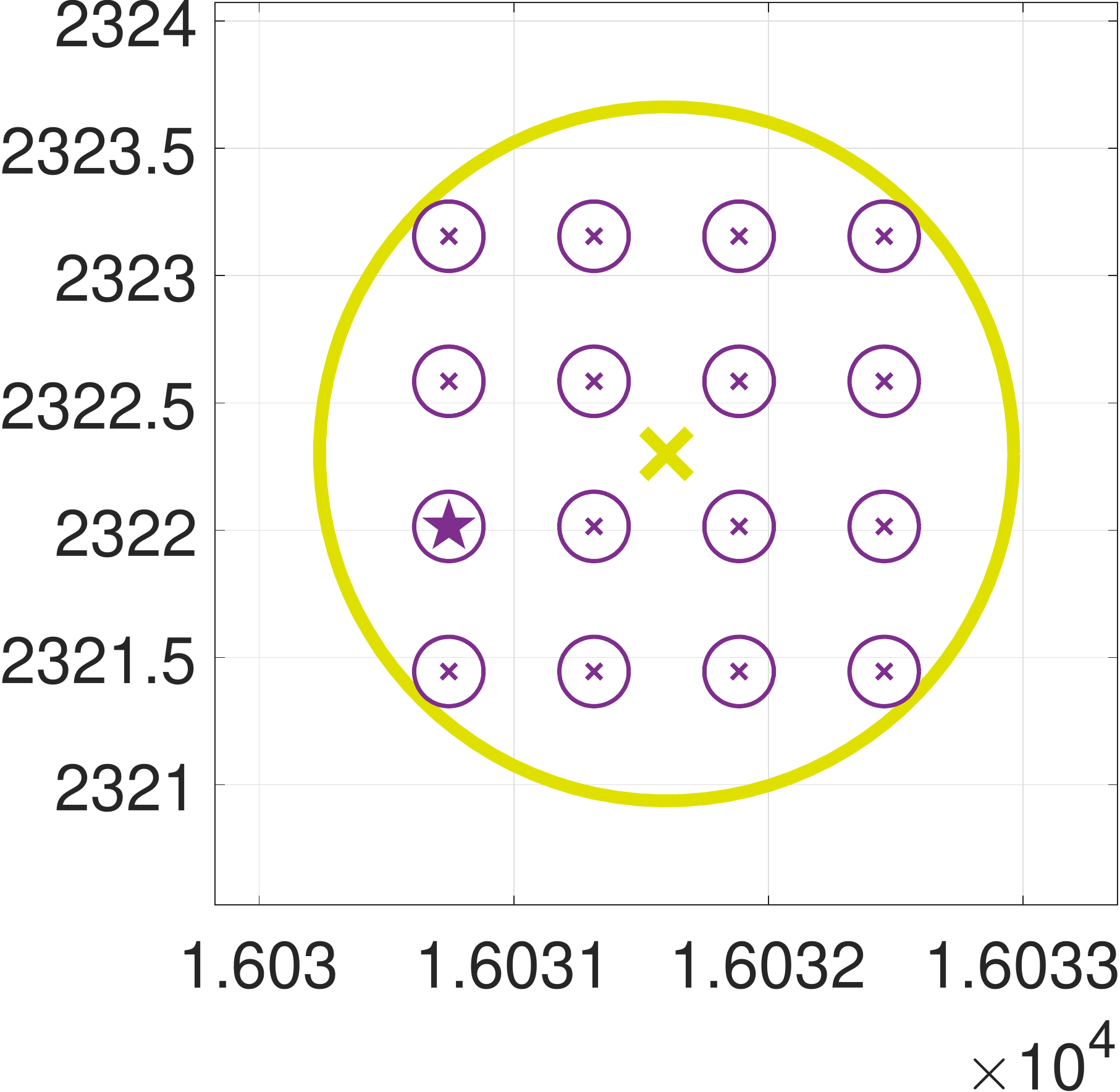}
         \caption{}
         \label{fig: VLMAS scale m}
     \end{subfigure}
       \centering
          \hfill
          \begin{subfigure}[b]{0.165\textwidth}
         \centering
         \includegraphics[width=\textwidth, trim={0cm 0cm 0cm 0cm
         },clip]{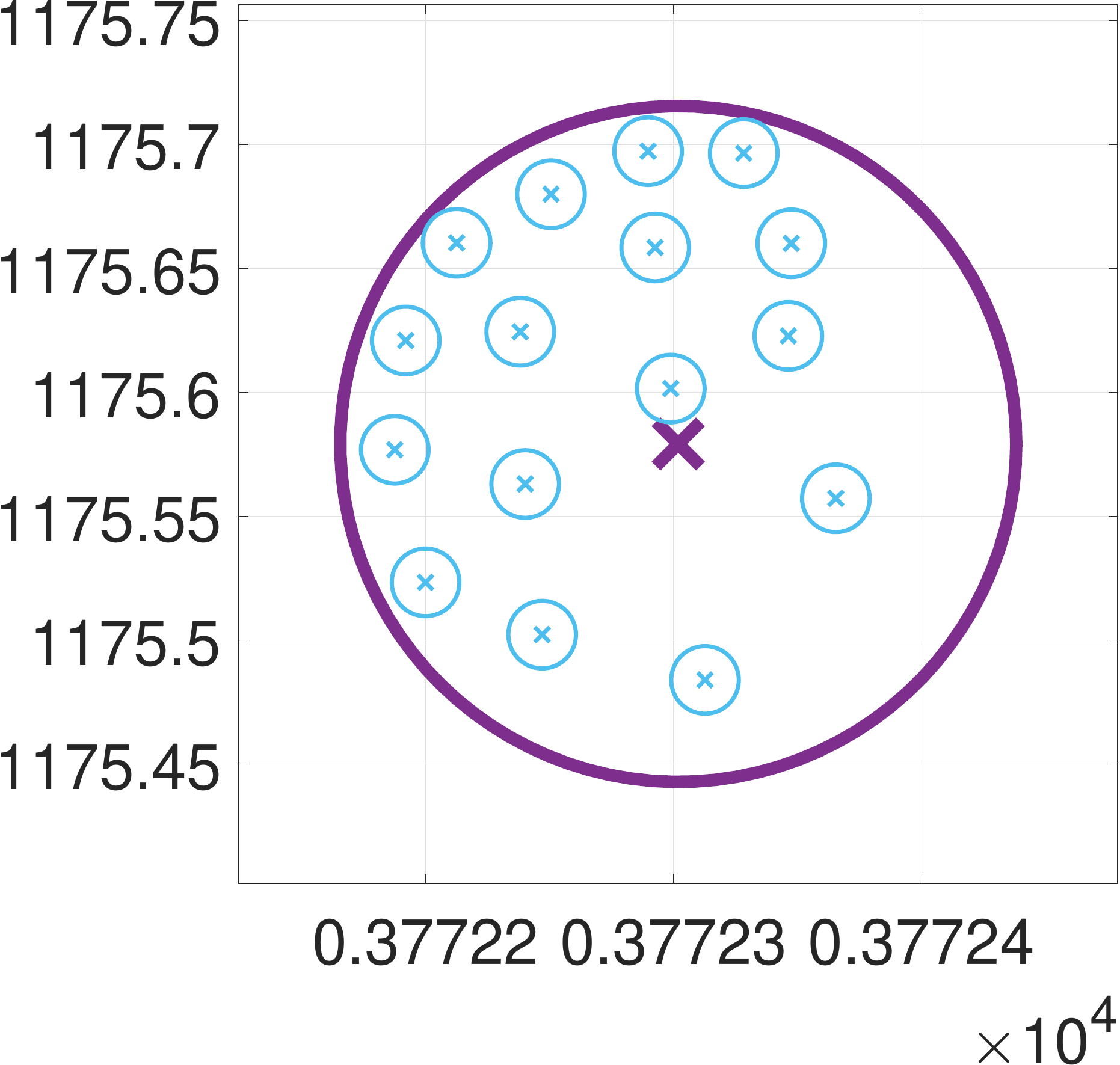}
         \caption{}
         \label{fig: VLMAS scale n}
     \end{subfigure}
            \centering
          \hfill
          \begin{subfigure}[b]{0.16\textwidth}
         \centering
         \includegraphics[width=\textwidth, trim={0cm 0cm 0cm 0cm
         },clip]{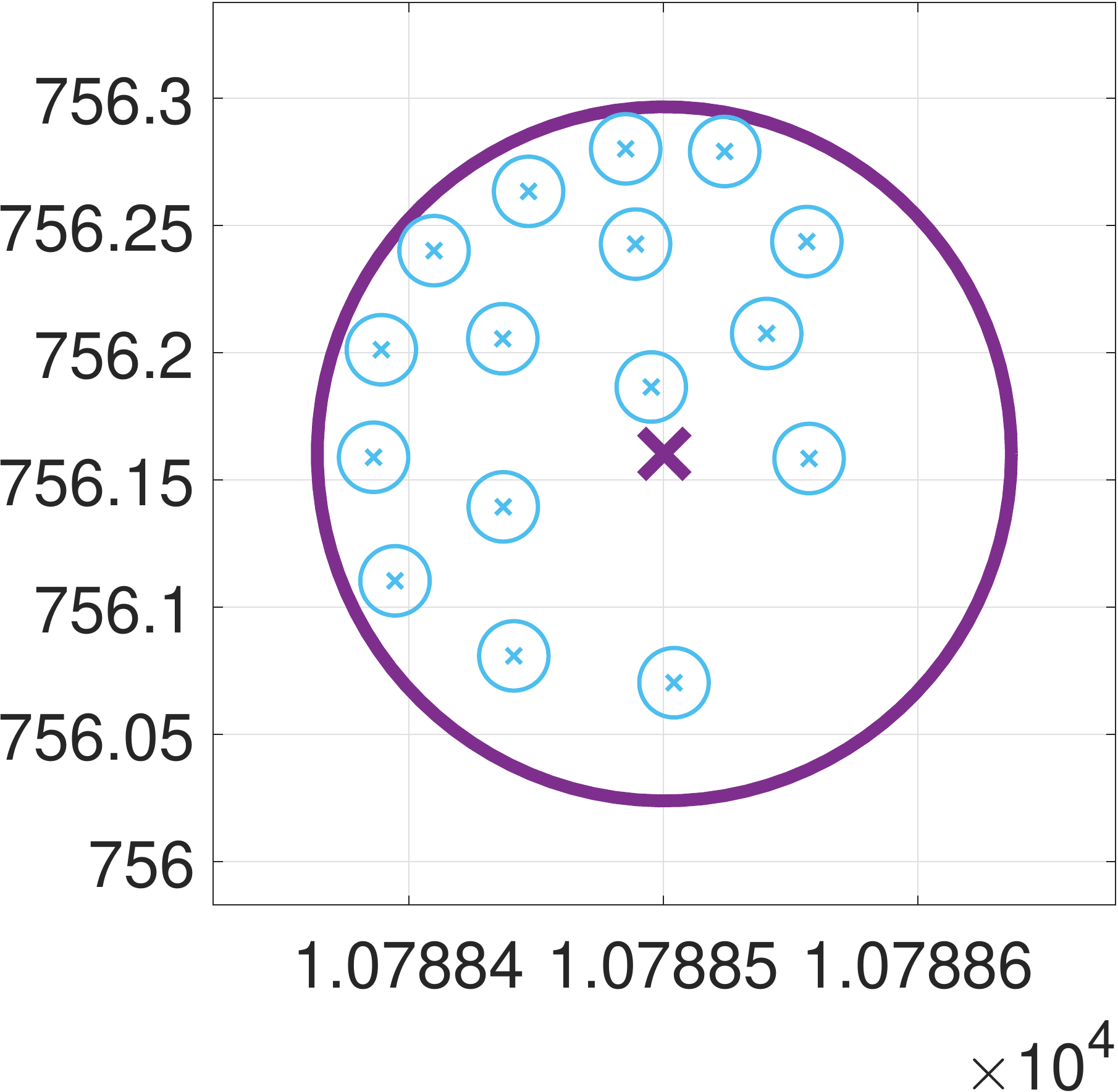}
         \caption{}
         \label{fig: VLMAS scale o}
     \end{subfigure}
            \centering
          \hfill
          \begin{subfigure}[b]{0.17\textwidth}
         \centering
         \includegraphics[width=\textwidth, trim={0cm 0cm 0cm 0cm
         },clip]{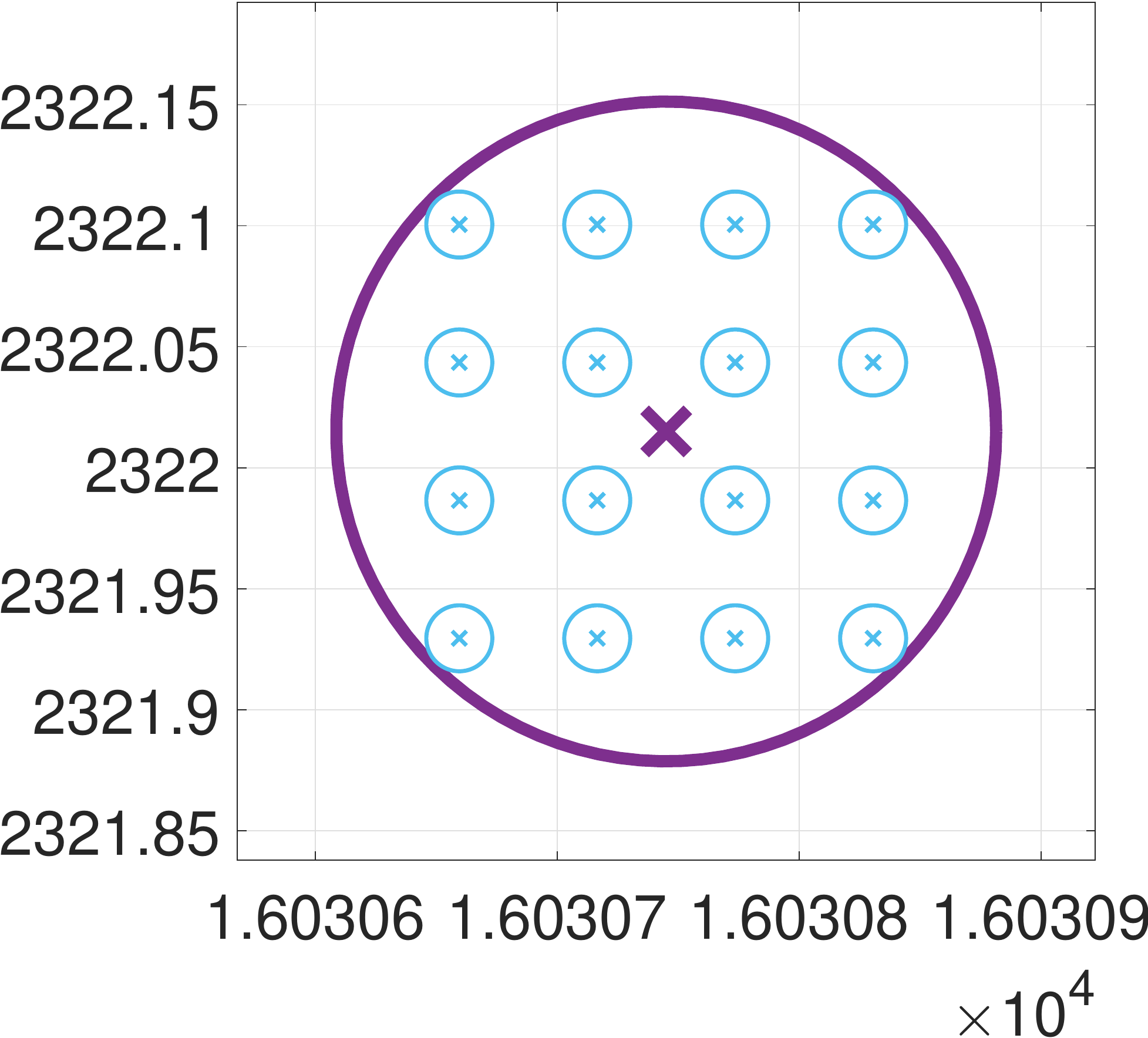}
         \caption{}
         \label{fig: VLMAS scale p}
     \end{subfigure}
        \caption{\textbf{A six-level VLMAS scenario.} Each level corresponds to a specific color ($1$: blue, $2$: red, $3$: green, $4$: yellow, $5$: purple, $6$: cyan). Fig. (a)-(d) show levels $1$ and $2$ at time instants $k=30,60,80,100$, respectively. The initial (left) and target (right) distributions are shown with dotted ellipses. Fig. (e)-(g) display one specific level-$2$ clique and its level-$3$ subcliques at $k=30,60,100$. Similarly, Fig. (h)-(j) show one specific level-$3$ clique and its level-$4$ subcliques, Fig. (k)-(m) show one specific level-$4$ clique and its level-$5$ subcliques, and Fig. (n)-(p) show one specific level-$4$ clique and its level-$6$ subcliques, for the same time instants. Each clique we are ``zooming`` into, is shown with a star in the figures of the above level.}
        \label{fig: VLMAS scale}
\end{figure*}


\section{Simulation Results}

This section provides simulation experiments that verify the effectiveness and scalability of the proposed DHDC framework. In Section \ref{subsec: small-scale scenario}, we consider a two-level small-scale system to facilitate the exposition of how DHDE and DHDS operate. A larger three-level system of agents is then used in Section \ref{subsec: large scenario} to further illustrate how the two proposed sub-frameworks work. Subsequently, in Section \ref{subsec: vlmas scenario}, we showcase the applicability of DHDC on a six-level VLMAS of two million robots. Finally, we compare the proposed approach against other CS methods in Section \ref{subsec: results comp}, and illustrate its superior computational efficiency and safety performance. The reader is encouraged to also refer to the supplementary video\footnote{\url{https://youtu.be/0QPyR4bD2q0}} for a full illustration of the results.

\subsection{Small-Scale Scenario}
\label{subsec: small-scale scenario}

It this task, we consider a two-level small-scale hierarchy with $5$ cliques of $4$ agents, thus a total of $20$ agents (Fig. \ref{fig: small scale}). All agents are modeled with 2D double integrator dynamics. The time horizon of the task is $N=100$. For additional details on the dynamics and algorithmic parameters, the reader is referred to Section \ref{SM sec: sim details} of the SM. The $99.7 \%$-confidence ellipses of the initial Gaussian distributions of the robots are shown in Fig. \ref{fig: small scale a} with red color. The level-$1$ distributions that are computed with DHDE also shown with blue color. As expected, the confidence ellipses of the assigned distributions for the level-$1$ cliques encompass the ones of the distributions of the robots (level $2$). Similarly, the target distributions of all levels are illustrated in Fig. \ref{fig: small scale b}. 

The goal of all agents is to reach to a target distribution that is in diametrically opposite position to their initial one, while of course avoiding collisions with the rest of the agents. Figures \ref{fig: small scale c}-\ref{fig: small scale h} show snapshots of the motion of all distributions for time instants $k=0,20,30,40,50,60,70,100$, respectively. As illustrated, DHDS is able to successfully steer all the level-$1$ distributions to the targets provided by DHDE. In the meantime, the distributions of all agents are also steered to their targets while satisfying the constraint that forces their distributions to lie within their parent clique distributions. Thanks to the fact that the solution of the level-$1$ problems guarantees that the agents of one clique will not collide with the ones of other cliques, the level-$2$ subproblems only take into account collision avoidance constraints with agents form the same clique.  In Fig. \ref{fig: small scale h}, all level-$1$ and level-$2$ distributions have successfully reached to their targets. 

\subsection{Large-Scale Scenario}
\label{subsec: large scenario}

Next, a larger scale $3$-level system of $2 \times 8 \times 9 = 144$ agents (Fig. \ref{fig: mid scale}) is used to further exhibit the effectiveness of the proposed method. First, we focus on illustrating the performance of the DHDE sub-framework. Figure \ref{fig: mid scale a} shows the initial and target distributions of the robots (level $3$) and the ones corresponding to the level-$2$ cliques that are obtained through DHDE. In Fig. \ref{fig: mid scale b}, the level-$1$ distributions are shown, while Fig. \ref{fig: mid scale c} shows the results DHDE would provide if the inter-clique constraints \eqref{multi clique est problem: ellipses overlap constraint} had been omitted. This highlights the importance of including these constraints and of using the advanced Problem \ref{estimation problem multi clique} formulation in our setup instead of the more simplistic Problem \ref{estimation problem single clique} one.

The performance of the DHDS algorithm is then demonstrated in Figs. \ref{fig: mid scale d}-\ref{fig: mid scale i}. In particular, the motion of the distributions of all levels is shown in Figs. \ref{fig: mid scale d}-\ref{fig: mid scale f} as they are being steered towards their target while avoiding the obstacles in the middle. In Figs. \ref{fig: mid scale g}-\ref{fig: mid scale i}, we focus into the black dotted box of the previous plots to further emphasize on the motion of the robots (level $3$). As shown, the distributions of the robots are steered while staying within the distributions of their parent cliques and not overlapping with each other.

\subsection{VLMAS Scenario}
\label{subsec: vlmas scenario}

Subsequently, we consider a VLMAS with a $6$-level hierarchical clustered structure, where the first level has $2$ cliques, each clique in levels $2,\dots,4$ contains $16$ sub-cliques, and finally each level-$5$ clique contains $16$ agents (level $6$). Therefore, this VLMAS consists of $2 \times 16^5 = 2,097,152$ agents. As in the previous tasks, the initial and target distributions of all cliques of all levels are first estimated, and subsequently, the actual distributions are steered while satisfying the probabilistic safety constraints for collision and obstacle avoidance. 

Figures \ref{fig: VLMAS scale a}-\ref{fig: VLMAS scale d} show the distributions of the cliques of levels $1$ and $2$. In particular, the $99.7\%$-confidence regions of the initial (left) and target (right) distributions are shown with dotted ellipses. Note that these distributions are the result of DHDE, first computing the level-$5$ distributions, then the level-$4$ ones, and so on - for a detailed demonstration, the reader is referred to the supplementary video. The motion of all level-$1$ (blue) and level-$2$ (red) distributions is shown in Figs. \ref{fig: VLMAS scale a}-\ref{fig: VLMAS scale d} for time instants $k=30,60,80,100$. As DHDS is a top-down framework, the level-$1$ distributions are first steered to their targets while successfully avoiding the obstacle. Subsequently, all level-$2$ clique distributions are also steered to their corresponding targets while staying within the limits of their parent cliques and avoiding collisions with each other. In Fig. \ref{fig: VLMAS scale d} all level-$1,2$ clique distributions have successfully reached to their targets.

In Figs. \ref{fig: VLMAS scale e}-\ref{fig: VLMAS scale g}, we focus into a randomly selected level-$2$ clique and illustrate the motion of the level-$3$ (green) distributions that belong in this clique, for time instants $k=30,60,100$, respectively. As shown, all level-$3$ cliques remain within the limits of their parent cliques, while also avoiding collisions. In Fig. \ref{fig: VLMAS scale g}, the level-3 cliques have reached to their target configuration. Similarly, Figs. \ref{fig: VLMAS scale h}-\ref{fig: VLMAS scale j} focus into a level-$3$ clique and illustrate the motion of all level-$4$ (yellow) sub-cliques that belong in this clique. Again, all cliques are able to safely reach to their target distributions. The motion of the distributions of specific cliques in levels $5$ (purple) and $6$ (cyan) are also shown in Figs. \ref{fig: VLMAS scale k}-\ref{fig: VLMAS scale m} and \ref{fig: VLMAS scale n}-\ref{fig: VLMAS scale p}, respectively. All distributions of all levels are successfully steered to their targets, which indicates that all agents have reached their target distributions.

\subsection{Comparison with Related Methods}
\label{subsec: results comp}

Finally, we highlight the significant computational efficiency and safety capabilities of DHDC by comparing it with the Centralized CS (CCS) \cite{balci2021covariance} and Distributed CS (DCS) \cite{saravanos2021distributed} methods. In CCS, the full level-$L$ CS problem is solved without any splitting being considered. In DCS, the level-$L$ CS problem is solved using an ADMM-based distributed approach, which however cannot benefit from the hierarchical structure of the VLMAS.

\subsubsection{Computational Demands}
\begin{figure}[!t]
\centering
\includegraphics[width=0.5\textwidth, trim={0cm 0cm 0cm 0cm},clip]{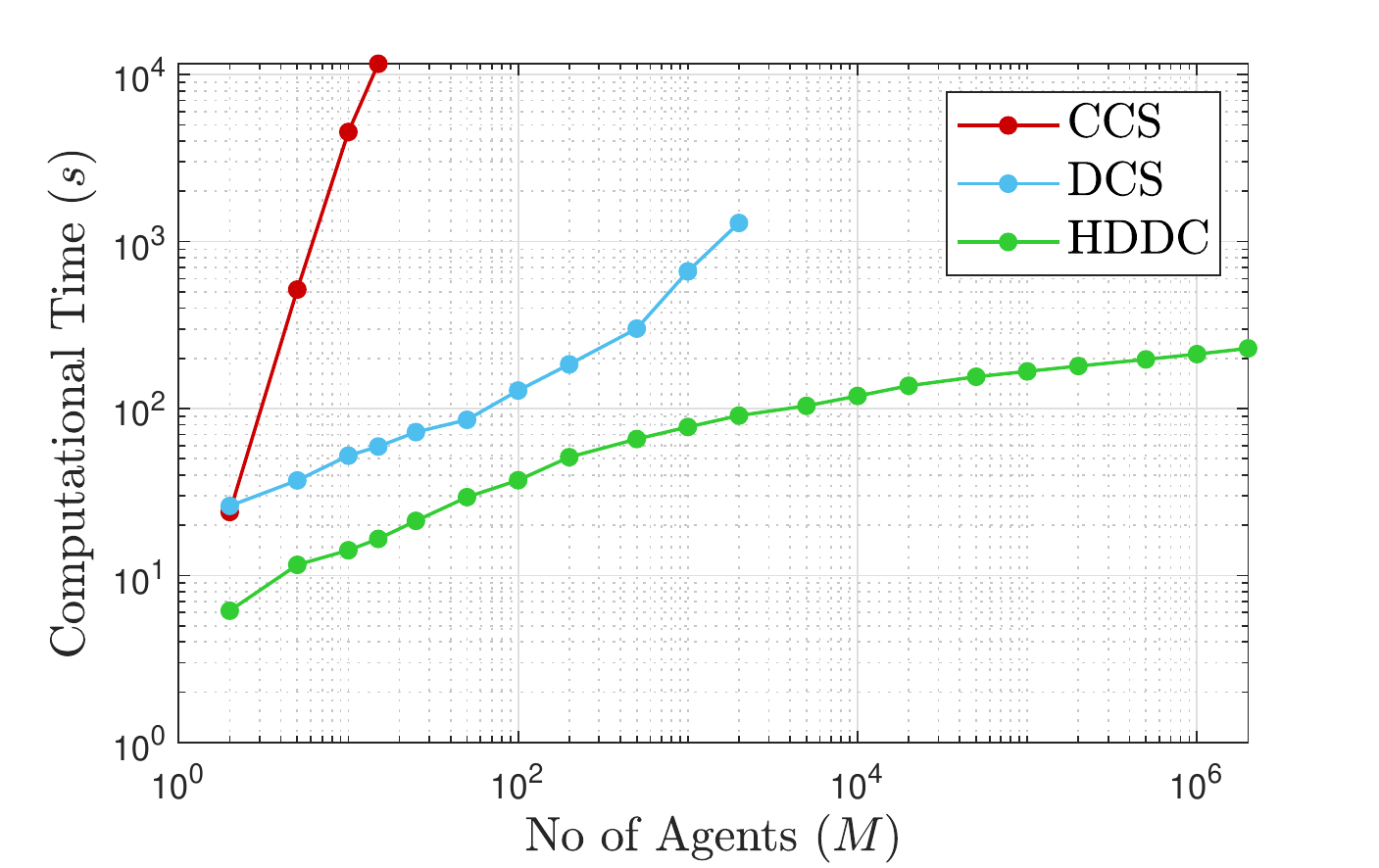}
\caption{Comparison of computational times between DHDC (proposed), CCS and DCS.}
\label{fig: comp times}
\end{figure}

 To perform a computational demands comparison, we repeat the task of Section \ref{subsec: vlmas scenario}, starting from small numbers of agents. All simulations were performed in Matlab R2021b using MOSEK 9.1.9 \cite{mosek} as the optimization solver and a laptop computer with an 11th Gen Intel(R) Core(TM) i7-11800H @ 2.30GHz and a 32GB RAM memory. The computational times for tasks ranging from two agents to two million agents, are shown in Fig. \ref{fig: comp times}. The results highlight the superior scalability of DHDC against DCS and CCS. This is mainly attributed to the following reasons: i) First, DHDC does not require solving problems of a significantly large scale given that there exist no cliques that contain a very large number of cliques in the level below them. ii) Second, the efficient variables splitting of the subproblems in DHDC reduces the amount of necessary copy variables, making DHDC significantly more computationally efficient than DCS. iii) Third, in CCS, the multi-agent problem is always solved in a centralized fashion, which leads to high-dimensional semidefinite programming problems that soon become computationally intractable. 
 
\subsubsection{Safety} 
\begin{figure}[!t]
\centering
\includegraphics[width=0.5\textwidth, trim={0cm 0cm 0cm 0cm},clip]{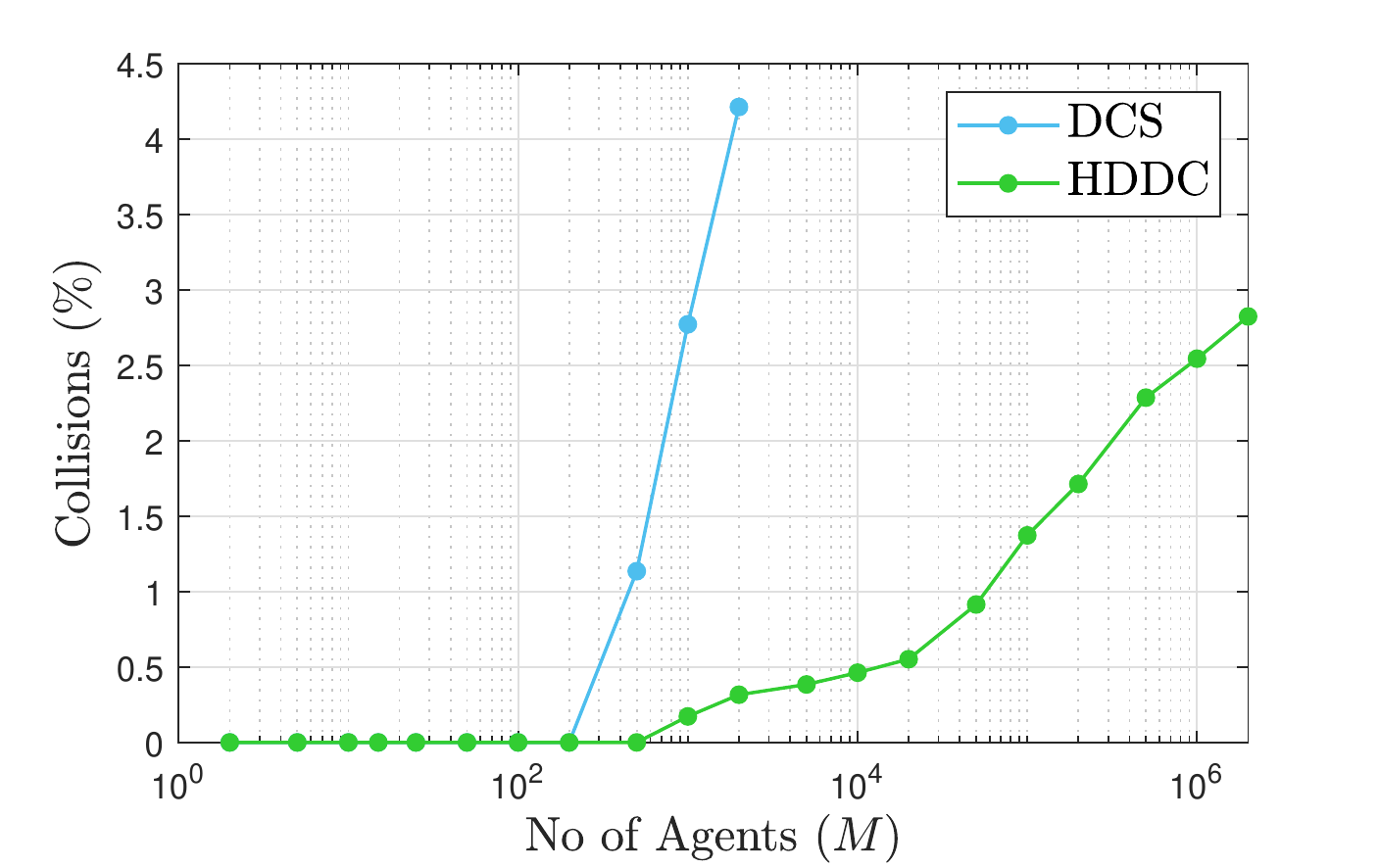}
\caption{Comparison of collision percentages between DHDC (proposed) and DCS.}
\label{fig: collisions}
\end{figure}

Next, we conduct a comparison on the safety capabilities of each method. Given that CCS cannot scale for large-scale systems, we exclude it from the comparison. In Fig. \ref{fig: collisions}, we demonstrate the collisions percentages for DHDC and DCS. As both methods enforce consensus through soft constraints, it follows that as tasks get more complicated, collisions might appear. Nevertheless, DHDC outperforms DCS in terms of safety as well, since the cliques/agents always need to only consider safety constraints associated with the other cliques/agents that are within the same clique, thus making the optimization problems easier to solve.

\section{Conclusion} 
\label{sec: conclusion}

This paper proposes a scalable hierarchical distributed control framework named DHDC for the control of VLMAS that admit a multi-level hierarchical clustered structure.  The first part of the framework, DHDE, associates the initial and target configurations of the cliques of all levels to representative Gaussian distributions that satisfy all the requirements of the hierarchical structure. The second part, DHDS, steers the distributions of all cliques and agents to their prescribed target distributions. Simulation experiments demonstrate the scalability of DHDC to VLMAS with up to two million robots. Therefore, DHDC is shown to be able to control systems of a very-large scale by exploiting the control of distributions within hierarchical structures.

Future work will focus on further expanding DHDC for more general problem setups. A straightforward extension would be to extend the framework for systems nonlinear dynamics by incorporating nonlinear versions of CS \cite{ridderhof2019nonlinear, saravanos2022dmpcs}. In addition, we wish to extend DHDC for the case where the hierarchical structure is unknown by exploring the incorporation of hierarchical distribution alignment methods in unsupervised learning such as \cite{lee2019hierarchical, yurochkin2019hierarchical}. Finally, future work will also focus on establishing formal convergence guarantees for the proposed method.

\section*{Acknowledgments}
This research was supported by ARO contract W911NF2010151 and  NSF CMMI-1936079. Augustinos Saravanos acknowledges financial support by the A. Onassis Foundation Scholarship. 


\bibliographystyle{plainnat}
\bibliography{references}

\newpage
\section*{Supplementary Material}

The following part serves as supplementary material including the proofs and additional details that are not covered in the main paper.

\section{DHDE Details}

\subsection{Proof of Proposition \ref{single clique proposition}}
\label{SM sec: proposition 1 proof}
First, it is straightforward to show the equivalence between minimizing the costs in \eqref{single clique problem: cost} and \eqref{single clique prop: cost}, since $\KL ( \calN_n^{\ell+1} \| \calN_i^{\ell} )$ is given by 
\begin{align}
& \KL ( \calN_n^{\ell+1} \| \calN_i^{\ell} ) = 
\frac{1}{2} \bigg[ \log \frac{|\Sigma_i|}{|\Sigma_n|} - n_x + \mathrm{tr}(\Sigma_i^{-1} \Sigma_n) 
\\
& ~~~~~~~~~~~~~~~~~~~~~~~~~~~~~~~~~~~~ + (\mu_i - \mu_n)^\T \Sigma_i^{-1} (\mu_i - \mu_n) \bigg]
\nonumber
\end{align}
which yields $\hat{J}_i^{\mathrm{e}}(Q_i,q_i)$ after substituting with $Q_i = \Sigma_i^{-1}$, $q_i = \Sigma_i^{-1} \mu_i$ and neglecting the constant terms. Note that the objective function $\hat{J}_i^{\mathrm{e}}(Q_i,q_i)$ is jointly convex for $Q_i,q_i$. 

%

Next, we show the equivalence between the constraints \eqref{single clique problem: ellipses constraint} and \eqref{single clique prop: ellipses constraint 1}-\eqref{single clique prop: ellipses constraint 2}. For the convenience of the reader, let us first restate a result known as the S-Lemma by \citet{yakubovich1977s}. According to the S-Lemma, given two functions $f_1,f_2: \Rb^n \rightarrow \Rb$ with $f_1(x) = x^\T A_1 x + b_1^\T x + c_1$ and $f_2(x) = x^\T A_2 x + b_2^\T x + c_2$ and if there exists an $\bar{x}$ such that $f_1(\bar{x}) > 0$, then the following is true
\begin{equation}
f_1(x) \geq 0 \Rightarrow f_2(x) \geq 0, \quad  \forall x,
\end{equation}
if and only if there exists a $\tau \geq 0$ such that $f_2(x) \geq \tau f_1(x), \ \forall x$. In constraint \eqref{single clique problem: ellipses constraint}, we enforce that if $x \in \Rb^{n_p}$ is such that 
\begin{equation}
(x - \bar{\mu}_n)^\T \bar{\Sigma}_n^{-1} (x - \bar{\mu}_n)
\leq \alpha, 
\end{equation}
then it should follow that 
\begin{equation}
(x - \bar{\mu}_i)^\T \bar{\Sigma}_i^{-1} (x - \bar{\mu}_i)
\leq \alpha.
\end{equation}
Using the S-Lemma, this is equivalent with imposing the constraints $\tau_n \geq 0, \ \forall n \in \calC_i^{\ell}$ and
\begin{equation*}
\alpha - (x - \bar{\mu}_i)^\T \bar{\Sigma}_i^{-1} (x - \bar{\mu}_i)
\geq 
\tau \big( \alpha - 
(x - \bar{\mu}_n)^\T \bar{\Sigma}_n^{-1} (x - \bar{\mu}_n) \big),
\end{equation*}
which can be written in matrix form as
\begin{equation}
\hat{x}^\T V_n \hat{x} \geq 0,
\label{matrix form}
\end{equation}
where $\hat{x} = [x; 1]$ and 
\begin{equation*}
V_n = 
\begin{bmatrix}
V_{11} & V_{12}
 \\
V_{12}^\T
& 
V_{22}
\end{bmatrix},
\end{equation*}
with
\begin{align*}
V_{11} & = - \bar{\Sigma}_i^{-1} + \tau_n \bar{\Sigma}_n^{-1}, \  
V_{12} = \bar{\Sigma}_i^{-1} \bar{\mu}_i - \tau_n \bar{\Sigma}_n^{-1} \bar{\mu}_n,
\\
V_{22} & = \alpha - \tau_n \alpha - \bar{\mu}_i^\T \bar{\Sigma}_i^{-1} \bar{\mu}_i + \tau_n \bar{\mu}_n^\T \bar{\Sigma}_n^{-1} \bar{\mu}_n.
\nonumber
\end{align*}
%
By definition, the constraint \eqref{matrix form} is equivalent with $V_n \succeq 0$. Furthermore, by applying the Schur complement w.r.t. $V_{22}$, it follows that $V_n \succeq 0$ is equivalent with $S_n \succeq 0$, where 
\begin{equation}
S_n = 
\begin{bmatrix}
S_{11} & S_{12} & 0
\\
S_{12}^\T & S_{22} & S_{23}
\\
0^\T & S_{23}^\T & S_{33}
\end{bmatrix},
\end{equation}
with
\begin{align*}
S_{11} & = V_{11}, \ S_{12} = V_{12},
\ S_{22} = \alpha - \tau_n \alpha + \tau_n \bar{\mu}_n^\T \bar{\Sigma}_n^{-1} \bar{\mu}_n,
\\
 \ 
S_{23} & = (\bar{\Sigma}_i^{-1} \bar{\mu}_i)^\T, \ S_{33} = \bar{\Sigma}_i^{-1}.
\nonumber
\end{align*}
The expressions in \eqref{single clique prop: S terms} follow after substituting with $Q_i$ and $q_i$. Finally, it is evident that the constraints \eqref{single clique problem: semidef constraint} and \eqref{single clique prop: semidef constraint} are equivalent since $\Sigma_i \succ 0$ if and only if $\Sigma_i^{-1} \succ 0$. Note that all constraints are convex as well. Therefore, the problem presented in Proposition \ref{single clique proposition} is a convex optimization one.

\subsection{Proof of Proposition \ref{multi clique proposition}}
\label{SM sec: proposition 2 proof}
The equivalence between the costs \eqref{multi clique est problem: cost} and \eqref{multi clique prop: cost}, as well as the equivalence between the constraints \eqref{multi clique est problem: ellipses constraint} and \eqref{multi clique prop: ellipses constraint 1}-\eqref{multi clique prop: ellipses constraint 2} follow directly from Proposition \ref{single clique proposition}. Next, we show that if the constraints \eqref{multi clique prop: ellipses overlap constraint 1} and \eqref{multi clique prop: ellipses overlap constraint 2} are satisfied, then the constraint \eqref{multi clique est problem: ellipses overlap constraint} is also satisfied.  In fact, the constraint 
\begin{equation}
\calE_\theta [\bar{\mu}_i, \bar{\Sigma}_i] \cap \calE_\theta [\bar{\mu}_j, \bar{\Sigma}_j] = \emptyset
\end{equation}
will hold if the following constraint holds 
\begin{equation}
\label{multi clique prop proof: circles constraint}
\calbC \big[ \calE_\theta [ \bar{\mu}_i, \bar{\Sigma}_i] \big] \cap \calbC \big[ \calE_\theta [\bar{\mu}_j, \bar{\Sigma}_j] \big] = \emptyset,
\end{equation}
where $\calbC[\calE]$ denotes the minimum area enclosing circle of an ellipse $\calE$. Of course, $\calbC \big[ \calE_\theta [ \bar{\mu}_i, \bar{\Sigma}_i] \big]$ is a circle with center $\bar{\mu}_i$ and radius $\sqrt{\alpha \lambda_{\max}(\bar{\Sigma}_i)}$, which is the major axis length of $\calE_\theta [\bar{\mu}_i, \bar{\Sigma}_i]$. Hence, the constraint \eqref{multi clique prop proof: circles constraint} can be rewritten as 
\begin{equation}
\| \bar{\mu}_i - \bar{\mu}_j \|_2 \geq \sqrt{\alpha \lambda_{\max}(\bar{\Sigma}_i)} + \sqrt{\alpha \lambda_{\max}(\bar{\Sigma}_j)}.
\end{equation}
or equivalently as 
\begin{equation}
\label{multi clique prop proof: circles constraint 2}
\| \bar{Q}_i^{-1} \bar{q}_i - \bar{Q}_j^{-1} \bar{q}_j \|_2 \geq \frac{\sqrt{\alpha}}{\sqrt{\lambda_{\min}(\bar{Q}_i)}} + \frac{\sqrt{\alpha}}{\sqrt{\lambda_{\min}(\bar{Q}_j)}}.
\end{equation}
By introducing the auxiliary variables $\phi_i, \phi_j$, the constraint \eqref{multi clique prop proof: circles constraint 2} is equivalent with the set of constraints 
\begin{subequations}
\begin{align}
& \| \bar{Q}_i^{-1} \bar{q}_i - \bar{Q}_j^{-1} \bar{q}_j \|_2 \geq \phi_i^{-1/2} + \phi_j^{-1/2}, 
\label{multi clique prop proof: circles constraint 3a}
\\
& \phi_l^{-1/2} \geq \frac{\sqrt{\alpha}}{\sqrt{\lambda_{\min}(\bar{Q}_l)}}, \ l \in \{i,j\}
\label{multi clique prop proof: circles constraint 3b}
\end{align}
\end{subequations}
where \eqref{multi clique prop proof: circles constraint 3a} is the same as \eqref{multi clique prop: ellipses overlap constraint 1}. The constraint \eqref{multi clique prop proof: circles constraint 3b} can be rewritten as 
\begin{equation} 
\bar{Q}_l \succeq \phi_l \alpha I
\end{equation}
which yields \eqref{multi clique prop: ellipses overlap constraint 2}. Finally, the constraints \eqref{multi clique est problem: semidef constraint} and \eqref{multi clique prop: semidef constraint} are equivalent.

\subsection{ADMM Derivation}
\label{SM sec: DHDE admm}

After introducing the augmented variables $\tilde{Q}_i$, $\tilde{q}_i$, $\tilde{\phi}_i$, and the global ones $G$, $g$, $z$, the problem presented in Proposition \ref{multi clique proposition} can be reformulated as
\begin{subequations}
\label{estimation: admm proof align}
\begin{align}
& ~~~~~~
\min \sum_{i \in \calC_a^{\ell-1}} \hat{J}_i^{\mathrm{e}}(Q_i, q_i)
\\[0.2cm]
\mathrm{s.t} \quad 
& S_{i,n}(Q_i, q_i, \tau_{i,n}) \succeq
0, \ n \in \calC_i^{\ell},
\label{estimation: admm proof constraint 1}
\\
& \tau_{i,n} \geq 0, \ n \in \calC_i^{\ell},
\label{estimation: admm proof constraint 2}
\\
& h_i(\tilde{Q}_i, \tilde{q}_i, \tilde{\phi}_i) \leq 0,
\label{estimation: admm proof constraint 3}
\\
& T_i(Q_i, \phi_i) \succeq 0,
\label{estimation: admm proof constraint 4}
\\ 
& Q_i \succeq 0, 
\label{estimation: admm proof constraint 5}
\\ 
& \tilde{Q}_i = \tilde{G}_i, \ \tilde{q}_i = \tilde{g}_i, \ \tilde{\phi}_i = \tilde{z}_i, \ i \in \calC_a^{\ell-1},
\end{align}
\end{subequations}
where $h_i(\tilde{Q}_i, \tilde{q}_i, \tilde{\phi}_i)$ is defined as 
\begin{equation}
h_i = [\{ h_{i,j}(Q_i, q_i, \phi_i, Q_j^i, q_j^i, \phi_j^i) \}_{j \in \caln[\calC_i^\ell]}].
\end{equation}
Let us also introduce the indicator functions $\calI_{S_i}(Q_i, q_i, \tau_{i,n})$, $\calI_{\tau_{i,n}}(\tau_{i,n})$, $\calI_{h_i}(\tilde{Q}_i, \tilde{q}_i, \tilde{\phi}_i)$, $\calI_{T_i}(Q_i, \phi_i)$, $\calI_{Q_i}(Q_i)$, which take a zero value if the constraints \eqref{estimation: admm proof constraint 1},  \eqref{estimation: admm proof constraint 2}, \eqref{estimation: admm proof constraint 3}, \eqref{estimation: admm proof constraint 4}, \eqref{estimation: admm proof constraint 5}, respectively, are satisfied, and become infinite, otherwise. Then, the Augmented Lagrangian (AL) for this problem can be formulated as
\begin{align*}
\calL & = \sum_{i \in \calC_a^{\ell-1}} \hat{J}_i^{\mathrm{e}}(Q_i, q_i)
+ \calI_{h_i}(\tilde{Q}_i, \tilde{q}_i, \tilde{\phi}_i)
+ \calI_{T_i}(Q_i, \phi_i)
\\
& ~~~~~~~~~ 
+ \calI_{Q_i}(Q_i) 
+ \sum_{n \in \calC_i^{\ell}} \calI_{S_{i,n}}(Q_i, q_i, \tau_{i,n})
+ \calI_{\tau_{i,n}}(\tau_{i,n})
\\[0.1cm]
& ~~~~~~~~~ 
+ \tr(\Xi_i^\T (\tilde{Q}_i - \tilde{G}_i)) 
+ \xi_i^\T (\tilde{q}_i - \tilde{g}_i)
+ y_i^\T (\tilde{\phi}_i - \tilde{z}_i)
\\[0.2cm]
& ~~~~~~~~~ 
+ \frac{\rho_Q}{2} \| \tilde{Q}_i - \tilde{G}_i \|_F^2
+ \frac{\rho_q}{2} \| \tilde{q}_i - \tilde{g}_i \|_2^2
+ \frac{\rho_\phi}{2} \| \tilde{\phi}_i - \tilde{z}_i \|_2^2.
\end{align*}

Therefore, the ADMM updates are derived as follows. First, the updates for the variables $\tilde{Q}_i$, $\tilde{q}_i$ and $\tilde{\phi}_i$, are given by
\begin{equation}
\label{estimation: admm proof local 1}
\{\tilde{Q}_i, \tilde{q}_i, \tilde{\phi}_i\} = \argmin \calL
\end{equation}
for all $i \in \calC_a^{\ell-1}$. The minimization in \eqref{estimation: admm proof local 1} leads to the local problems
\begin{subequations}
\begin{align}
& 
\{\tilde{Q}_i, \tilde{q}_i, \tilde{\phi}_i\} = \argmin \tilde{J}_i^{\mathrm{e}}(\tilde{Q}_i, \tilde{q}_i, \tilde{\phi}_i)
\\[0.1cm]
\mathrm{s.t} \quad 
& S_{i,n}(Q_i, q_i, \tau_{i,n}) \succeq
0, \ n \in \calC_i^{\ell},
\\
& \tau_{i,n} \geq 0, \ n \in \calC_i^{\ell},
\\
& h_i(\tilde{Q}_i, \tilde{q}_i, \tilde{\phi}_i) \leq 0,
\\
& T_i(Q_i, \phi_i) \succeq 0,
\\ 
& Q_i \succ 0,
\end{align}
\end{subequations}
where 
\begin{align}
\tilde{J}_i^{\mathrm{e}} & = \hat{J}_i^{\mathrm{e}}(Q_i, q_i) + \tr(\Xi_i^\T(\tilde{Q}_i - \tilde{G}_i)) + \xi_i^\T(\tilde{q}_i - \tilde{g}_i) 
\nonumber
\\[0.1cm]
& ~~ + y_i^\T(\tilde{\phi}_i - \tilde{z}_i)
+ \frac{\rho_Q}{2} \| \tilde{Q}_i - \tilde{G}_i \|_F^2
+ \frac{\rho_q}{2} \| \tilde{q}_i - \tilde{g}_i \|_2^2
\nonumber
\\
& ~~ + \frac{\rho_\phi}{2} \| \tilde{\phi}_i - \tilde{z}_i \|_2^2.
\end{align}
Subsequently, the global variables $G$, $g$ and $z$ are updated by 
\begin{equation}
\label{estimation: admm proof global 1}
\{G, g, z \} = \argmin \calL.
\end{equation}
using the updated values of $\tilde{Q}_i$, $\tilde{q}_i$ and $\tilde{\phi}_i$, $\forall i \in \calC_a^{\ell-1}$.
The minimization in \eqref{estimation: admm proof global 1} can be separated for all $G_i$, $g_i$ and $z_i$, leading to the following averaging steps
\begin{subequations}
\begin{align}
G_i & = \frac{1}{|\calm'[\calC_i^\ell]|} \sum_{j \in \calm'[\calC_i^\ell]} Q_i^j
\\
g_i & = \frac{1}{|\calm'[\calC_i^\ell]|} \sum_{j \in \calm'[\calC_i^\ell]} q_i^j
\\
z_i & = \frac{1}{|\calm'[\calC_i^\ell]|} \sum_{j \in \calm'[\calC_i^\ell]} \phi_i^j.
\end{align}
\end{subequations}
After these updates are performed, then the dual variables are updated through dual ascent steps, as follows
\begin{subequations}
\begin{align}
\Xi_i & \leftarrow \Xi_i + \rho_Q (\tilde{Q}_i - \tilde{G}_i)
\\ 
\xi_i & \leftarrow \xi_i + \rho_q (\tilde{q}_i - \tilde{g}_i)
\\
y_i & \leftarrow y_i + \rho_{\phi} (\tilde{\phi}_i - \tilde{z}_i), 
\end{align}
\end{subequations}
by all $i \in \calC_a^{\ell-1}$.

\subsection{Implementation Details}
\label{SM sec: DHDE details}

\subsubsection{Constraint Linearization}

In the local problems \eqref{estimation admm local updates}, all cost terms and constraints are convex, except for the constraint \eqref{DHDE local problem: ellipses overlap constraint 1}. We accommodate for that by linearizing the constraint in every ADMM iteration around the previous values of the included variables, which we denote with $\bar{Q}_i', \bar{q}_i', \phi_i', \bar{Q}_j{}', \bar{q}_j{}', \phi_j{}'$, where we drop the superscript $i$ to lighten the notation. The first order Taylor approximation of $h_{i,j}$ around $(\bar{Q}_i', \bar{q}_i', \phi_i', \bar{Q}_j{}', \bar{q}_j{}', \phi_j{}')$ is denoted by $\bar{h}_{i,j}(\bar{Q}_i, \bar{q}_i, \phi_i, \bar{Q}_j, \bar{q}_j, \phi_j)$ where 
\begin{align*}
\bar{h}_{i,j} & =
h_{i,j}(\bar{Q}_i', \bar{q}_i', \phi_i', \bar{Q}_j', \bar{q}_j', \phi_j') 
+ \mathrm{tr} \left( \nabla_{\bar{Q}_i} h_{i,j} \Bigr\rvert_{\bar{Q}_i'}^\T (\bar{Q}_i - \bar{Q}_i') \right)
\\
& ~~~~ 
+ \mathrm{tr} \left( \nabla_{\bar{Q}_j} h_{i,j} \Bigr\rvert_{\bar{Q}_j'}^\T (\bar{Q}_j - \bar{Q}_j') \right)
+ \nabla_{\bar{q}_i} h_{i,j} \Bigr\rvert_{\bar{q}_i'}^\T (\bar{q}_i - \bar{q}_i')
\\
& ~~~~ 
+ \nabla_{\bar{q}_j} h_{i,j} \Bigr\rvert_{\bar{q}_j'}^\T (\bar{q}_j - \bar{q}_j')
+ \frac{\partial h_{i,j}}{\partial \phi_j}
\biggr\rvert_{\phi_j'} 
(\phi_j - \phi_j') 
\\
& ~~~~ + \frac{\partial h_{i,j}}{\partial \phi_i}
\biggr\rvert_{\phi_i'} 
(\phi_i - \phi_i'), 
\end{align*}
with 
\begin{align*}
\nabla_{\bar{Q}_i} h_{i,j} & = \frac{1}{\| \omega_{i,j} \|_2} \bar{Q}_i^{-\mathrm{T}} \omega_{i,j} (\bar{Q}_i^{-1} \bar{q}_i)^\T
\\[0.2cm]
\nabla_{\bar{Q}_j} h_{i,j} & = - \frac{1}{\| \omega_{i,j} \|_2} \bar{Q}_j^{-\mathrm{T}} \omega_{i,j} (\bar{Q}_j^{-1} \bar{q}_j)^\T
\\[0.2cm]
\nabla_{\bar{q}_i} h_{i,j} & = - \frac{1}{\| \omega_{i,j} \|_2} \bar{Q}_i^{-\mathrm{T}} \omega_{i,j}
\\[0.2cm]
\nabla_{\bar{q}_j} h_{i,j} & = \frac{1}{\| \omega_{i,j} \|_2} \bar{Q}_j^{-\mathrm{T}} \omega_{i,j}
\\[0.2cm]
\omega_{i,j} & = \bar{Q}_i^{-1} \bar{q}_i - \bar{Q}_j^{-1} \bar{q}_j,  
\\[0.2cm]
\frac{\partial h_{i,j}}{\partial \phi_i} & = 
- \frac{1}{2} \phi_{i}^{-3/2}, 
\quad 
\frac{\partial h_{i,j}}{\partial \phi_j} = 
- \frac{1}{2} \phi_{j}^{-3/2}.
\end{align*}

\subsubsection{Termination Criterion}

We suggest two options for the termination criterion in Line 10 of Alg. \ref{DHDE Algorithm}. The first one that would not require any additional communication would be to just set a maximum amount of ADMM iterations. The second option would be to also check whether the ADMM primal and dual residuals norms are below some prespecified thresholds to allow for early termination. In particular, the primal residuals norms are given by 
\begin{align*}
\epsilon_{\mathrm{primal},1} & = 
\sum_{i \in \calC_a^{\ell-1}}
\| \tilde{Q}_i - \tilde{T}_i \|_F,
\\
\epsilon_{\mathrm{primal},2} & = 
\sum_{i \in \calC_a^{\ell-1}}
\| \tilde{q}_i - \tilde{t}_i \|_2,
\\
\epsilon_{\mathrm{primal},3} & = 
\sum_{i \in \calC_a^{\ell-1}}
\| \tilde{\phi}_i - \tilde{z}_i \|_2,
\end{align*}
while the dual residuals norms are given by
\begin{align*}
\epsilon_{\mathrm{dual},1} & = 
\rho_Q  \sum_{i \in \calC_a^{\ell-1}}
\|  \tilde{T}_i - \tilde{T}_{i,\mathrm{prev}} \|_F,
\\
\epsilon_{\mathrm{dual},2} & = 
\rho_q  \sum_{i \in \calC_a^{\ell-1}}
\|  \tilde{t}_i - \tilde{t}_{i,\mathrm{prev}} \|_2,
\\
\epsilon_{\mathrm{dual},3} & = 
\rho_\phi  \sum_{i \in \calC_a^{\ell-1}}
\|  \tilde{z}_i - \tilde{z}_{i,\mathrm{prev}} \|_2.
\end{align*}
Note that the latter approach would require all agents $i \in \calC_a^{\ell-1}$ sending their variables to agent $a$ so that the residuals are computed.

\section{DHDS Details}

\subsection{Detailed Expressions}
\label{SM sec: DHDS detailed expressions}

The decision variables $\bar{u}_i \in \Rb^{N n_u}$, $L_i \in \Rb^{N n_u \times n_x}$ and $K_i \in \Rb^{N n_u \times N n_x}$ are given by $\bar{u}_i = [\bar{u}_{i,0}; \dots; \bar{u}_{i,N-1}]$, 
\begin{align*}
& ~~~~~~~~~~~~ \bar{u}_i = 
\begin{bmatrix}
\bar{u}_{i,0}^\T & \bar{u}_{i,1}^\T & \cdots & \bar{u}_{i,N-1}^\T
\end{bmatrix}^{\T},
\\[0.2cm]
& ~~~~~~~~~~~~ L_i = 
\begin{bmatrix}
L_{i,0}^\T & L_{i,1}^\T & \cdots & L_{i,N-1}^\T
\end{bmatrix}^{\T},
\\[0.2cm]
K_i & = 
\begin{bmatrix}
0 & 0 & \dots & 0 & 0
\\
K_{i,(0,0)} & 0 & \dots & 0 & 0
\\
K_{i,(1,0)} & K_{i,(1,1)} & \dots & 0 & 0
\\
\vdots & \vdots & \ddots & \vdots & \vdots
\\
K_{i,(N-2,0)} & K_{i,(N-2,1)} & \dots & K_{i,(N-2,N-2)} & 0
\end{bmatrix}.
\end{align*}
The matrices $\Psi_0$, $\Psi_u$ and $\Psi_w$ have the following form
\begin{align*}
& \Psi_{0} =
\begin{bmatrix}
I & A^{\T} & \cdots & A^N{}^{\T}
\end{bmatrix}^{\T},
\\[0.2cm]
& \Psi_{u} = \begin{bmatrix} 0 & 0 & \dots & 0 \\
B &  0 & \cdots & 0 \\
A B & B & \cdots & 0 \\
\vdots & \vdots & \ddots & \vdots \\
A^{N-1} B & A^{N-2} B &
\cdots & B\end{bmatrix},
\\[0.2cm]
& \Psi_{w} = \begin{bmatrix} 0 & 0 & \dots & 0 \\
I &  0 & \cdots & 0 \\
A & I & \cdots & 0 \\
\vdots & \vdots & \ddots & \vdots \\
A^{N-1} & A^{N-2} &
\cdots & I \end{bmatrix}.
\end{align*}
The mean state $\mu_{i,k}$ is given by 
\begin{equation}
\mu_{i,k} = f_{i,k}(\bar{u}_i) =  P_{k} f_i(\bar{u}_i),
\end{equation}
where
\begin{equation}
f_i(\bar{u}_i) = \Psi_0 \mu_{i,0} + \Psi_u \bar{u}_i,
\end{equation}
and $P_{k} := \begin{bmatrix}
0, \dots, I, \dots, 0
\end{bmatrix} \in \Rb^{N_x \times (N+1) N_x}$. Furthermore, the state covariance $\Sigma_{i,k}$ is given by
\begin{equation}
\Sigma_{i,k} = F_{i,k}(L_i, K_i) = P_k F_i(L_i, K_i) P_k^\T,
\end{equation}
where
\begin{align*}
F_i(L_i, K_i) & := (\Psi_0 + \Psi_u L_i) \Sigma_{i,0} (\Psi_0 + \Psi_u L_i)^\T
\nonumber
\\
& ~~~~ + (\Psi_w + \Psi_u K_i) \bar{W} (\Psi_w + \Psi_u K_i)^\T.
\end{align*}
with $\bar{W} = \mathrm{blkdiag}(W, \dots, W) \in \Sb_{N n_x}^+$.

\subsection{Proof of Proposition \ref{proposition: multi clique steering}}
\label{SM sec: proposition 3 proof}

First, let us show the equivalence between costs \eqref{multi clique steering problem: cost} and \eqref{prop: multi clique steering problem: cost}. The cost function $J_i^{\mathrm{s}}(u_i^{\ell})$ can be rewritten as
%
\begin{align*}
J_i^{\mathrm{s}}(u_i)
& = \Eb[u_i^\T \bar{R} u_i] 
= \Eb \big[ \tr(\bar{R}_i u_i u_i^\T) \big]
= \tr \big( \Eb[\bar{R}_i u_i u_i^\T ] \big).
\end{align*}
Using \eqref{control policy}, we obtain 
\begin{align*}
J_i^{\mathrm{s}}(u_i) & = \hat{J}_i^{\mathrm{s}}(\bar{u}_i, L_i, K_i)
\\ 
& =
\tr \big( \Eb[\bar{R}_i (\bar{u}_i + L_i \tilde{x}_{i,0} + K_i w_i) (\bar{u}_i + L_i \tilde{x}_{i,0} + K_i w_i)^\T] \big)
\\
& = 
\tr( \bar{R} \bar{u}_i \bar{u}_i^\T + \bar{R} K_i W K_i^\T+ \bar{R} L_i \Sigma_{i,0} L_i^\T)
\\ 
& = 
\bar{u}_i^\T \bar{R} \bar{u}_i + \tr(\bar{R} K_i W K_i^\T+ \bar{R} L_i \Sigma_{i,0} L_i^\T),
\end{align*}
where $\tilde{x}_{i,0} = x_{i,0} - \mu_{i,0}$ and we used the facts that $\Eb[\tilde{x}_{i,0}] = 0$, $\Eb[\tilde{x}_{i,0} w_i^\T] = 0$, $\Eb[w_i w_i^\T] = \bar{W}$ and $\Eb[\tilde{x}_{i,0} \tilde{x}_{i,0}^\T] = \Sigma_{i,0}$.
Furthermore, the dynamics constraints \eqref{multi clique steering problem: dynamics} are implicitly satisfied since in all expressions we use \eqref{state mean covs} for the state means and covariances. 

It is also trivial to show that the constraint $\calf_{i,N}(\bar{u}_i) = 0$ is equivalent to $\Eb[x_{i,N}] = \mu_{i,\mathrm{f}}$. Moreover, if we write $F_i(L_i, K_i) = \Phi_i(L_i, K_i) \Phi_i(L_i, K_i)^\T$ with  
\begin{equation}
\Phi_i(L_i, K_i) =
\begin{bmatrix}
(\Psi_0 + \Psi_u L_i) & (\Psi_w + \Psi_u K_i) 
\end{bmatrix} 
\Omega_i,
\end{equation}
where $\Omega_i \Omega_i^\T = \mathrm{blkdiag}(\Sigma_{i,0}, W)$ and define $\Phi_{i,k}(L_i, K_i) = P_k \Phi_i(L_i, K_i)$, then the constraint $\Sigma_{i,\mathrm{f}} \succeq F_{i,N}(L_i, K_i) =  \Phi_{i,k}(L_i, K_i) \Phi_{i,k}(L_i, K_i)^\T$ is equivalent with 
\begin{equation}
\calF_{i,N}(L_i, K_i) = 
\begin{bmatrix}
\Sigma_{i,\mathrm{f}} & \Phi_{i,N}(L_i, K_i) \\ 
\Phi_{i,N}(L_i, K_i)^\T & I
\end{bmatrix}
\succeq 0
\end{equation}
by using the Schur complement of $\calF_{i,N}(L_i, K_i)$ w.r.t. $I$.
Subsequently, we show that if the constraints $\calQ_{i,k}(L_i, K_i) \succeq 0$ and $\calq_{i,j,k}(\bar{u}_i, \bar{u}_j) \geq 0$ are satisfied, then the constraint $q_{i,j,k}(p_{i,k}, p_{j,k}) \geq 0$ is satisfied as well. In particular, the latter constraint will be true if the following inequalities hold,
\begin{subequations}
\begin{align}
& \| \bar{\mu}_{i,k} - \bar{\mu}_{j,k} \|_2 \geq d_{\mathrm{inter}} + 2r,
\label{proof mean}
\\
& ~~~~~ \sqrt{\alpha \lambda_{\mathrm{max}} (\bar{\Sigma}_{i,k})} \leq r,
\label{proof cov}
\end{align}
\end{subequations}
where we drop the superscripts $\ell$ for notational convenience. If we plug the mean state expressions into \eqref{proof mean}, then we obtain 
\begin{equation}
\| \bar{f}_{i,k}(\bar{u}_i) - \bar{f}_{j,k}(\bar{u}_j) \|_2 \geq d_{\mathrm{inter}} + 2r, 
\end{equation}
which yields the constraint $\calq_{i,j,k}(\bar{u}_i, \bar{u}_j) \geq 0$. Furthermore, the constraint \eqref{proof cov} can be rewritten as
\begin{equation}
\lambda_{\mathrm{max}} (\bar{\Sigma}_{i,k}) \leq \frac{r^2}{\alpha}.
\end{equation}
which is equivalent with 
\begin{equation}
\label{proof cov 2}
H F_{i,k}(L_i, K_i) H^\T - \frac{r^2}{\alpha} \preceq 0.
\end{equation}
or using again the Schur complement with
\begin{equation}
\begin{bmatrix}
\big(\frac{r^\ell}{\sqrt{\alpha}}\big)^2 I & H \Phi_{i,k}(L_i, K_i) \\ 
\Phi_{i,k}(L_i, K_i)^\T H^\T & I
\end{bmatrix} \succeq 0
\end{equation}
which is identical with $\calQ_{i,k}(L_i, K_i) \succeq 0$.
With similar arguments, it can be shown that if the constraints $\calQ_{i,k}(L_i, K_i) \succeq 0$ and $\cals_{i,k} (\bar{u}_i) \geq 0$ are satisfied, then the constraint $s_{i,k}(p_{i.k}) \geq 0$ is also satisfied. 

Finally, we wish to show that if the constraint $\calp(\bar{u}_i) \leq 0$ is true, then the constraint \eqref{multi clique steering problem: in ellipse constraint} is also true. Since \eqref{proof cov} holds, then it suffices to enforce a contraint that $\bar{\mu}_{i,k}^\ell$ should lie within an ellipse with center $\bar{\mu}_{a,k}^{(\ell-1)}$, major axis length $\sqrt{\alpha \lambda_{\mathrm{max}} ( \bar{\Sigma}_{a,k}^{(\ell-1)})}
- r$, minor axis length $\sqrt{\alpha \lambda_{\mathrm{min}} ( \bar{\Sigma}_{a,k}^{(\ell-1)})}
- r$, and the same orientation as the ellipse $\calE_\theta [\bar{\mu}_{a,k}^{\ell-1} ,  \bar{\Sigma}_{a,k}^{\ell-1}]$. 
These specifications can be captured if the following inequality holds
\begin{equation}
\label{proof nice ellipse}
(\bar{\mu}_{i,k}^\ell - \bar{\mu}_{a,k}^{\ell-1})^\T \hat{P} (\bar{\mu}_{i,k}^\ell-\bar{\mu}_{a,k}^{\ell-1}) \leq 1, 
\end{equation}
where 
\begin{align*}
& ~~~ \hat{P} = \frac{1}{\alpha} U \hat{\Lambda}^{-1} U^\T,
\\
& \hat{\Lambda} 
= \left( \Lambda^{1/2} - \frac{r}{\sqrt{\alpha}} I  \right)^2,
\end{align*}
and $[\Lambda,U]$ is the eigendecomposition of $\bar{\Sigma}_{a,k}^{\ell-1}$. This is true since the ellipse $\hat{P}$ and $\bar{\Sigma}_{a,k}^{\ell-1}$ have the same eigenvectors, the major axis length of the ellipse in \eqref{proof nice ellipse} is 
\begin{align*}
\sqrt{ \frac{1}{\lambda_{\min}(\hat{P})}} 
& = \sqrt{\alpha \frac{1}{\lambda_{\min}(\hat{\Lambda}^{-1})}}
= \sqrt{\alpha \lambda_{\max}(\hat{\Lambda})}
\\
& = \sqrt{\alpha} \lambda_{\max} \Big(\Lambda^{1/2} - \frac{r}{\sqrt{\alpha}} I \Big)
\\
& = \sqrt{\alpha} \Big( \lambda_{\max} \big(\Lambda^{1/2} \big) - \frac{r}{\sqrt{\alpha}} \Big)
\\
& = \sqrt{\alpha \lambda_{\mathrm{max}} ( \bar{\Sigma}_{a,k}^{(\ell-1)})}
- r
\end{align*}
and similarly it can be shown that the minor axis length is 
\begin{align*}
\sqrt{ \frac{1}{\lambda_{\max}(\hat{P})}} 
= \sqrt{\alpha \lambda_{\mathrm{min}} ( \bar{\Sigma}_{a,k}^{(\ell-1)})}
- r.
\end{align*}

\subsection{ADMM Derivation}
\label{SM sec: DHDS admm}
The derivation is similar with the one in Section \ref{SM sec: DHDE admm} of the SM. With the introduction of the augmented variables $\tilde{u}_i$ and global variable $b$, problem \eqref{DHDS ubar: problem} can be reformulated as
\begin{subequations}
\begin{align}
& \{ \bar{u}_i \}_{i \in \calC_a^{\ell-1}} = \argmin \sum_{i \in \calC_a^{\ell-1}} 
\hat{J}_{i,1}^{\mathrm{s}}(\bar{u}_i)
\\[0.1cm]
\mathrm{s.t.} \quad
& \calf_{i,N}(\bar{u}_i) = 0, ~
\cals_{i,k}(\bar{u}_i) \geq 0, ~ 
\calp_{i,k}(\bar{u}_i) \leq 0, 
\\[0.1cm]
& \calq_{i,k}(\tilde{u}_i) \geq 0,
~ \ k \in \llbracket 0, N \rrbracket,
\\[0.1cm]
& \tilde{u}_i = \tilde{b}_i, \   
i \in \calC_a^{\ell-1},
\end{align}
\end{subequations}
with $h_i = [\{ h_{i,j}(\bar{u}_i, \bar{u}_j^i) \}_{j \in \caln[\calC_i^\ell]}]$. The AL for this problem is given by 
\begin{align*}
\calL & = \sum_{i \in \calC_a^{\ell-1}} \hat{J}_i^{\mathrm{s}}(\bar{u}_i)
+ \calI_{\calf_i}(\bar{u}_i)
+ \calI_{\cals_i}(\bar{u}_i)
+ \calI_{\calp_i}(\bar{u}_i)
\\
& ~~~~~~~~~~~ + \calI_{\calq_i}(\tilde{u}_i) 
+ v_i^\T (\tilde{u}_i - \tilde{b}_i)
+ \frac{\rho_u}{2} \| \tilde{u}_i - \tilde{b}_i \|_2^2, 
\end{align*}
where the indicator functions are of the same form as in Section \ref{SM sec: DHDE admm}. The updates for the variables $\tilde{u}_i$, are given by $\tilde{u}_i = \argmin \calL$, which leads to the local problems
\begin{subequations}
\begin{align}
& ~~~~~~~~~ \tilde{u}_i = \argmin 
\tilde{J}_{i,1}^{\mathrm{s}}(\tilde{u}_i)
\\[0.1cm]
\mathrm{s.t.} \quad
& \calf_{i,N}(\bar{u}_i) = 0, ~
\cals_{i,k}(\bar{u}_i) \geq 0, ~ 
\calp_{i,k}(\bar{u}_i) \leq 0, 
\\[0.1cm]
& \calq_{i,j,k}(\bar{u}_i, \bar{u}_j^i) \geq 0,
~  j \in \caln[\calC_i^\ell], ~ k \in \llbracket 0, N \rrbracket,
\end{align}
\end{subequations}
with 
\begin{align}
\tilde{J}_{i,1}^{\mathrm{s}}(\tilde{u}_i) & = \hat{J}_{i,1}^{\mathrm{s}}(\bar{u}_i) 
+ v_i^\T(\tilde{u}_i - \tilde{b}_i)
+ \frac{\rho_u}{2} \| \tilde{u}_i - \tilde{b}_i \|_2^2.
\end{align}
The global update given by $b = \argmin \calL$, leads to the update rules
\begin{equation}
b_i = \frac{1}{|\calm'[\calC_i^\ell]|} \sum_{j \in \calm'[\calC_i^\ell]} \bar{u}_i^j
\end{equation}
using the updated values of $\bar{u}_i^j$. Finally, the dual updates are given by 
\begin{equation}
\label{DHDS ubar: dual update supp}
v_i \leftarrow v_i + \rho_{u} (\tilde{u}_i - \tilde{b}_i).
\end{equation}

\subsection{Implementation Details}
\label{SM sec: DHDS details}
\subsubsection{Constraint Linearization}

In problems \eqref{DHDS ubar: local update}, all cost terms and constraints are convex, except for the constraints $\calq_{i,j,k}(\bar{u}_i, \bar{u}_j) \geq 0$ and $\cals_{i,k} (\bar{u}_i) \geq 0$. To address these non-convexities, we linearize the constraints in every ADMM iteration around $\bar{u}_i', \bar{u}_j'$, which are the previous values of $\bar{u}_i, \bar{u}_j$. Thus, we replace the aforementioned constraints with 
\begin{subequations}
\begin{align}
\bar{\calq}_{i,j,k}(\bar{u}_i, \bar{u}_j^i) & \geq 0,
~  j \in \caln[\calC_i^\ell], ~ k \in \llbracket 0, N \rrbracket,
\\[0.1cm]
\bar{\cals}_{i,k}(\bar{u}_i) & \geq 0, ~ k \in \llbracket 0, N \rrbracket,
\end{align}
\end{subequations}
where
\begin{align*}
\bar{\calq}_{i,j,k}(\bar{u}_i, \bar{u}_j) & =
\calq_{i,j,k}(\bar{u}_i', \bar{u}_j')
+ \nabla_{\bar{u}_i} \calq_{i,j,k} \Bigr\rvert_{\bar{u}_i'}^\T (\bar{u}_i - \bar{u}_i')
\\
& ~~~~ 
+ \nabla_{\bar{u}_j} \calq_{i,j,k} \Bigr\rvert_{\bar{u}_j'}^\T (\bar{u}_j - \bar{u}_j'), 
\\
\bar{\cals}_{i,k}(\bar{u}_i) & =
\cals_{i,k}(\bar{u}_i')
+ \nabla_{\bar{u}_i} \cals_{i,k} \Bigr\rvert_{\bar{u}_i'}^\T (\bar{u}_i - \bar{u}_i')
\end{align*}
and
\begin{align*}
\nabla_{\bar{u}_i} \calq_{i,j,k} & = \frac{1}{\| \zeta_{i,j,k} \|_2} (H P_k \Psi_u)^\T \zeta_{i,j,k}
\\[0.2cm]
\nabla_{\bar{u}_j} \calq_{i,j,k} & = - \frac{1}{\| \zeta_{i,j,k} \|_2} (H P_k \Psi_u)^\T \zeta_{i,j,k}
\\[0.2cm]
\zeta_{i,j,k} & = H P_k \big( \Psi_0 (\mu_{i,0} - \mu_{j,0}) + \Psi_u (\bar{u}_i - \bar{u}_j) \big)
\\[0.2cm]
\nabla_{\bar{u}_i} \cals_{i,k} & = \frac{1}{\| \eta_{i,k} \|_2} (H P_k \Psi_u)^\T\eta_{i,k}
\\[0.2cm]
\eta_{i,k} & = H P_k \big( \Psi_0 \mu_{i,0} + \Psi_u \bar{u}_i \big) - p_o.
\end{align*}

\subsubsection{Termination Criterion}

The termination criterion in Line 10 of Alg. \ref{DHDS Algorithm} is similar with one presented in Section \ref{SM sec: DHDE details} of the SM. In particular, we either set a maximum amount of ADMM iterations or check whether the residual norms 
\begin{align*}
\epsilon_{\mathrm{primal}} & = 
\sum_{i \in \calC_a^{\ell-1}}
\| \tilde{u}_i - \tilde{b}_i \|_2,
\\
\epsilon_{\mathrm{dual}} & = 
\rho_u  \sum_{i \in \calC_a^{\ell-1}}
\|  \tilde{b}_i - \tilde{b}_{i,\mathrm{prev}} \|_2,
\end{align*}
are below some predefined thresholds. Note that in the latter case, all agents $i \in \calC_a^{\ell-1}$ would be required to send their variables to agent $a$ during every ADMM iteration. 

\section{Simulation Details}
\label{SM sec: sim details}

In the simulation experiments, all agents are modeled with 2D double integrator dynamics which are discretized with $dt = 0.05s$. The time horizon is $N=100$ for all tasks. For the first two tasks, the noise covariance is $W=\mathrm{diag}(0.02,0.02,0.2,0.2)^2$, while for the third one it is $W=\mathrm{diag}(10^{-3},10^{-3},10^{-2},10^{-2})^2$. For all tasks, we set $\theta = 0.997$. For both DHDE and DHDS, the maximum amount of ADMM iterations is set to $20$. All penalty parameters are selected to be $\rho = 10^3$.

\end{document}